\def\year{2022}\relax
\documentclass[letterpaper]{article} 
\usepackage{aaai22}  
\usepackage{times}  
\usepackage{helvet}  
\usepackage{courier}  
\usepackage[hyphens]{url}  
\usepackage{graphicx} 
\usepackage{natbib}  
\usepackage{caption} 
\DeclareCaptionStyle{ruled}{labelfont=normalfont,labelsep=colon,strut=off} 
\usepackage{algorithm}
\usepackage{algorithmic}

\usepackage{amsmath}
\usepackage{amssymb}
\usepackage{amsthm}
\usepackage{units}
\usepackage{babel}
\usepackage{tikz}
\usepackage{multirow}
\usetikzlibrary{shapes.geometric, arrows, automata, positioning}
\usepackage{makecell}
\usepackage{bbding}
\usepackage{dsfont}
\usepackage{bbm}
\usepackage{colortbl}
\usepackage{wrapfig}
\usepackage{soul}

\newtheorem{theorem}{Theorem}
\newtheorem{lemma}{Lemma}

\newtheorem{corollary}{Corollary}

\newtheorem{definition}{Definition}
\newtheorem{remark}{Remark}

\newenvironment{proofsketch}{%
  \proof}{\endproof}

\global\long\def\E{\mathbb{E}}
\newcommand{\ind}{\mathbb{I}}
\newcommand{\indevent}[1]{\ind \{ #1 \}}
\newcommand{\Kexp}{\mathcal{K}_{exp}}

\newcommand{\picheat}{\bar{\pi}}
\newcommand{\regret}{\mathcal{R}_K}
\newcommand{\wt}{\widetilde}
\newcommand{\Kcal}{\mathcal{K}}
\newcommand{\Fcal}{\mathcal{F}}
\newcommand{\Mcal}{\mathcal{N}}
\newcommand{\Scal}{\mathcal{S}}
\newcommand{\Acal}{\mathcal{A}}
\newcommand{\Ksa}{\mathcal{K}}
\newcommand{\epsilontilde}{\tilde{\epsilon}}

\usepackage[capitalise,nameinlink]{cleveref}
\Crefname{theorem}{Theorem}{Theorems}
\Crefname{corollary}{Corollary}{corollaries}

\usepackage{newfloat}
\usepackage{listings}

\floatstyle{ruled}
\newfloat{listing}{tb}{lst}{}
\floatname{listing}{Listing}
\title{Learning Adversarial Markov Decision Processes with Delayed Feedback}
\author{
    Tal Lancewicki,\equalcontrib\textsuperscript{\rm 1} Aviv Rosenberg,\equalcontrib\textsuperscript{\rm 1}
    Yishay Mansour\textsuperscript{\rm 1,2}
}
\affiliations{
   \textsuperscript{\rm 1} Tel Aviv University, Israel \\
    \textsuperscript{\rm 2} Google Research, Israel \\
    lancewicki@mail.tau.ac.il,
    avivros007@gmail.com, mansour.yishay@gmail.com
}

\begin{document}

\maketitle

\begin{abstract}
    Reinforcement learning typically assumes that agents observe feedback for their actions immediately, but in many real-world applications (like recommendation systems) feedback is observed in delay.
    This paper studies online learning in episodic Markov decision processes (MDPs) with unknown transitions, adversarially changing costs and unrestricted delayed feedback.
    That is, the costs and trajectory of episode $k$ are revealed to the learner only in the end of episode $k + d^k$, where the delays $d^k$ are neither identical nor bounded, and are chosen by an oblivious adversary.
    We present novel algorithms based on policy optimization that achieve near-optimal high-probability regret of $\sqrt{K + D}$ under full-information feedback, where $K$ is the number of episodes and $D = \sum_{k} d^k$ is the total delay.
    Under bandit feedback, we prove similar $\sqrt{K + D}$ regret assuming the costs are stochastic, and $(K + D)^{2/3}$ regret in the general case.
    We are the first to consider regret minimization in the important setting of MDPs with delayed feedback.
\end{abstract}

\section{Introduction}

Delayed feedback is a fundamental challenge in sequential decision making arising in almost all practical applications.
For example, recommendation systems learn the utility of a recommendation by detecting occurrence of certain events (e.g., user conversions), which may happen with a variable delay
after the recommendation was issued.
Other examples include display advertising, autonomous vehicles, video streaming \citep{changuel2012online}, delays in communication between learning agents \citep{chen2020delay} and system delays in robotics \citep{mahmood2018setting}.

Although handling feedback delays is crucial for applying reinforcement learning (RL) in practice, it was only barely studied from a theoretical perspective, as most of the RL literature focuses on the MDP model in which the agent observes feedback regarding her immediate reward and transition to the next state right after performing an action.

This paper makes a substantial step towards closing the major gap on delayed feedback in the RL literature.
We consider the challenging adversarial episodic MDP setting where cost functions change arbitrarily between episodes while the transition function remains stationary over time (but unknown to the agent).
We present the \emph{adversarial MDP with delayed feedback} model in which the agent observes feedback for episode $k$ only in the end of episode $k + d^k$, where the delays $d^k$ are unknown and not restricted in any way.
This model generalizes standard adversarial MDPs (where $d^k=0 \  \forall k$), and encompasses great challenges that do not arise in standard RL models, e.g., exploration without feedback and latency in policy updates. 
Adversarial models are extremely important in practice, as they allow dependencies between costs, unlike stochastic models that assume i.i.d samples. 
This is especially important in the presence of delays (that are also adversarial in our model), since it allows dependencies between costs and delays which are well motivated in practice \citep{lancewicki2021stochastic}.

We develop novel policy optimization (PO) algorithms that perform their updates whenever feedback is available and ignore feedback with large delay, and prove that they obtain high-probability regret bounds of order $\sqrt{K + D}$ under full-information feedback and $(K + D)^{2/3}$ under bandit feedback, where $K$ is the number of episodes and $D$ is the sum of delays.
Unlike simple reductions that can only handle fixed delay $d$, our algorithms are robust to any kind of variable delays and do not require any prior knowledge.
Furthermore, we show that a naive adaptation of existing algorithms suffers from sub-optimal dependence in the number of actions, and present a novel technique that forces exploration in order to achieve tight bounds.
To complement our results, we present nearly matching lower bounds of order $\sqrt{K + D}$.
See detailed bounds in \cref{table:comparison}.

\begin{table*}[t]
    \scriptsize
   \centering
    \begin{tabular}{|c|c|c|c|}
        \hline
        & \makecell{Known Transition \\ + Delayed Trajectory} & \makecell{Unknown Transition \\ + Delayed Cost} & \makecell{Unknown Transition \\ + Delayed Trajectory}
        \\
         \hline \hline
        \rowcolor{lightgray} D-O-REPS (full) & $H \sqrt{K + D}$ & $H^{3/2} S \sqrt{A K} + H \sqrt{D}$ & $H^2 S \sqrt{A K} + H^{3/2} \sqrt{S D}$
        \\ [2pt]
        \hline
        \rowcolor{lightgray} D-OPPO (full) & $H^2 \sqrt{K + D}$ & $H^{3/2} S \sqrt{A K} + H^2 \sqrt{D}$ & $H^2 S \sqrt{A K} + H^{3/2} \sqrt{S D}$
        \\ [2pt]
        \hline
        Lower Bound (full) & $H \sqrt{K + D}$ & $H^{3/2} \sqrt{S A K} + H \sqrt{D}$ & $H^{3/2} \sqrt{S A K} + H \sqrt{D}$
        \\ [2pt]
        \hline \hline
        \rowcolor{lightgray} D-OPPO (bandit) & $H S \sqrt{A} K^{2/3} + H^2 D^{2/3}$ & $H S \sqrt{A} K^{2/3} + H^2 D^{2/3}$ & $H S \sqrt{A} K^{2/3} + H^2 D^{2/3}$
        \\ [2pt]
        \hline
        Lower Bound (bandit) & $H \sqrt{S A K} + H \sqrt{D}$ & $H^{3/2} \sqrt{S A K} + H \sqrt{D}$ & $H^{3/2} \sqrt{S A K} + H \sqrt{D}$
        \\ [2pt]
        \hline
    \end{tabular}
    \caption{\small Regret bounds comparison (ignoring constant and poly-logarithmic factors) between our algorithms Delayed OPPO (D-OPPO) and Delayed O-REPS (D-O-REPS), and our lower bound under full-information (full) and bandit feedback (bandit). 
    ``Known Transition'' assumes dynamics are known to the learner in advance, and ``Unknown Transition'' means that the learner needs to learn the dynamics.
    ``Delayed Cost'' assumes only costs are observed in delay, while in ``Delayed Trajectory'' the trajectory is also observed in delay, together with the costs.}
    \label{table:comparison}
\end{table*}

\subsection{Related work}   \label{sec:related_work}

\paragraph{Delays in RL.}
Although delay is a common challenge RL algorithms need to face in practice \citep{schuitema2010control,liu2014impact,changuel2012online,mahmood2018setting}, the theoretical literature on the subject is very limited.
Previous work only studied \emph{delayed state observability} \citep{katsikopoulos2003markov} where the state is observable in delay and the agent picks actions without full knowledge of its current state.
This setting is much related to partially observable MDPs (POMDPs) and motivated by scenarios like robotics system delays.
Unfortunately, even planning is computationally hard (exponential in the delay $d$) for delayed state observability \citep{walsh2009learning}.

This paper studies a different setting that we call \emph{delayed feedback}, where the delay only affects the information available to the agent, and not the execution of its policy.
Delayed feedback is also an important setting, as it is experienced in recommendation systems and applications where the policy is executed by a different computational unit than the main algorithm (e.g., policy is executed by a robot with limited computational power, while heavy computations are done by the main algorithm on another computer that receives data from the robot in delay).
Importantly, unlike delayed state observability, it is not computationally hard to handle delayed feedback, as we show in this paper.
The challenges of delayed feedback are very different than the ones of delayed state observability, and include policy updates that occur in delay and exploration without observing feedback.

\paragraph{Delays in multi-armed bandit.}
Delays were extensively studied in MAB recently as a fundamental issue that arises in many real applications \citep{vernade2017stochastic,pike2018bandits,cesa2018nonstochastic,zhou2019learning,manegueu2020stochastic,lancewicki2021stochastic}. 
Our work is most related to the literature on delays in adversarial MAB, starting with \citet{cesa2016delay} that showed the optimal regret for MAB with fixed delay $d$ is of order $\sqrt{(A + d) K}$, where $A$ is the number of actions.
Even earlier, variable delays were studied by \citet{quanrud2015online} in online learning with full-information feedback, where they showed optimal $\sqrt{K + D}$ regret.
More recently, \citet{thune2019nonstochastic,bistritz2019online,zimmert2020optimal,gyorgy2020adapting} studied variable delays in MAB, proving optimal $\sqrt{AK + D}$ regret.
Unlike MDPs, in MAB there is no underlying dynamics, and the only challenge is feedback about the cost arriving in delay.

\paragraph{Regret minimization in stochastic MDPs.}
There is a vast literature on regret minimization in RL that mostly builds on the optimism in face of uncertainty principle.
Most literature focuses on the tabular setting, where the number of states is small (see, e.g., \citet{jaksch2010near,azar2017minimax,jin2018q,zanette2019tighter}).
Recently it was extended to function approximation under various assumptions (see, e.g., \citet{jin2020provably,yang2019sample,zanette2020frequentist,zanette2020learning}).

\paragraph{Adversarial MDPs.}
Early works on adversarial MDPs \citep{even2009online,neu2010ossp,neu2012unknown,neu2014bandit} focused on known transitions and used various reductions to MAB.
\citet{zimin2013online} presented O-REPS -- a reduction to online linear optimization achieving optimal regret bounds with known dynamics.
Later, O-REPS was extended to unknown dynamics \citep{rosenberg2019full,rosenberg2019bandit,jin2019learning} obtaining near-optimal regret bounds.
Recently, \citet{cai2019provably,efroni2020optimistic,he2021nearly} proved similar regret for PO methods (that are widely used in practice).

\section{Setting}

An episodic adversarial MDP is defined by  a tuple $\mathcal{M} = ( \mathcal{S} , \mathcal{A} , H , p, \{ c^k \}_{k=1}^K )$, where $\mathcal{S}$ and $\mathcal{A}$ are finite state and action spaces of sizes $S$ and $A$, respectively, $H$ is the episode length, $p = \{ p_h : \mathcal{S} \times \mathcal{A} \to \Delta_{\mathcal{S}} \}_{h=1}^H$ is the transition function, and $c^k = \{ c^k_h : \mathcal{S} \times \mathcal{A}\to [0,1] \}_{h=1}^H$ is the cost function for episode $k$.
For simplicity, $S \ge \max \{ A , H^2 \}$.

The interaction between the learner and the environment proceeds as follows.
At the beginning of episode $k$, the learner starts in a fixed initial state\footnote{The algorithm readily extends to a fixed initial distribution.} $s^k_1 = s_\text{init} \in S$ and picks a policy $\pi^k = \{ \pi^k_{h}: \mathcal{S} \to \Delta_{\mathcal{A}} \}_{h=1}^H$ where $\pi^k_{h}(a \mid s)$ gives the probability that the agent takes action $a$ at time $h$ given that the current state is $s$.
Then, the policy is executed in the MDP generating a trajectory $U^k = \{ (s^k_h,a^k_h) \}_{h=1}^H$, where $a^k_h \sim \pi^k_h(\cdot | s_h^k)$ and $s^k_{h+1} \sim p_h(\cdot | s^k_h,a^k_h)$.
With no delays, the learner observes the feedback in the end of the episode, that is, the trajectory $U^k$ and either the entire cost function $c^k$ under \emph{full-information feedback} or the suffered costs $\{ c^k_h(s^k_h,a^k_h) \}_{h=1}^H$ under \emph{bandit feedback}.
In contrast, with delayed feedback, these are revealed to the learner only in the end of episode $k + d^k$, where the delays $\{ d^k \}_{k=1}^K$ are unknown and chosen by an oblivious adversary before the interaction starts.
Denote the total delay by $D = \sum_{k} d^k$ and the maximal delay by $d_{max} = \max_{k} d^k$.
Note that standard adversarial MDPs are a special case in which $d^k = 0 \ \forall k$.

For a given policy $\pi$, we define its expected cost with respect to cost function $c$, when starting from state $s$ at time $h$, as
$
    V_{h}^{\pi}(s) 
    = 
    \E \bigl[ \sum_{h'=h}^{H}c_{h'}(s_{h'},a_{h'}) | s_{h}=s,\pi,p \bigr],
$
where the expectation is taken over the randomness of the transition function $p$ and the policy $\pi$.
This is known as the \emph{value function} of $\pi$, and we also define the \emph{$Q$-function} by
$
    Q_{h}^{\pi}(s,a) 
    = 
    \E \bigl[ \sum_{h'=h}^{H}c_{h'}(s_{h'},a_{h'})| s_h = s,a_h = a,\pi,p \bigr].
$
It is well-known (see \citet{sutton2018reinforcement}) that the value function and $Q$-function satisfy the Bellman equations:
\begin{align}
    \nonumber
    Q_{h}^{\pi}(s,a) 
    & = 
    c_{h}(s,a)+\langle p_h(\cdot\mid s,a) , V_{h+1}^{\pi}\rangle
    \\
    V_{h}^{\pi}(s)  
    & =
    \langle \pi_{h}(\cdot\mid s),Q_h^\pi(s,\cdot)\rangle,
    \label{eq:bellman2}
\end{align}
where $\langle \cdot, \cdot \rangle$ is the dot product.
Let $V^{k,\pi}$ be the value function of $\pi$ with respect to $c^k$.
We measure the performance of the learner by the \emph{regret} -- the cumulative difference between the value of the learner's policies and the value of the best fixed policy in hindsight, i.e.,
\begin{align*}
    \regret
    =
    \sum_{k=1}^{K} V_{1}^{k,\pi^{k}}(s_1^k) - \min_{\pi} \sum_{k=1}^{K}  V_{1}^{k,\pi}(s_1^k).
\end{align*}

\paragraph{Notations.}
Episode indices appear as superscripts and in-episode steps as subscripts. 
$\mathcal{F}^k = \{j:j+d^j = k\}$ denotes the set of episodes that their feedback arrives in the end of episode $k$, and the of number visits to state-action pair $(s,a)$ at time $h$ by the end of episode $k-1$ is denoted by $m_{h}^{k}(s,a)$. 
Similarly, $n_{h}^{k}(s,a)$ denotes the number of these visits for which feedback was observed until the end of episode $k -1$. 
$\E^\pi[\cdot] = \E[ \cdot \mid s^k_1 = s_\text{init}, \pi, p ]$ denotes the expectation given a policy $\pi$,
the notation $\wt O(\cdot)$ ignores constant and poly-logarithmic factors and $x \vee y = \max \{x,y\}$.
We denote the set $\{1,\dots,n\}$ by $[n]$, and the indicator of event $E$ by $\mathbb{I}\{E\}$.

\section{Warm-up: a black-box reduction}
\label{sec:black-box-red}

One simple way to deal with delays (adopted in several MAB and online optimization works, e.g., \citet{weinberger2002delayed,joulani2013online}) is to simulate a non-delayed algorithm and use its regret guarantees.
Specifically, we can maintain $d_{max} + 1$ instances of the non-delayed algorithm, running the $i$-th instance on the episodes $k$ such that $k = i \mod (d_{max} + 1)$.
That is, at the first $d_{max} + 1$ episodes, the learner plays the first policy that each instance outputs.
By the end of episode $d_{max} + 1$, the feedback for the first episode is observed, allowing the learner to feed it to the first instance.
The learner would then play the second output of that instance, and so on. 
Effectively, each instance plays $\nicefrac{K}{(d_{max}+1)}$ episodes, so we can use the regret of the non-delayed algorithm $\wt{\mathcal{R}}_K$ in order to bound $
    \regret
    \leq 
    (d_{max} + 1) \wt{\mathcal{R}}_{\nicefrac{K}{(d_{max}+1)}}.
$
Plugging in standard adversarial MDP regret bounds \citep{rosenberg2019full,jin2019learning}, we obtain the following regret for both full-information and bandit feedback:
\[
    \regret
    =
    \wt O \bigl( H^2 S \sqrt{A K (d_{max}+1)} + H^2 S^2 A (d_{max}+1) \bigr).
\]

While simple in concept, the black-box reduction suffers from many evident shortcomings.
First, it is highly non-robust to variable delays as its regret scales with the \emph{worst-case delay} $K d_{max}$ which becomes very large even if the feedback from just one episode is missing.
One of the major challenges that we tackle in the rest of the paper is to achieve regret bounds that are independent of $d_{max}$ and scale with the \emph{average delay}, i.e., the total delay $D$ which is usually much smaller than worst-case.
Second, even if we ignore the problematic dependence in the worst-case delay, this regret bound is still sub-optimal as it suggests a multiplicative relation between $d_{max}$ and $A$ (and $S^2$) which does not appear in the MAB setting.
Our analysis focuses on eliminating this sub-optimal dependence through a clever algorithmic feature that forces exploration and ensures tight near-optimal regret.
Finally, the reduction is highly inefficient as it requires running $d_{max}+1$ different algorithms in parallel.
Moreover, the $\sqrt{Kd_{max}}$ regret under bandit feedback is only achievable using O-REPS algorithms that are extremely inefficient to implement in practice.
In contrast, our algorithm is based on efficient and practical PO methods.
Its running time is independent of the delays and it does not require any prior knowledge or parameter tuning (unlike the reduction needs to know $d_{max}$). 
In \cref{sec:discussion} we present experiments showing that our algorithm outperforms generic approaches, such as black-box reduction, not only theoretically but also empirically.

\section{Delayed OPPO}

\begin{algorithm}[t]
    \caption{Delayed OPPO}  
    \label{alg:POMD}
    \begin{algorithmic}
        \STATE \textbf{Input:} $\mathcal{S} , \mathcal{A} , H , \eta > 0 , \gamma > 0 , \delta > 0$.
        \STATE \textbf{Initialization:} 
        Set $\pi_{h}^{1}(a \mid s) = \nicefrac{1}{A}$ for every $(s,a,h)$.
        \FOR{$k=1,2,\dots,K$}
            \STATE Play episode $k$ with policy $\pi^k$.
            
            \STATE Observe feedback from all episodes $j \in \mathcal{F}^k$.
            
            \STATE Compute cost estimators $\hat c^j$ and  confidence set $\mathcal{P}^{k}$.
            
            \STATE {\color{gray} \# Policy Evaluation}
            
            \FOR{$j\in \mathcal{F}^k$}
            
                \STATE Set $V_{H+1}^j(s)=0$ for every $s \in \mathcal{S}$.
                
                \FOR{$h = H,\dots,1$ and $(s,a) \in \mathcal{S} \times \mathcal{A}$}
                
                    \STATE $\hat p_h^j(\cdot | s,a) \in {\arg\min}_{p'_h(\cdot | s,a) \in \mathcal{P}_h^{k}(s,a)} \langle p'_h(\cdot | s,a) , V_{h+1}^j \rangle$.
                    
                    \STATE $Q_{h}^{j}(s,a) = \hat{c}_{h}^{j}(s,a)+\langle \hat{p}_{h}^j(\cdot\mid s,a),V_{h+1}^{j} \rangle$.
                    
                    \STATE $V_{h}^{j}(s) = \langle Q_{h}^{j}(s,\cdot),\pi_{h}^{j}(\cdot \mid s)\rangle$.
                
                \ENDFOR
            
            \ENDFOR
            
            \STATE {\color{gray} \# Policy Improvement}
            
            \STATE $\pi^{k+1}_h(a | s) = \frac{\pi^k_h(a \mid s) \exp \bigl( -\eta \sum_{j\in\mathcal{F}^{k}} Q_{h}^{j}(s,a) \bigr)}{\sum_{a' \in \mathcal{A}} \pi^k_h(a' \mid s) \exp \bigl( -\eta \sum_{j\in\mathcal{F}^{k}} Q_{h}^{j}(s,a') \bigr)}$. 
            
        \ENDFOR
        
    \end{algorithmic}
\end{algorithm}

In this section we present \emph{Delayed OPPO} (\cref{alg:POMD} and with more details in \cref{sec:full-algo}) -- the first algorithm for regret minimization in adversarial MDPs with delayed feedback.
Delayed OPPO is a policy optimization algorithm, and therefore implements a smoother version of Policy Iteration \citep{sutton2018reinforcement}, i.e., it alternates between a policy evaluation step -- where an optimistic estimate for the $Q$-function of the learner's policy is computed, and a policy improvement step -- where the learner's policy is improved in a ``soft'' manner regularized by the KL-divergence.

Delayed OPPO is based on the optimistic proximal policy optimization (OPPO) algorithm \citep{cai2019provably,efroni2020optimistic}.
As a policy optimization algorithm, it enjoys many merits of practical PO algorithms that have had great empirical success in recent years, e.g., TRPO \citep{schulman2015trust}, PPO \citep{schulman2017proximal} and SAC \citep{haarnoja2018soft} -- It is easy to implement, computationally efficient and readily extends to function approximation.

The main difference that Delayed OPPO introduces is performing updates using all the available feedback at the current time step.
Furthermore, in \cref{sec:traj-delay,sec:skipping-and-doubling} we equip our algorithm with novel mechanisms that make it robust to all kinds of variable delays without any prior knowledge and enable us to prove tight regret bounds.
Importantly, these mechanisms improve existing results even for the fundamental problem of delayed MAB.
Even with these algorithmic mechanisms, proving our regret bounds requires careful analysis and new ideas that do not appear in the MAB with delays literature, as we tackle the much more complex MDP environment.

In the beginning of episode $k$, the algorithm computes an optimistic estimate $Q^j$ of $Q^{\pi^j}$ for all the episodes $j$ that their feedback just arrived.
To that end, we maintain confidence sets that contain the true transition function $p$ with high probability, and are built using all the trajectories available at the moment.
That is, for every $(s,a,h)$, we compute the empirical transition function $\bar p^k_h(s' \mid s,a)$ and define the confidence set $\mathcal{P}^k_h(s,a)$ as the set of transition functions $p'_h(\cdot \mid s,a)$ such that, for every $s' \in \Scal$,
\[
    | p'_h(s' \mid s,a) - \bar p^k_h(s' \mid s,a) |
    \le 
    \epsilon^k_h(s' \mid s,a),
\]
where $\epsilon^k_h(s' | s,a) = \wt \Theta ( \sqrt{\nicefrac{\bar p^k_h(s' | s,a)}{n^k_h(s,a)}} + \nicefrac{1}{n^k_h(s,a)} )$ is the confidence set radius.
Then, the confidence set for episode $k$ is defined by $\mathcal{P}^k = \{ \mathcal{P}^k_h(s,a) \}_{s,a,h}$.
Under bandit feedback, the computation of $Q^j$ also requires estimating the cost function $c^j$ in state-action pairs that were not visited in that episode.
For building these estimates, we utilize optimistic importance-sampling estimators \citep{jin2019learning} that first optimistically estimate the probability to visit each state $s$ in each time $h$ of episode $j$ by $u^j_h(s) = \max_{p' \in \mathcal{P}^j} \Pr[s_h=s \mid s_1=s_\text{init},\pi^j,p']$ and then set the estimator to be
$
    \hat c^j_h(s,a)
    =
    \frac{c^j_h(s,a) \mathbb{I} \{ s_h^j=s,a_h^j=a \}}{u_h^j(s) \pi^j_h(a \mid s) + \gamma}
$ with an exploration parameter $\gamma > 0$.

After the optimistic $Q$-functions are computed, we use them to improve the policy via a softmax update, i.e., we update
$
    \pi^{k+1}_h(a \mid s) \propto \pi^k_h(a \mid s) \exp (-\eta \sum_{j\in\mathcal{F}^{k}} Q_{h}^{j}(s,a))
$ for learning rate $\eta > 0$.
This update form, which may be characterized as an online mirror descent \citep{beck2003mirror} step with KL-regularization, stands in the heart of the following regret analysis (full proofs in \cref{appendix:proof-basic-regret}).
We note that \cref{thm:regret-bound-unknown-p-non-delayed-traj} handles only delayed feedback regarding the costs, while assuming that feedback regarding the learner's trajectory arrives without delay.

\begin{theorem}
    \label{thm:regret-bound-unknown-p-non-delayed-traj}
    Running Delayed OPPO with delayed cost feedback and non-delayed trajectory feedback guarantees, with probability $1 - \delta$, under full-information feedback:
    \[
        \regret
        =
        \wt O (H^{3/2} S \sqrt{A K} + H^2 \sqrt{D}),
    \]
    and under bandit feedback:
    \[
        \regret
        =
        \wt O (H S \sqrt{A} K^{2/3} + H^2 D^{2/3} + H^2 d_{max}).
    \]
\end{theorem}

\begin{proofsketch}
    With standard regret decomposition (based on the value difference lemma), we can show that the regret scales with two main terms: $(A) = \sum_k V_{1}^{\pi^{k}}(s_1^k) - V_{1}^{k}(s_1^k)$ is the bias between the estimated and true value of $\pi^k$; and $(B) = \sum_{k,h} \E^\pi [ \langle Q_{h}^{k}(s_h^k,\cdot),\pi_{h}^{k}(\cdot \mid s^k_{h}) - \pi_{h}(\cdot\mid s^k_{h}) \rangle ]$ which, for a fixed $(s,h) \in \mathcal{S} \times [H]$, can be viewed as the regret of a delayed MAB algorithm with full-information feedback, where the losses are the estimated $Q$-functions.
    
    Since the trajectories are not observed in delay, we can bound term (A) similarly to \citet{efroni2020optimistic} using our confidence sets that shrink over time.
    To bound term (B), we fix $(s,h)$ and follow a ``cheating'' algorithm technique \citep{gyorgy2020adapting}.
    To that end, we define the ``cheating'' algorithm that does not experience delay and sees one step into the future, i.e., in episode $k$ it plays policy $\bar \pi^{k+1}_h(a | s) \propto e^{-\eta \sum_{j=1}^{k} Q^j_h(s,a)}$.
    Then, we can break term (B) into two terms:
    (i) The regret of the ``cheating'' algorithm which is bounded by $\frac{\log A}{\eta}$ using a Be-The-Leader argument (see, e.g., \citet{joulani2020modular}), and (ii) The difference between $\bar \pi^{k+1}$ and $\pi^k$ which we can bound by looking at the exponential weights update form.
    Specifically, we bound the ratio $\nicefrac{\bar \pi^{k+1}_h(a|s)}{\pi^{k}_h(a|s)}$ from below by $1 - \eta \sum_{j \le k, j+d^j \ge k} Q^j_h(s,a)$ and this bounds term (ii) in terms of the missing feedback, i.e.,
    \begin{align*}
        (ii)
        & =
        \sum_{k=1}^K \sum_{a \in \mathcal{A}} Q_{h}^{k}(s,a) \bigl( \pi_{h}^{k}(a \mid s) - \bar \pi^{k+1}_{h}(a \mid s) \bigr)
        \\
        & =
        \sum_{k=1}^K \sum_{a \in \mathcal{A}} \pi_{h}^{k}(a | s) Q_{h}^{k}(s,a) \Bigl( 1 - \frac{\bar \pi^{k+1}_{h}(a \mid s)}{\pi_{h}^{k}(a \mid s)} \Bigr)
        \\
        & \le
        \eta \sum_{k=1}^K \sum_{a \in \mathcal{A}} \pi_{h}^{k}(a \mid s) Q_{h}^{k}(s,a) \sum_{j \le k, j+d^j \ge k} Q^j_h(s,a).
    \end{align*}
    Under full-information feedback, our estimates of the $Q$-function are always bounded by $H$, which leads to
    \begin{align*}
        (ii)
        & \le
        \eta H^2 \sum_{k=1}^K \sum_{j=1}^K \mathbb{I} \{ j \le k , j+d^j \ge k \}
        \\
        & =
        \eta H^2 \sum_{j=1}^K \sum_{k=1}^K \mathbb{I} \{ j \le k \le j+d^j \}
        \\
        & \le
        \eta H^2 \sum_{j=1}^K ( 1 + d^j) 
        = 
        \eta H^2 (K + D).
    \end{align*}
    To finish the proof we set $\eta = \nicefrac{1}{H \sqrt{K + D}}$.
    Under bandit feedback, this argument  becomes a lot more delicate because the $Q$-function estimates are naively bounded only by $\nicefrac{H}{\gamma}$.
    Thus, we need to prove concentration of $\sum_k V_h^k(s)$ around $\sum_k V_h^{\pi^k}(s)$ (which is indeed bounded by $HK$).
\end{proofsketch}

Notice that the regret bound in \cref{thm:regret-bound-unknown-p-non-delayed-traj} overcomes the major problems that we had with the black-box reduction approach.
Namely, the regret scales with the total delay $D$ and not the worst-case delay $K d_{max}$ (the extra additive dependence in $d_{max}$ is avoided altogether in \cref{sec:skipping-and-doubling}), and $D$ is not multiplied by neither $S$ nor $A$.
Finally, as a direct corollary of \cref{thm:regret-bound-unknown-p-non-delayed-traj}, we deduce the regret bound for the known transitions case, in which term (A) does not appear (at least under full-information feedback).
Notice that with known transitions, there is no need to handle delays in the trajectory feedback since dynamics are known.

\begin{theorem}
    \label{thm:regret-bound-known-p}
    Running Delayed OPPO with known transition function guarantees, with probability $1 - \delta$, under full-information feedback:
    $
        \regret
        =
        \wt O (H^2 \sqrt{K + D})
    $, and bandit feedback:
    $
        \regret
        =
        \wt O (H S \sqrt{A} K^{2/3} + H^2 D^{2/3} + H^2 d_{max}).
    $
\end{theorem}

\subsection{Handling delayed trajectories}
\label{sec:traj-delay}

Previously, we analyzed the Delayed OPPO algorithm in the setting where only cost is observed in delay.
In this section, we face the \emph{delayed trajectory feedback} setting in which the trajectory of episode $k$ is observed only in the end of episode $k+d^k$ together with the cost.
We emphasize that, while the trajectory from episode $k$ is observed in delay, the policy $\pi^k$ is executed regularly (see discussion in \cref{sec:related_work}).
Delayed trajectory feedback is a unique challenge in MDPs that does not arise in MAB, as no underlying dynamics exist.
Next, we provide the first analysis for delays of this kind and present novel ideas which are crucial for obtaining optimal bounds.
Some of the ideas in this section are applicable in other regimes and allow, for example, to enhance the famous UCB algorithm for stochastic MAB to be optimal in the presence of delays (see discussion in \cref{sec:discussion}).
To convey the main ideas, we focus on full-information feedback (for bandit see \cref{appendix:proof-basic-regret}).

The most natural approach to handle delayed trajectory feedback is simply to update the confidence sets once data becomes available, and then investigate the stochastic process describing the way that the confidence sets shrink over time (with the effect of the delays).
With a naive analysis of this approach, we can bound term (A) from the proof of \cref{thm:regret-bound-unknown-p-non-delayed-traj} by $H^2 S \sqrt{A (K + D)}$.
However, as discussed before, this is far from optimal since the total delay should not scale with the number of states and actions.

To get a tighter regret bound, we must understand the new challenges that cause this sub-optimality.
The main issue here is \emph{wasted exploration} due to unobserved feedback.
To tackle this issue, we leverage the following key observation: the importance of unobserved exploration becomes less significant as time progresses, since our understanding of the underlying dynamics is already substantial.
With this in mind we propose a new technique to analyze term (A): isolate the first $d_{max}$ visits to each state-action pair, and for other visits use the fact that some knowledge of the transition function is already evident.
With this technique we are able to get the improved bound $(A) \lesssim H^2 S \sqrt{A K} + H^2 S A d_{max}$.

This is a major improvement especially whenever $d_{max} < \sqrt{D}$, and even if not, the second term can be always substituted for $H^{3/2}\sqrt{SAD}$ using the skipping scheme in \cref{sec:skipping-and-doubling}.
However, we still see the undesirable multiplicative relation with $S$ and $A$.
To tighten the regret bound even further we propose a novel algorithmic mechanism to specifically direct wasted exploration.
The mechanism, that we call \emph{explicit exploration}, forces the agent to explore until it observes sufficient amount of feedback.
Specifically, until we observe feedback for $2 d_{max} \log \frac{H S A}{\delta}$ visits to state $s$ at time $h$, we pick actions uniformly at random in this state.
The explicit exploration mechanism directly improves the bound on term (A) by a factor of $A$ (as shown in the following theorem), and is in fact a necessary mechanism for optimistic algorithms in the presence of delays (see \cref{sec:discussion}).

\begin{theorem}
    \label{thm:regret-bound-unknown-p-delayed-traj}
    Running Delayed OPPO with explicit exploration, with delayed cost feedback and delayed trajectory feedback guarantees, with probability $1 - \delta$, under full-information feedback:
    \[
        \regret
        =
        \wt O (H^2 S \sqrt{A K} + H^2 \sqrt{D} + H^2 S d_{max}),
    \]
    and under bandit feedback:
    \[
        \regret
        =
        \wt O (H S \sqrt{A} K^{2/3}
        + H^2 D^{2/3} + H^2 S A d_{max}).
    \]
\end{theorem}

\begin{proofsketch}
    We start by isolating episodes in which we visit some state for which we observed less than $2 d_{max} \log \frac{H S A}{\delta}$ visits.
    Since $d_{max}$ is the maximal delay, there are only $\wt O (H S d_{max})$ such episodes (and the cost in each episode is at most $H$).
    For the rest of the episodes, by virtue of explicit exploration, we now have that the number of observed visits to each $(s,a,h)$ is at least $d_{max} / A$.
    
    Term (A) that measures the $Q$-functions estimation error is controlled by the rate at which the confidence sets shrink. 
    Let $\mathbb{I}_h^k(s,a) = \mathbb{I} \{s_h^k=s,a_h^k=a \}$, we can bound it as follows with standard analysis,
    \begin{align}
        \label{eq:estimation-error-term}
        (A)
        \lesssim
        H \sqrt{S} \sum_{s\in S} \sum_{a \in A} \sum_{h=1}^H \sum_k \frac{\mathbb{I}_h^k (s,a) }{\sqrt{n_h^{k} (s,a)}}.
    \end{align}
    Now we address the delays.
    Fix $(s,a,h)$ and denote the number of unobserved visits by 
    $N^k_h(s,a) = (m^k_h(s,a) - n^{k}_h(s,a))$.
    Next, we decouple the statistical estimation error and the effect of the delays in the following way,
    \begin{align}
        \label{eq:for-better-regret-with-stochastic-delay}
        \sum_k & \frac{\mathbb{I}_h^k(s,a) }{\sqrt{n_h^{k} (s,a)}}
        =
        \sum_k \frac{\mathbb{I}_h^k(s,a) }{\sqrt{m_h^k(s,a)}} \sqrt{\frac{m_h^k(s,a)}{n_h^{k}(s,a)}}
        \\
        \nonumber
        & \le
        \sum_k \frac{\mathbb{I}_h^k(s,a) }{\sqrt{m_h^k(s,a)}} \sqrt{1 + \frac{N_h^k(s,a)}{n_h^{k} (s,a)}}
        \\
        \nonumber
        & \le
        \sum_k \frac{\mathbb{I}_h^k(s,a) }{\sqrt{m_h^k(s,a)}} + \sum_k \frac{\mathbb{I}_h^k(s,a) }{\sqrt{m_h^k(s,a)}} \sqrt{\frac{N_h^k(s,a)}{n_h^{k}(s,a)}}.
    \end{align}
    The first term is unaffected by delays and bounded by $H^2 S \sqrt{A K}$.
    For the second term, we utilize explicit exploration in the sense that $n^{k}_h(s,a) \ge d_{max} / A$.
    Combine this with the observation that $N^k_h(s,a) \le d_{max}$ (since $d_{max}$ is the maximal delay), to obtain the bound $H^2 S A \sqrt{K}$.
    Finally, to get the tight bound (i.e., eliminate the extra $\sqrt{A}$), we split the second sum into: (1) episodes with $n^{k}_h(s,a) \ge d_{max}$ where $\nicefrac{N_h^k(s,a)}{n_h^{k}(s,a)}$ is tightly bounded by $1$ (and not $A$), and (2) episodes with $n^{k}_h(s,a) < d_{max}$ in which the regret scales as $\sqrt{d_{max}}$ (which is at most $\sqrt{K}$).
\end{proofsketch}

\subsection{Large delays and unknown total delay}
\label{sec:skipping-and-doubling}

In this section we address two final issues with our Delayed OPPO algorithm. 
First, we eliminate the dependence in the maximal delay $d_{max}$ that may be as large as $K$ even when the total delay is relatively small.
Second, we avoid the need for any prior knowledge regarding the delays which is hardly ever available, making the algorithm \emph{parameter-free}.

To handle large delays, we use a \emph{skipping} technique \citep{thune2019nonstochastic}.
That is, if some feedback arrives in delay larger than $\beta$ (where $\beta > 0$ is a skipping parameter), we just ignore it.
Thus, effectively, the maximal delay experienced by the algorithm is $\beta$, but we also need to bound the number of skipped episodes.
To that end, let $\mathcal{K}_\beta$ be the set of skipped episodes and note that $D = \sum_{k=1}^K d^k \ge |\mathcal{K}_\beta| \beta$, implying that the number of skipped episodes is bounded by $|\mathcal{K}_\beta| \le D/\beta$.
In \cref{appendix:skipping} we apply the skipping technique to all the settings considered in the paper to obtain the final regret bounds in \cref{table:comparison}.
Here, we take the unknown transitions case with delayed trajectory feedback and under full-information feedback as an example.
Setting $\beta = \sqrt{\nicefrac{D}{S H}}$ yields the following bound that is independent of the maximal delay $d_{max}$:
\begin{align*}
    \regret
    & =
    \wt O ( H^2 S \sqrt{A K} + H^2 \sqrt{D} + H^2 S \beta + H D/\beta )
    \\
    & =
    \wt O ( H^2 S \sqrt{A K} + H^{3/2} \sqrt{S D} ).
\end{align*}

To address unknown number of episodes and total delay, we design a new \emph{doubling} scheme. 
Unlike \citet{bistritz2019online} that end up with a worse bound in delayed MAB due to doubling, our carefully tuned mechanism obtains the same regret bounds (as if $K$ and $D$ were known). 
Moreover, when applied to MAB, our technique confirms the conjecture of \citet{bistritz2019online} that optimal regret with unknown $K$ and $D$ is achievable using a doubling scheme (due to lack of space, we defer the details to \cref{appendix:doubling}).
Note that $K$ and $D$ are the only parameters that the algorithm requires, since the skipping scheme replaces the need to know $d_{max}$ with the parameter $\beta$ (which is tuned using $D$).
The doubling scheme manages the tuning of the algorithm's parameters $\eta,\gamma,\beta$, making it completely parameter-free and eliminating the need for any prior knowledge regarding the delays.

The doubling scheme maintains an optimistic estimate of $D$ and uses it to tune the algorithm's parameters.
Every time the estimate doubles, the algorithm is restarted with the new doubled estimate. 
This ensures that our optimistic estimate is always relatively close to the true value of $D$ and that the number of restarts is only logarithmic, allowing us to keep the same regret bounds.
The optimistic estimate of $D$ is computed as follows.
Let $M^k$ be the number of episodes with missing feedback at the end of episode $k$.
Notice that $\sum_{k=1}^K M^k \le D$ because the feedback from episode $j$ was missing in exactly $d^j$ episodes.
Thus, at the end of episode $k$ our optimistic estimate is $\sum_{j=1}^k M^j$.
So for every episode $j$ with observed feedback, its delay is estimated by exactly $d^j$, and if its feedback was not observed, then we estimate it as if feedback will be observed in the next episode.

In \cref{appendix:doubling} we give the full pseudo-code of Delayed OPPO when combined with doubling, and formally prove that our regret bounds are not damaged by doubling.

\section{Additional results and empirical evaluation}
\label{sec:discussion}

\paragraph{Lower bound.}
For episodic stochastic MDPs, the optimal minimax regret bound is $\wt \Theta (H^{3/2} \sqrt{S A K})$ \citep{azar2017minimax,jin2018q}.
As adversarial MDPs generalize the stochastic MDP model, this lower bound also applies to our setting.
The lower bound for multi-arm bandits with delays is based on a simple reduction to non-delayed MAB with full-information feedback.
Namely, we can construct a non-delayed algorithm for full-information feedback using an algorithm $\mathcal{A}$ for fixed delay $d$ by simply feeding $\mathcal{A}$ with the same cost function for $d$ consecutive rounds.
Using the known lower bound for full-information MAB, this yields a $\Omega(\sqrt{dK}) = \Omega(\sqrt{D})$ lower bound which easily translates to a $\Omega(H\sqrt{D})$ lower bound in adversarial MDPs.
Combining these two bounds gives a lower bound of $\Omega(H^{3/2} \sqrt{S A K} + H \sqrt{D})$ for all settings, except for full-information feedback with known dynamics where the lower bound is $\Omega(H \sqrt{K+D})$.
In light of this lower bound, we discuss the regret of Delayed OPPO and open problems.

For bandit feedback, our $(K+D)^{2/3}$ regret bounds are still far from the lower bound.
However, it is important to emphasize that we cannot expect more from PO methods. 
Our bounds match state-of-the-art regret bounds for policy optimization under bandit feedback \citep{efroni2020optimistic}.
It is an open problem whether PO methods can obtain $\sqrt{K}$ regret in adversarial MDPs under bandit feedback (even with known dynamics).
Currently, the only algorithm with $\sqrt{K}$ regret for this setting is O-REPS \citep{jin2019learning}. 
It remains an important and interesting open problem to extend it to delayed feedback in the bandit case (see next paragraph).

Under full-information feedback, our regret bounds match the lower bound up to a factor of $\sqrt{S}$ (there is also sub-optimal dependence in $H$ but it can be avoided with Delayed O-REPS as discussed in the next paragraph).
However, this extra $\sqrt{S}$ factor already appears in the regret bounds for adversarial MDPs without delays \citep{rosenberg2019full,jin2019learning}. 
Determining the correct dependence in $S$ for adversarial MDPs is an important open problem that must be solved without delays first.
We note that if only cost feedback is delayed (and not trajectory feedback), then the delays are not entangled in the estimation of the transition function, and therefore the $\sqrt{D}$ term in our regret is optimal!

Another important note: even with delayed trajectory feedback, our $\sqrt{D}$ term is still optimal for a wide class of delays -- \emph{monotonic delays}. 
That is, if the sequence of delays is monotonic, i.e., $d^j \le d^k$ for $j < k$, then the $\sqrt{D}$ term of our regret bound for delayed trajectory feedback is not multiplied by $\sqrt{S}$.
This follows because in this case term $(A)$ that handles estimation error of $p$ can be analysed with respect to the \textit{actual} number of visitation,
since by the time we estimate $Q^k$ at the end of episode $k+d^k$ we already have all the feedback for $j < k$.
Monotonic delays include the fundamental setting of a fixed delay $d$.

\paragraph{O-REPS vs OPPO.}
PO methods directly optimize the policy. 
Practically, this translates to estimating the $Q$-function and then applying a closed-form update to the policy in each state.
Alternatively, O-REPS methods \citep{zimin2013online} optimize over the state-action occupancy measures instead
of directly on policies.
This requires solving a global convex optimization problem of size $H S^2 A$ \citep{rosenberg2019full} in the beginning of each episode, which has no closed-form solution and is extremely inefficient computationally.
Another significant shortcoming of O-REPS is the difficulty to scale it up to function approximation, since the constrained optimization problem becomes non-convex.
On the other hand, PO methods extend naturally to function approximation and enjoy great empirical success (e.g., \citet{haarnoja2018soft}).

Other than their practical merits, this paper reveals an important theoretical advantage of PO methods over O-REPS -- simple update form.
We utilize the exponential weights update form of Delayed OPPO in order to investigate the propagation of delayed feedback through the episodes.
This results in an intuitive analysis that achieves the best available PO regret bounds even when feedback is delayed.
On the other hand, there is very limited understanding regarding the solution for the O-REPS optimization problem, making it very hard to extend beyond its current scope.
Specifically, studying the effect of delays on this optimization problem is extremely challenging and takes involved analysis.
While we were able to analyze Delayed O-REPS under full-information feedback (\cref{appendix:delayed-o-reps}) and give tight regret bounds (\cref{thm:regret-delayed-o-reps-main-paper}), we were not able to extend our analysis to bandit feedback because it involves a complicated in-depth investigation of the difference between any two consecutive occupancy measures chosen by the algorithm.
Our analysis bounds this difference under full-information feedback, but in order to bound the regret under bandit feedback its ratio (and the high variance of importance-sampling estimators) must also be bounded.
Extending Delayed O-REPS to bandit feedback remains an important open problem, for which our analysis lays the foundations, and is currently the only way that can achieve $\sqrt{K}$ regret in the presence of delays.

\begin{theorem}
    \label{thm:regret-delayed-o-reps-main-paper}
    Running Delayed O-REPS under full-information feedback guarantees, with probability $1 - \delta$, with known transitions: $\regret = \wt O (H \sqrt{K + D})$, and with unknown dynamics, delayed cost feedback and non-delayed trajectory feedback: $\regret = \wt O (H^{3/2} S \sqrt{A K} + H \sqrt{D})$.
\end{theorem}

\paragraph{Stochastic MDP with delayed feedback.}
Most of the RL literature has focused on stochastic MDPs -- a special case of adversarial MDPs where cost $c^k_h(s,a)$ of episode $k$ is sampled i.i.d from a fixed distribution $C_h(s,a)$.
Thus, studying the effects of delayed feedback on stochastic MDPs is a natural question.
With stochastic costs, OPPO obtains $\sqrt{K}$ regret even under bandit feedback, since we can replace importance-sampling estimators with an empirical average.
This means that with stochastic costs and bandit feedback, our Delayed OPPO algorithm obtains the same near-optimal regret bounds as under full-information feedback.
However, the $\sqrt{D}$ lower bound heavily relies on adversarial costs, as it uses a sequence of costs that change every $d$ episodes, suggesting that $\sqrt{D}$ dependence might not be necessary.

Indeed, for stochastic cost, delayed versions of optimistic algorithms (e.g., \citet{zanette2018problem}) have regret scaling as the estimation error (term (A) in \cref{eq:estimation-error-term}), which means that our analysis (\cref{sec:traj-delay}) proves regret that does not scale with $\sqrt{D}$ but only with $H^2 S A d_{max}$.
Again, this can be improved to $H^2 S d_{max}$ using explicit exploration. 

\begin{theorem}                 
    \label{thm:regret-bound-stochastic}
    Running an optimistic algorithm with explicit exploration, with delayed bandit cost feedback and delayed trajectory feedback guarantees, with probability $1 - \delta$, regret bound of
    $
        \wt O (H^2 S \sqrt{A K} + H^2 S d_{max})
    $ in stochastic MDPs.
\end{theorem}

This contribution is important, even for the rich literature on delayed stochastic MAB. 
\citet{lancewicki2021stochastic} show that the (optimistic) UCB algorithm may suffer sub-optimal regret of $A d_{max}$. Furthermore, they were able to remove the $A$ factor by an action-elimination algorithm which explores active arms equally. 
Since optimism is currently the only approach for handling unknown transitions in adversarial MDPs, it was crucial for us to find a novel solution to handle delays in optimistic algorithms. \cref{thm:regret-bound-stochastic} shows that optimistic algorithms (like UCB) can indeed be ``fixed'' to handle delays optimally, using explicit exploration.

\paragraph{Empirical evaluation.}
\begin{figure}[t]
    \centering 
    \includegraphics[width=0.42\textwidth,height=55pt]{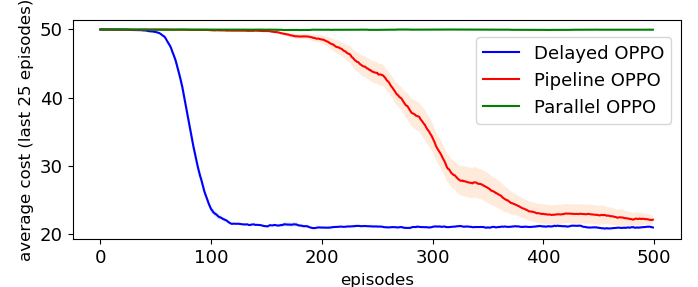}
  \caption{Average cost of delayed algorithms in grid world with geometrically distributed delays.}
  \vspace{-10pt}
  \label{fig:dmax_main}
\end{figure}
We used synthetic experiments to compare the performance of \emph{Delayed OPPO} to two other generic approaches for handling delays: \emph{Parallel-OPPO} -- running in parallel $d_{max}$ online algorithms, as described in \cref{sec:black-box-red}, and \emph{Pipeline-OPPO} -- another simple approach for turning a non-delayed algorithm to an algorithm that handles delays by simply waiting for the first $d_{max}$ episodes and then feeding the feedback always with delay $d_{max}$.
We used a simple $10 \times 10$ grid world (with $H = 50,K = 500$) where the agent starts in one corner and needs to reach the opposite corner, which is the goal state. 
The cost is $1$ in all states except for $0$ cost in the goal state.
Delays are drawn i.i.d from a geometric distribution with mean $10$, and
the maximum delay $d_{max}$ is computed on the sequence of realized delays (it is roughly $10 \log K\approx 60$).

\cref{fig:dmax_main} shows Delayed OPPO significantly outperforms the other approaches, thus highlighting the importance of handling variable delays and not simply considering the worst-case delay $d_{max}$.
An important note is that, apart from its very high cost, Parallel-OPPO also requires much more memory (factor $d_{max}$ more).
For more implementation details and additional experiments, see \cref{appendix:experiments}.

\section*{Acknowledgements}

This project has received funding from the European Research Council (ERC) under the European Union’s Horizon 2020 research and innovation program (grant agreement No. 882396), by the Israel Science Foundation (grant number 993/17), Tel Aviv University Center for AI and Data Science (TAD), and the Yandex Initiative for Machine Learning at Tel Aviv University.

\bibliography{aaai22}

\begin{thebibliography}{51}
\providecommand{\natexlab}[1]{#1}

\bibitem[{Azar, Osband, and Munos(2017)}]{azar2017minimax}
Azar, M.~G.; Osband, I.; and Munos, R. 2017.
\newblock Minimax regret bounds for reinforcement learning.
\newblock In \emph{Proceedings of the 34th International Conference on Machine
  Learning-Volume 70}, 263--272. JMLR. org.

\bibitem[{Beck and Teboulle(2003)}]{beck2003mirror}
Beck, A.; and Teboulle, M. 2003.
\newblock Mirror descent and nonlinear projected subgradient methods for convex
  optimization.
\newblock \emph{Operations Research Letters}, 31(3): 167--175.

\bibitem[{Bistritz et~al.(2019)Bistritz, Zhou, Chen, Bambos, and
  Blanchet}]{bistritz2019online}
Bistritz, I.; Zhou, Z.; Chen, X.; Bambos, N.; and Blanchet, J. 2019.
\newblock Online exp3 learning in adversarial bandits with delayed feedback.
\newblock In \emph{Advances in Neural Information Processing Systems},
  11349--11358.

\bibitem[{Cai et~al.(2020)Cai, Yang, Jin, and Wang}]{cai2019provably}
Cai, Q.; Yang, Z.; Jin, C.; and Wang, Z. 2020.
\newblock Provably efficient exploration in policy optimization.
\newblock In \emph{International Conference on Machine Learning}, 1283--1294.
  PMLR.

\bibitem[{Cesa-Bianchi, Gentile, and Mansour(2018)}]{cesa2018nonstochastic}
Cesa-Bianchi, N.; Gentile, C.; and Mansour, Y. 2018.
\newblock Nonstochastic bandits with composite anonymous feedback.
\newblock In \emph{Conference On Learning Theory}, 750--773.

\bibitem[{Cesa-Bianchi et~al.(2016)Cesa-Bianchi, Gentile, Mansour, and
  Minora}]{cesa2016delay}
Cesa-Bianchi, N.; Gentile, C.; Mansour, Y.; and Minora, A. 2016.
\newblock Delay and cooperation in nonstochastic bandits.
\newblock In \emph{Conference on Learning Theory}, 605--622.

\bibitem[{Changuel, Sayadi, and Kieffer(2012)}]{changuel2012online}
Changuel, N.; Sayadi, B.; and Kieffer, M. 2012.
\newblock Online learning for QoE-based video streaming to mobile receivers.
\newblock In \emph{2012 IEEE Globecom Workshops}, 1319--1324. IEEE.

\bibitem[{Chen et~al.(2020)Chen, Xu, Liu, Li, and Zhao}]{chen2020delay}
Chen, B.; Xu, M.; Liu, Z.; Li, L.; and Zhao, D. 2020.
\newblock Delay-aware multi-agent reinforcement learning.
\newblock \emph{arXiv preprint arXiv:2005.05441}.

\bibitem[{Even-Dar, Kakade, and Mansour(2009)}]{even2009online}
Even-Dar, E.; Kakade, S.~M.; and Mansour, Y. 2009.
\newblock Online Markov decision processes.
\newblock \emph{Mathematics of Operations Research}, 34(3): 726--736.

\bibitem[{Gael et~al.(2020)Gael, Vernade, Carpentier, and
  Valko}]{manegueu2020stochastic}
Gael, M.~A.; Vernade, C.; Carpentier, A.; and Valko, M. 2020.
\newblock Stochastic bandits with arm-dependent delays.
\newblock In \emph{International Conference on Machine Learning}, 3348--3356.
  PMLR.

\bibitem[{Gy{\"o}rgy and Joulani(2020)}]{gyorgy2020adapting}
Gy{\"o}rgy, A.; and Joulani, P. 2020.
\newblock Adapting to Delays and Data in Adversarial Multi-Armed Bandits.
\newblock \emph{arXiv preprint arXiv:2010.06022}.

\bibitem[{Haarnoja et~al.(2018)Haarnoja, Zhou, Abbeel, and
  Levine}]{haarnoja2018soft}
Haarnoja, T.; Zhou, A.; Abbeel, P.; and Levine, S. 2018.
\newblock Soft Actor-Critic: Off-Policy Maximum Entropy Deep Reinforcement
  Learning with a Stochastic Actor.
\newblock In \emph{International Conference on Machine Learning}, 1861--1870.

\bibitem[{Hazan(2019)}]{hazan2019introduction}
Hazan, E. 2019.
\newblock Introduction to online convex optimization.
\newblock \emph{arXiv preprint arXiv:1909.05207}.

\bibitem[{He, Zhou, and Gu(2021)}]{he2021nearly}
He, J.; Zhou, D.; and Gu, Q. 2021.
\newblock Nearly Optimal Regret for Learning Adversarial MDPs with Linear
  Function Approximation.
\newblock \emph{arXiv preprint arXiv:2102.08940}.

\bibitem[{Jaksch, Ortner, and Auer(2010)}]{jaksch2010near}
Jaksch, T.; Ortner, R.; and Auer, P. 2010.
\newblock Near-optimal Regret Bounds for Reinforcement Learning.
\newblock \emph{Journal of Machine Learning Research}, 11(4).

\bibitem[{Jin et~al.(2018)Jin, Allen-Zhu, Bubeck, and Jordan}]{jin2018q}
Jin, C.; Allen-Zhu, Z.; Bubeck, S.; and Jordan, M.~I. 2018.
\newblock Is q-learning provably efficient?
\newblock In \emph{Advances in Neural Information Processing Systems},
  4863--4873.

\bibitem[{Jin et~al.(2020{\natexlab{a}})Jin, Jin, Luo, Sra, and
  Yu}]{jin2019learning}
Jin, C.; Jin, T.; Luo, H.; Sra, S.; and Yu, T. 2020{\natexlab{a}}.
\newblock Learning adversarial markov decision processes with bandit feedback
  and unknown transition.
\newblock In \emph{International Conference on Machine Learning}, 4860--4869.
  PMLR.

\bibitem[{Jin et~al.(2020{\natexlab{b}})Jin, Yang, Wang, and
  Jordan}]{jin2020provably}
Jin, C.; Yang, Z.; Wang, Z.; and Jordan, M.~I. 2020{\natexlab{b}}.
\newblock Provably efficient reinforcement learning with linear function
  approximation.
\newblock In \emph{Conference on Learning Theory}, 2137--2143.

\bibitem[{Joulani, Gyorgy, and Szepesv{\'a}ri(2013)}]{joulani2013online}
Joulani, P.; Gyorgy, A.; and Szepesv{\'a}ri, C. 2013.
\newblock Online learning under delayed feedback.
\newblock In \emph{International Conference on Machine Learning}, 1453--1461.

\bibitem[{Joulani, Gy{\"o}rgy, and Szepesv{\'a}ri(2020)}]{joulani2020modular}
Joulani, P.; Gy{\"o}rgy, A.; and Szepesv{\'a}ri, C. 2020.
\newblock A modular analysis of adaptive (non-) convex optimization: Optimism,
  composite objectives, variance reduction, and variational bounds.
\newblock \emph{Theoretical Computer Science}, 808: 108--138.

\bibitem[{Katsikopoulos and Engelbrecht(2003)}]{katsikopoulos2003markov}
Katsikopoulos, K.~V.; and Engelbrecht, S.~E. 2003.
\newblock Markov decision processes with delays and asynchronous cost
  collection.
\newblock \emph{IEEE transactions on automatic control}, 48(4): 568--574.

\bibitem[{Lancewicki et~al.(2021)Lancewicki, Segal, Koren, and
  Mansour}]{lancewicki2021stochastic}
Lancewicki, T.; Segal, S.; Koren, T.; and Mansour, Y. 2021.
\newblock Stochastic Multi-Armed Bandits with Unrestricted Delay Distributions.
\newblock In \emph{Proceedings of the 38th International Conference on Machine
  Learning, {ICML} 2021, 18-24 July 2021, Virtual Event}, 5969--5978. {PMLR}.

\bibitem[{Liu, Wang, and Liu(2014)}]{liu2014impact}
Liu, S.; Wang, X.; and Liu, P.~X. 2014.
\newblock Impact of communication delays on secondary frequency control in an
  islanded microgrid.
\newblock \emph{IEEE Transactions on Industrial Electronics}, 62(4):
  2021--2031.

\bibitem[{Mahmood et~al.(2018)Mahmood, Korenkevych, Komer, and
  Bergstra}]{mahmood2018setting}
Mahmood, A.~R.; Korenkevych, D.; Komer, B.~J.; and Bergstra, J. 2018.
\newblock Setting up a reinforcement learning task with a real-world robot.
\newblock In \emph{2018 IEEE/RSJ International Conference on Intelligent Robots
  and Systems (IROS)}, 4635--4640. IEEE.

\bibitem[{Maurer and Pontil(2009)}]{maurer2009empirical}
Maurer, A.; and Pontil, M. 2009.
\newblock Empirical Bernstein Bounds and Sample Variance Penalization.
\newblock \emph{stat}, 1050: 21.

\bibitem[{Neu, Gy{\"{o}}rgy, and Szepesv{\'{a}}ri(2010)}]{neu2010ossp}
Neu, G.; Gy{\"{o}}rgy, A.; and Szepesv{\'{a}}ri, C. 2010.
\newblock The Online Loop-free Stochastic Shortest-Path Problem.
\newblock In \emph{Conference on Learning Theory {(COLT)}}, 231--243.

\bibitem[{Neu, Gy{\"{o}}rgy, and Szepesv{\'{a}}ri(2012)}]{neu2012unknown}
Neu, G.; Gy{\"{o}}rgy, A.; and Szepesv{\'{a}}ri, C. 2012.
\newblock The adversarial stochastic shortest path problem with unknown
  transition probabilities.
\newblock In \emph{Proceedings of the Fifteenth International Conference on
  Artificial Intelligence and Statistics, {(AISTATS)}}, 805--813.

\bibitem[{Neu et~al.(2014)Neu, Gy{\"{o}}rgy, Szepesv{\'{a}}ri, and
  Antos}]{neu2014bandit}
Neu, G.; Gy{\"{o}}rgy, A.; Szepesv{\'{a}}ri, C.; and Antos, A. 2014.
\newblock Online {Markov Decision Processes} Under Bandit Feedback.
\newblock \emph{{IEEE} Trans. Automat. Contr.}, 59(3): 676--691.

\bibitem[{Pike-Burke et~al.(2018)Pike-Burke, Agrawal, Szepesvari, and
  Grunewalder}]{pike2018bandits}
Pike-Burke, C.; Agrawal, S.; Szepesvari, C.; and Grunewalder, S. 2018.
\newblock Bandits with delayed, aggregated anonymous feedback.
\newblock In \emph{International Conference on Machine Learning}, 4105--4113.
  PMLR.

\bibitem[{Quanrud and Khashabi(2015)}]{quanrud2015online}
Quanrud, K.; and Khashabi, D. 2015.
\newblock Online learning with adversarial delays.
\newblock \emph{Advances in neural information processing systems}, 28:
  1270--1278.

\bibitem[{Rosenberg et~al.(2020)Rosenberg, Cohen, Mansour, and
  Kaplan}]{cohen2020ssp}
Rosenberg, A.; Cohen, A.; Mansour, Y.; and Kaplan, H. 2020.
\newblock Near-optimal Regret Bounds for Stochastic Shortest Path.
\newblock In \emph{International Conference on Machine Learning}, 8210--8219.
  PMLR.

\bibitem[{Rosenberg and Mansour(2019{\natexlab{a}})}]{rosenberg2019full}
Rosenberg, A.; and Mansour, Y. 2019{\natexlab{a}}.
\newblock Online Convex Optimization in Adversarial Markov Decision Processes.
\newblock In \emph{International Conference on Machine Learning}, 5478--5486.

\bibitem[{Rosenberg and Mansour(2019{\natexlab{b}})}]{rosenberg2019bandit}
Rosenberg, A.; and Mansour, Y. 2019{\natexlab{b}}.
\newblock Online Stochastic Shortest Path with Bandit Feedback and Unknown
  Transition Function.
\newblock In \emph{Advances in Neural Information Processing Systems},
  2209--2218.

\bibitem[{Schuitema et~al.(2010)Schuitema, Bu{\c{s}}oniu, Babu{\v{s}}ka, and
  Jonker}]{schuitema2010control}
Schuitema, E.; Bu{\c{s}}oniu, L.; Babu{\v{s}}ka, R.; and Jonker, P. 2010.
\newblock Control delay in reinforcement learning for real-time dynamic
  systems: a memoryless approach.
\newblock In \emph{2010 IEEE/RSJ International Conference on Intelligent Robots
  and Systems}, 3226--3231. IEEE.

\bibitem[{Schulman et~al.(2015)Schulman, Levine, Abbeel, Jordan, and
  Moritz}]{schulman2015trust}
Schulman, J.; Levine, S.; Abbeel, P.; Jordan, M.; and Moritz, P. 2015.
\newblock Trust region policy optimization.
\newblock In \emph{International conference on machine learning}, 1889--1897.

\bibitem[{Schulman et~al.(2017)Schulman, Wolski, Dhariwal, Radford, and
  Klimov}]{schulman2017proximal}
Schulman, J.; Wolski, F.; Dhariwal, P.; Radford, A.; and Klimov, O. 2017.
\newblock Proximal policy optimization algorithms.
\newblock \emph{arXiv preprint arXiv:1707.06347}.

\bibitem[{Shani et~al.(2020)Shani, Efroni, Rosenberg, and
  Mannor}]{efroni2020optimistic}
Shani, L.; Efroni, Y.; Rosenberg, A.; and Mannor, S. 2020.
\newblock Optimistic Policy Optimization with Bandit Feedback.
\newblock In \emph{International Conference on Machine Learning}, 8604--8613.
  PMLR.

\bibitem[{Sutton and Barto(2018)}]{sutton2018reinforcement}
Sutton, R.~S.; and Barto, A.~G. 2018.
\newblock \emph{Reinforcement learning: An introduction}.
\newblock MIT press.

\bibitem[{Thune, Cesa-Bianchi, and Seldin(2019)}]{thune2019nonstochastic}
Thune, T.~S.; Cesa-Bianchi, N.; and Seldin, Y. 2019.
\newblock Nonstochastic multiarmed bandits with unrestricted delays.
\newblock In \emph{Advances in Neural Information Processing Systems},
  6541--6550.

\bibitem[{Vernade, Capp{\'e}, and Perchet(2017)}]{vernade2017stochastic}
Vernade, C.; Capp{\'e}, O.; and Perchet, V. 2017.
\newblock Stochastic Bandit Models for Delayed Conversions.
\newblock In \emph{Conference on Uncertainty in Artificial Intelligence}.

\bibitem[{Walsh et~al.(2009)Walsh, Nouri, Li, and Littman}]{walsh2009learning}
Walsh, T.~J.; Nouri, A.; Li, L.; and Littman, M.~L. 2009.
\newblock Learning and planning in environments with delayed feedback.
\newblock \emph{Autonomous Agents and Multi-Agent Systems}, 18(1): 83.

\bibitem[{Weinberger and Ordentlich(2002)}]{weinberger2002delayed}
Weinberger, M.~J.; and Ordentlich, E. 2002.
\newblock On delayed prediction of individual sequences.
\newblock \emph{IEEE Transactions on Information Theory}, 48(7): 1959--1976.

\bibitem[{Yang and Wang(2019)}]{yang2019sample}
Yang, L.; and Wang, M. 2019.
\newblock Sample-optimal parametric Q-learning using linearly additive
  features.
\newblock In \emph{International Conference on Machine Learning}, 6995--7004.
  PMLR.

\bibitem[{Zanette et~al.(2020{\natexlab{a}})Zanette, Brandfonbrener, Brunskill,
  Pirotta, and Lazaric}]{zanette2020frequentist}
Zanette, A.; Brandfonbrener, D.; Brunskill, E.; Pirotta, M.; and Lazaric, A.
  2020{\natexlab{a}}.
\newblock Frequentist regret bounds for randomized least-squares value
  iteration.
\newblock In \emph{International Conference on Artificial Intelligence and
  Statistics}, 1954--1964.

\bibitem[{Zanette and Brunskill(2018)}]{zanette2018problem}
Zanette, A.; and Brunskill, E. 2018.
\newblock Problem dependent reinforcement learning bounds which can identify
  bandit structure in mdps.
\newblock In \emph{International Conference on Machine Learning}, 5747--5755.

\bibitem[{Zanette and Brunskill(2019)}]{zanette2019tighter}
Zanette, A.; and Brunskill, E. 2019.
\newblock Tighter Problem-Dependent Regret Bounds in Reinforcement Learning
  without Domain Knowledge using Value Function Bounds.
\newblock In \emph{International Conference on Machine Learning}, 7304--7312.

\bibitem[{Zanette et~al.(2020{\natexlab{b}})Zanette, Lazaric, Kochenderfer, and
  Brunskill}]{zanette2020learning}
Zanette, A.; Lazaric, A.; Kochenderfer, M.; and Brunskill, E.
  2020{\natexlab{b}}.
\newblock Learning near optimal policies with low inherent bellman error.
\newblock In \emph{International Conference on Machine Learning}, 10978--10989.
  PMLR.

\bibitem[{Zhou, Xu, and Blanchet(2019)}]{zhou2019learning}
Zhou, Z.; Xu, R.; and Blanchet, J. 2019.
\newblock Learning in generalized linear contextual bandits with stochastic
  delays.
\newblock In \emph{Advances in Neural Information Processing Systems},
  5197--5208.

\bibitem[{Zimin(2013)}]{ziminonline}
Zimin, A. 2013.
\newblock \emph{Online Learning in Markovian Decision Processes}.
\newblock Ph.D. thesis, Central European University.

\bibitem[{Zimin and Neu(2013)}]{zimin2013online}
Zimin, A.; and Neu, G. 2013.
\newblock Online learning in episodic Markovian decision processes by relative
  entropy policy search.
\newblock In \emph{Advances in Neural Information Processing Systems 26: 27th
  Annual Conference on Neural Information Processing Systems 2013. Proceedings
  of a meeting held December 5-8, 2013, Lake Tahoe, Nevada, United States},
  1583--1591.

\bibitem[{Zimmert and Seldin(2020)}]{zimmert2020optimal}
Zimmert, J.; and Seldin, Y. 2020.
\newblock An optimal algorithm for adversarial bandits with arbitrary delays.
\newblock In \emph{International Conference on Artificial Intelligence and
  Statistics}, 3285--3294. PMLR.

\end{thebibliography}

\clearpage
\onecolumn
\appendix

\section{The Delayed OPPO Algorithm}
\label{sec:full-algo}

\begin{algorithm}
    \caption{Delayed OPPO with known transition function}  
    \label{:delayed-OPPO-unknown-p}
    \begin{algorithmic}
        \STATE \textbf{Input:} State space $\mathcal{S}$, Action space $\mathcal{A}$, Horizon $H$, Transition function $p$, Learning rate $\eta > 0$, Exploration parameter $\gamma > 0$.
        
        \STATE \textbf{Initialization:} 
        Set $\pi_{h}^{1}(a \mid s) = \nicefrac{1}{A}$ for every $(s,a,h) \in \mathcal{S} \times \mathcal{A} \times [H]$.
        
        \FOR{$k=1,2,\dots,K$}
            
            \STATE Play episode $k$ with policy $\pi^k$.
            
            \STATE Observe feedback from all episodes $j \in \mathcal{F}^k$.
            
            \STATE {\color{gray} \# Policy Evaluation}
            
            \FOR{$j\in \mathcal{F}^k$}
            
                \STATE $\forall s \in \mathcal{S} : V_{H+1}^j(s)=0$.
                
                \FOR{$h = H,\dots,1$ and $(s,a) \in \mathcal{S} \times \mathcal{A}$}
                    
                    \IF{bandit feedback}
                    
                        \STATE $\hat c^j_h(s,a) = \frac{c_h^j(s,a) \cdot \mathbb{I} \{ s_h^j=s,a_h^j=a \}}{q^{p,\pi^j}_h(s) \pi^j_h(a \mid s) + \gamma}$.
                    
                    \ELSIF{full-information feedback}
                    
                        \STATE $\hat c^j_h(s,a) = c_h^j(s,a)$.
                        
                    \ENDIF
                    
                    \STATE $Q_{h}^{j}(s,a) = \hat{c}_{h}^{j}(s,a)+\langle p_{h}(\cdot\mid s,a),V_{h+1}^{j} \rangle$.
                    
                    \STATE $V_{h}^{j}(s) = \langle Q_{h}^{j}(s,\cdot),\pi_{h}^{j}(\cdot \mid s)\rangle$.
                
                \ENDFOR
            
            \ENDFOR
            
            \STATE {\color{gray} \# Policy Improvement}
            
            \FOR{$(s,a,h) \in \mathcal{S} \times \mathcal{A} \times [H]$}
            
                \STATE $\pi^{k+1}_h(a | s) = \frac{\pi^k_h(a \mid s) \exp (-\eta \sum_{j\in\mathcal{F}^{k}} Q_{h}^{j}(s,a))}{\sum_{a' \in \mathcal{A}} \pi^k_h(a' \mid s) \exp (-\eta \sum_{j\in\mathcal{F}^{k}} Q_{h}^{j}(s,a'))}$.
                
            \ENDFOR
            
        \ENDFOR
        
    \end{algorithmic}
\end{algorithm}

\begin{algorithm}
    \caption{Delayed OPPO with unknown transition function}  
    \label{alg:delayed-OPPO-unknown-p}
    \begin{algorithmic}
        \STATE \textbf{Input:} State space $\mathcal{S}$, Action space $\mathcal{A}$, Horizon $H$, Learning rate $\eta > 0$, Exploration parameter $\gamma > 0$, Confidence parameter $\delta > 0$, Explicit exploration parameter $\texttt{UseExplicitExploration} \in \{ \texttt{true},\texttt{false} \}$, maximal delay $d_{max}$.
        
        \STATE \textbf{Initialization:} 
        Set $\pi_{h}^{1}(a \mid s) = \nicefrac{1}{A}$ for every $(s,a,h) \in \mathcal{S} \times \mathcal{A} \times [H]$, $\Kexp = \emptyset$.
        
        \FOR{$k=1,2,\dots,K$}
        
            \FOR{$s \in \mathcal{S}$}
            
                \IF{$\texttt{UseExplicitExploration} = \texttt{true}$ and $n^k_h(s) \le 2 d_{max} \log \frac{HSA}{\delta}$}
                    
                    \STATE $\forall a \in \mathcal{A}: \tilde \pi^k_h(a \mid s) = \nicefrac{1}{A}$.
                
                \ELSE
                
                    \STATE $\forall a \in \mathcal{A}: \tilde \pi^k_h(a \mid s) = \pi^k_h(a \mid s)$.
                
                \ENDIF
            
            \ENDFOR
            
            \STATE Play episode $k$ with policy $\tilde \pi^k$.
            
            \IF{trajectory feedback is not delayed}
            
                \STATE Observe trajectory $U^k = \{ (s_h^k,a_h^k) \}_{h=1}^H$.
                
            \ENDIF
            
            \STATE Observe feedback from all episodes $j \in \mathcal{F}^k$ and update $n^{k+1}$ and $\bar p^{k+1}$.
            
            \STATE If explicit exploration was used in some $j \in \mathcal{F}^k$, then add $j$ to $\Kexp$.
            
            \STATE Compute confidence set $\mathcal{P}^{k} = \{\mathcal{P}^{k}_h(s,a) \}_{s,a,h}$, where $\mathcal{P}^{k}_h(s,a)$ contains all transition functions $p'_h(\cdot \mid s,a) $ such that for every $s' \in S$,
            \begin{align*}
                | p'_h(s' | s,a) - \bar p^{k}_h(s' | s,a) |
                & \le
                \epsilon_h^{k}(s' | s,a)
                \\
                & =
                4\sqrt{ \frac{\bar{p}_{h}^{k}(s'| s,a) (1-\bar{p}_{h}^{k}(s'| s,a)) \ln\frac{HSAK}{4\delta}}{n_{h}^{k}(s,a) \vee 1}} + 10 \frac{\ln\frac{HSAK}{4\delta}}{n_{h}^{k}(s,a) \vee 1}.
            \end{align*}
            
            \STATE {\color{gray} \# Policy Evaluation}
            
            \FOR{$j\in \mathcal{F}^k \setminus \Kexp$}
            
                \STATE $\forall s \in \mathcal{S} : V_{H+1}^j(s)=0$.
                
                \FOR{$h = H,\dots,1$ and $(s,a) \in \mathcal{S} \times \mathcal{A}$}
                    
                    \IF{bandit feedback}
                    
                        \STATE $u^j_h(s) = \max_{p' \in \mathcal{P}^{j}} q^{p',\pi^j}_h(s) = \max_{p' \in \mathcal{P}^{j}} \Pr[s_h=s \mid s_1=s_\text{init},\pi^j,p']$.
                    
                        \STATE $\hat c^j_h(s,a) = \frac{c_h^j(s,a) \cdot \mathbb{I} \{ s_h^j=s,a_h^j=a \}}{u_h^j(s) \pi^j_h(a \mid s) + \gamma}$.
                        
                        \STATE $\hat p_h^j(\cdot | s,a) \in \underset{p'_h(\cdot | s,a) \in \mathcal{P}_h^{j}(\cdot \mid s,a)}{\arg\min} \langle p'_h(\cdot | s,a) , V_{h+1}^j \rangle$.
                    
                    \ELSIF{full-information feedback}
                    
                        \STATE $\hat c^j_h(s,a) = c_h^j(s,a)$.
                        
                        \STATE $\hat p_h^j(\cdot | s,a) \in \underset{p'_h(\cdot | s,a) \in \mathcal{P}^{k}_h(\cdot \mid s,a)}{\arg\min} \langle p'_h(\cdot | s,a) , V_{h+1}^j \rangle$.
                        
                    \ENDIF
                    
                    \STATE $Q_{h}^{j}(s,a) = \hat{c}_{h}^{j}(s,a)+\langle \hat{p}_{h}^j(\cdot\mid s,a),V_{h+1}^{j} \rangle$.
                    
                    \STATE $V_{h}^{j}(s) = \langle Q_{h}^{j}(s,\cdot),\pi_{h}^{j}(\cdot \mid s)\rangle$.
                
                \ENDFOR
            
            \ENDFOR
            
            \STATE {\color{gray} \# Policy Improvement}
            
            \FOR{$(s,a,h) \in \mathcal{S} \times \mathcal{A} \times [H]$}
            
                \STATE $\pi^{k+1}_h(a | s) = \frac{\pi^k_h(a \mid s) \exp (-\eta \sum_{j\in\mathcal{F}^{k} \setminus \Kexp} Q_{h}^{j}(s,a))}{\sum_{a' \in \mathcal{A}} \pi^k_h(a' \mid s) \exp (-\eta \sum_{j\in\mathcal{F}^{k} \setminus \Kexp} Q_{h}^{j}(s,a'))}$.
                
            \ENDFOR
            
        \ENDFOR
        
    \end{algorithmic}
\end{algorithm}

\newpage

\section{Full proofs of main theorems} \label{appendix:proof-basic-regret}

We start by defining failure events. 
The rest of the analysis focuses on the good event: the event in which non of the failure events occur. 
We show that in order to guarantee a regret bound that holds with probability of $1 - \delta$, we only need to pay a factor which is logarithmic in $1/\delta$. 
In \cref{appendix:fail-events} we define the failure events and bound their probability in \Cref{lemma:basic-good-event,lemma:conditioned-good-event}. 

In \cref{appendix:basic-bandit-proof} we prove the regret bound for the most challenging case: unknown transition function + delayed trajectory feedback + bandit feedback, that is, \cref{thm:regret-bound-unknown-p-delayed-traj} with bandit feedback.
Then, the rest of the proofs follow easily as corollaries.

In \cref{appendix:basic-full-proof} we prove the regret for the case of unknown transition function + delayed trajectory feedback + full-information feedback, that is, \cref{thm:regret-bound-unknown-p-delayed-traj} with full-information feedback.

In \cref{appendix:no-trajectory-delay} we prove the regret for the case of unknown transition function but with non-delayed trajectory feedback, that is, \cref{thm:regret-bound-unknown-p-non-delayed-traj}.
Finally, in \cref{appendix:known-dynamics} we prove the regret for the case of known transition function, that is, \cref{thm:regret-bound-known-p}.

\begin{remark}
    The analysis for bandit feedback uses estimated $Q$-functions $Q^j$ that were computed with the confidence set $\mathcal{P}^j$ and not $\mathcal{P}^{j+d^j}$, as in the full-information case.
    This simplifies concentration arguments but ignores a lot of data.
    It also means that under bandit feedback the algorithm has worse space complexity, since we may need to keep up to $d_{max}$ empirical transition functions.
    However, the space complexity can be easily reduced by re-computing the confidence sets only when the number of visits to some state-action pair is doubled, and not in the end of every episode.
\end{remark}

\subsection{Failure events}     \label{appendix:fail-events}

Fix some probability $\delta'$. 
We now define basic failure events:

\begin{itemize}

    \item $F_{1}^{basic}
    = \left\{ \exists k,s',s,a,h:
    \left| p_{h}(s'\mid s,a)-\bar{p}_{h}^{k}(s'\mid s,a) \right|
    > \epsilon_{h}^{k}(s' \mid s,a) \right\}$  
    \begin{itemize}
        \item[$\blacktriangleright$] where $\epsilon_{h}^{k}(s' \mid s,a) = 4\sqrt{ \frac{\bar{p}_{h}^{k}(s' \mid s,a) (1-\bar{p}_{h}^{k}(s'\mid s,a)) \ln\frac{HSAK}{4\delta}}{n_{h}^{k}(s,a) \vee 1}} + 10 \frac{\ln\frac{HSAK}{4\delta}}{n_{h}^{k}(s,a) \vee 1}$.
    \end{itemize}
    
    \item $F_{2}^{basic} =
    \left\{ \exists k,s,a,h:
    \left\Vert p_{h}(\cdot \mid s,a) - \bar{p}_{h}^{k}(\cdot \mid s,a) \right\Vert_{1}
    >
    \sqrt{ \frac{14 S \ln\left( \frac{HSAK}{\delta'} \right)}{n_{h}^{k}(s,a)\vee1}} \right\} $
    
    \item $F_{3}^{basic}
    =\left\{ 
    \sum_{k,s,a,h}
    \bigl(q_{h}^{k}(s,a)-\indevent{s_{h}^{k}=s,a_{h}^{k}=a}\bigr) \min \{ 2,r_{h}^{k}(s,a) \}
    > 6\sqrt{K\ln\frac{1}{\delta}}\right\} $
    \begin{itemize}
        \item[$\blacktriangleright$] where 
        $r_{h}^{k}(s,a)
        = 8\sqrt{\frac{S\ln\frac{HSAK}{4\delta'}}{n_{h}^{k}(s,a)\vee1}} + 200 S\ln\frac{\ln\frac{HSAK}{4\delta'}}{n_{h}^{k}(s,a)\vee1}$ , $q_h^k(s,a) = q^{p,\pi^k}_h(s) \pi^k_h(a | s)$, and $q^{p,\pi}_h(s) = \Pr [s_h = s \mid s_1 = s_\text{init} , \pi , p]$.
    \end{itemize}
    
    \item $F_{4}^{basic}
    =\left\{ \exists k,s,a,h:\sum_{k'=1}^{k}\hat{c}_{h}^{k'}(s,a)-\frac{q_{h}^{k'}(s)}{u_{h}^{k'}(s)}c_{h}^{k'}(s,a)
    > \frac{\ln \frac{SAHK}{\delta'}}{2\gamma}\right\} $
    
    \item $F_{5}^{basic} =  \{ \exists k,s,a,h:
    n_{h}^{k}(s) \geq d_{max}\log\frac{HSA}{\delta} \text{ and }
    n_{h}^{k}(s,a)\leq\frac{d_{max}}{2A} \} \cap \{UseExplicitExploration = \texttt{true}\} $
    \begin{itemize}
        \item[$\blacktriangleright$] where 
        $n_{h}^{k}(s) = \sum_{a'} n_{h}^{k}(s,a')$.
    \end{itemize}
\end{itemize}
We define the basic good by $G^{basic} = \bigcap_{i=1}^{5} \lnot F_{i}^{basic}$.


\begin{lemma}   \label{lemma:basic-good-event}
    The basic good event $G^{basic}$, occurs with probability of at
    least $1-5\delta'$.
\end{lemma}
\begin{proof}
    \begin{itemize}
        \item[]
        
        \item By \citep[Theorem 4]{maurer2009empirical},
        $\Pr\left( F_{1}^{basic} \right)\leq\delta'$.
        

        \item  By  \citep[Lemma 17]{jaksch2010near} and union bounds,
        $\Pr\left( F_{2}^{basic} \right)\leq\delta'$.
        
        \item Let $Y_k = \sum_{s,a,h} \bigl(q_{h}^{k}(s,a) - \indevent{s_{h}^{k}=s,a_{h}^{k}=a}\bigr)
       \min\left(2,r_{h}^{k}(s,a)\right)$. Note that  $r_{h}^{k}$ depends on the history up to the end of episode $k-1$, $\mathcal{H}_{k-1}$.
        Therefore, 
        $\E[Y_k\mid \mathcal{H}_{k-1}] = 0$ (as $\E\bigl[\indevent{s_{h}^{k}=s,a_{h}^{k}=a} \mid{H}_{k-1}\bigr] = q_{h}^{k}(s,a)$). That is, $\sum_k Y_k$ is a martingale. Also, 
        $|Y_{k}| \leq 4 $. By Azuma-Hoeffding inequality,
        \[
            \Pr\left( F_{3}^{basic} \right) 
            = 
            \Pr\left( \sum_k Y_k > 6\sqrt{K \ln 1/\delta'} \right) 
            \leq 
            \delta'.
        \]

        
        \item By  \citep[Lemma 6]{efroni2020optimistic},
        $\Pr\left(F_{4}^{basic}\right)\leq\delta'$.\footnote{We have invoked Lemma 6 of \cite{efroni2020optimistic} with $\alpha_{h}^{k}(s',a') = \mathbb{I}\{ s'=s,a'=a\} $ and then take union bound over all $s,a$ and $h$.}
        
        \item Fix $h,s,a$ and $k$ such that $n_{h}^{k}(s) \geq d_{max}\log\frac{HSA}{\delta'}$, and let $k_{0}$ be the first episode such that $n_{h}^{k_{0}}(s)\geq d_{max}\log\frac{HSA}{\delta'}$.
        Since, for at least the first $d_{max}\log\frac{HSA}{\delta'}$ visits in $s$, we play a uniform policy in $s$, 
         $\E\left[n_{h}^{k_{0}}(s,a)\right]\geq\frac{d_{max}}{A}\log\frac{HSA}{\delta'}$. By Chernoff bound,
        \begin{align*}
            \Pr \left(n_{h}^{k_{0}}(s,a) \geq \frac{1}{2}\frac{d_{max}}{A}\right) 
            & \leq 
            e^{-\frac{1}{8}\frac{d_{max}}{A}\log\frac{HSA}{\delta'}} 
            \leq
            \frac{\delta'}{HSA},
        \end{align*} where the last holds whenever $d_{max}\geq8A$ (if not, we can actually get better regret bounds). Taking the union bound over $h,s,a$ and noting that it is sufficient to show the claim for the first $k$ that satisfies $n_{h}^{k}(s)\geq d_{max}\log\frac{HSA}{\delta'}$, gives us
        \begin{align*}
            \Pr (\lnot F_{5}^{basic}) 
            & \leq
            \delta'.
        \end{align*}

    \end{itemize}
    
Using the union bound on the above failure events and taking the complement gives us the desired result.

\end{proof}
Define $\epsilontilde^{k}_{h}(s' \mid s,a) = 8\sqrt{ \frac{p_{h}(s'\mid s,a) (1-p_{h}(s'\mid s,a)) \ln\frac{HSAK}{4\delta}}{n_{h}^{k}(s,a) \vee 1}}
    + 100\ln\frac{\ln\frac{HSAK}{4\delta}}{n_{h}^{k}(s,a) \vee 1}$.
\begin{lemma}   \label{lemma:consequences-good}
    Given the basic good event, the following relations holds:
    \begin{enumerate}
        \item $\left| p_{h}(s'\mid s,a)-\hat{p}_{h}^{k}(s'\mid s,a) \right| 
        \leq 
        \epsilontilde_{h}^{k}(s'\mid s,a)$.
        \item $\sum_{s'}\left| p_{h}(s'\mid s,a) - \hat{p}_{h}^{k}(s'\mid s,a) \right|
        \leq 
        r_{h}^{k}(s,a)$.
    \end{enumerate}
\end{lemma}
\begin{proof}

    \begin{enumerate}
        \item[]
        \item Under bandit feedback, the first inequality now holds by  $\lnot F_{1}^{basic}$ and \citep[Lemma B.13]{cohen2020ssp}.
        Under full information, recall that $\hat{p}^k$ is computed after episode $k + d^k$. That is, $\hat{p}^k \in \mathcal{P}^{k + d^k}$. Similar to the bandit case,
        \[
            \left| p_{h}(s'\mid s,a)-\hat{p}_{h}^{k}(s'\mid s,a) \right| 
            \leq 
            \epsilontilde_{h}^{k+d^k}(s'\mid s,a)
            \leq
            \epsilontilde_{h}^{k}(s'\mid s,a),
        \]
        where the second inequality is since $\epsilontilde_{h}^{k}(s'\mid s,a)$ is decreasing in $k$.
        \item The second relation simply holds by the first relation and Jensen's inequality.  
    \end{enumerate}

\end{proof}

We define the following bad events that will not occur with high probability, given that the basic good event occurs:

\begin{itemize}
    \item $F_{1}^{cond} = \left\{ 
    \sum_{k,s,a,h}q_{h}^{k} (s)\pi_{h}^{k}(s\mid a)(\\E\left[\hat{c}_{h}^{k}(s,a)\mid\mathcal{H}^{k-1}\right] - \hat{c}_{h}^{k}(s,a))
    >
    H\sqrt{K\frac{\ln H}{2\delta'}} \right\} $ 
    \begin{itemize}
        \item[$\blacktriangleright$] where $\mathcal{H}^{k-1}$ denotes the history up to episode $k$.
    \end{itemize}
    
    \item $F_{2}^{cond} = \left\{ 
    \exists h,s: \sum_{k=1}^K V_{h}^{k}(s) - V_{h}^{\pi^{k}}(s) 
    >
    \frac{H}{\gamma}\ln\frac{HSK}{\delta'}\right\} $ 
    
    \item $F_{3}^{cond} = \left\{ \exists s,a:
    \sum_{k = 1}^{K}  \sum_{h} \E^{\pi^{k},s}\left[\epsilontilde_{h}^{k}(\cdot\mid s,a)(V_{h+1}^{k}-V_{h+1}^{\pi^{k}}) \right]
    >
    \frac{H^{2}S}{2\gamma}\ln\frac{H^{2}SK}{\delta'} \right\} $

    \item $F_{4}^{cond}=\left\{ \exists h,s:\sum_{k}   \left| \{ j:j\leq k,j+d^{j}\geq k\} \right| \left(V_{h}^{k}(s)-V_{h}^{\pi^{k}}(s) \right)
    >
    d_{max}\frac{H}{\gamma}\ln\frac{HSK}{\delta'}\right\} $

\end{itemize}
We define the conditioned good event by $G^{cond}=\lnot F_{1}^{cond}\cap\lnot F_{2}^{cond}\cap\lnot F_{3}^{cond}\cap\lnot F_{4}^{cond}$.

\begin{lemma}   \label{lemma:conditioned-good-event}
    Conditioned on  $G^{basic}$, the conditioned good event $G^{cond}$, occurs with probability of at least $1-4\delta'$. That is,
\begin{align*}
    \Pr \left(G^{cond} \mid G^{basic} \right) \geq 1 - 4\delta'.
\end{align*}
\end{lemma}

\begin{proof}
\begin{itemize}
    \item[]
    \item Following the proof in \citep[appendix C.1.3]{efroni2020optimistic}, $\Pr \left( F_{i}^{cond} \mid G^{basic}  \right) \leq \delta'$  for $i=1,2,3$.
    \item By \citep[Lemmas 7 and 10]{efroni2020optimistic} and the fact that $\left| \{ j:j\leq k,j+d^{j}\geq k\} \right| \leq d_{max}$,  $\Pr \left( F_{4}^{cond}\mid G^{basic} \right) \leq \delta'$.
\end{itemize}

Using the union bound on the above failure events and taking the complement gives us the desired result.
\end{proof}
Finally, we define the global good event, $G := G^{basic}\cap G^{cond}$. Using the union bound, for $\delta > 0$ and $\delta' = \nicefrac{\delta}{9}$, $P(G) \geq 1-\delta$. 

\subsection{Proof of \cref{thm:regret-bound-unknown-p-delayed-traj} with Bandit Feedback}
\label{appendix:basic-bandit-proof}
With probability at least $1 - \delta$ the good event holds.
For now on, we assume we are outside the failure event, and then our regret bound holds with probability at least $1 - \delta$.

According to the value difference lemma of \citet{efroni2020optimistic},
\begin{align}
    \nonumber
    \regret
    & =
    \sum_{k=1}^{K}V_{1}^{\pi^{k}}(s_1^k)-V_{1}^{\pi}(s_1^k)
    \\
    \nonumber
    & =
    \sum_{k=1}^{K}V_{1}^{\pi^{k}}(s_1^k)-V_{1}^{k}(s_1^k) + \sum_{k=1}^{K}V_{1}^{k}(s_1^k)-V_{1}^{\pi}(s_1^k)
    \\
    \nonumber
    & =
    \underbrace{
    \sum_{k=1}^{K} V_{1}^{\pi^{k}}(s_1^k)-V_{1}^{k}(s_1^k)}
    _{(A)}
    \\
    \nonumber
    &  \qquad +
    \underbrace{
    \sum_{k=1}^{K}\sum_{h=1}^H \E^{\pi}\big[ \langle Q_{h}^{k}(s_h^k,\cdot),\pi_{h}^{k}(\cdot\mid s_h^k)-\pi_{h}(\cdot\mid s_h^k)\rangle ] }
    _{(B)}
    \\
    \label{eq:regret-decompose}
    & \qquad + 
    \underbrace{
    \sum_{k=1}^{K} \sum_{h=1}^H \E^{\pi} \big[ Q_{h}^{k}(s_h^k,a_h^k) - c_h^k(s_h^k,a_h^k) - \langle p_{h}(\cdot\mid s_h^k,a_h^k),V_{h+1}^{k}\rangle \big] }
    _{(C)}.
\end{align}

We continue bounding each of these terms separately.
Term (A) is bounded in \cref{sec:bound-term-a} by $\wt O \bigl(H^2 S \sqrt{A K} + H^2 S A d_{max} + \gamma K H S A + \frac{H^2 S}{\gamma} + H^2 S^2 A \bigr)$, Term (B) is bounded in \cref{sec:bound-term-b} by $\wt O \bigl( \frac{H}{\eta} + \frac{\eta}{\gamma} H^3 (K+D) + \frac{\eta}{\gamma^2} d_{max} H^3 \bigr)$ and Term (C) is bounded in \cref{sec:bound-term-c} by $\wt O \bigl( \frac{H}{\gamma} \bigr)$.
This gives a total regret bound of
\begin{align*}
    \regret
    & =
    \wt O \Bigl( H^2 S A d_{max} 
    + H^2 S \sqrt{A K} 
    + \frac{H^2 S}{\gamma} + \gamma K H S A 
    \\
    & \qquad \quad + \frac{H}{\eta} + \frac{\eta}{\gamma} H^3 (K+D) + \frac{\eta}{\gamma^2} d_{max} H^3 + \frac{H}{\gamma} + H^2 S^2 A \Bigr).
\end{align*}
Choosing $\eta = \frac{1}{H (A^{3/2}K + D)^{2/3}}$ and $\gamma = \frac{1}{ (A^{3/2}K + D)^{1/3}}$ gives the theorem's statement.

\begin{remark}[Delayed OPPO with stochastic costs under bandit feedback]
    As shown by \citet{efroni2020optimistic}, when the costs are stochastic, we can replace the importance-sampling estimator with a simple empirical average.
    This means that our estimator is now bounded by $H$ and eliminates all the terms that depend on $\gamma$.
    Thus, we obtain a regret of $\wt O (H^2 S \sqrt{A K} + H^2 \sqrt{D} + H^2 S^2 A + H^2 S A d_{max})$ that is similar to the full-information feedback case.
    Moreover, in this case we can again use explicit exploration to reduce the last term to $\wt O(H^2 S d_{max})$
\end{remark}

\subsubsection{Bounding Term (A)}
\label{sec:bound-term-a}

\begin{lemma}  
    \label{lemma:estimation-error}
    Conditioned on the good event $G$,
    \[
        \sum_{k=1}^{K} V_{1}^{\pi^{k}}(s_1^k)-V_{1}^{k}(s_1^k) 
        =
        \wt O \Bigl( H^2 S \sqrt{A K} + H^2 S^2 A + \frac{H^2 S}{\gamma} + \gamma K H S A + H^{2} S A d_{max} \Bigr).
    \]
\end{lemma}

\begin{proof}
    We start with a value difference lemma \citep{efroni2020optimistic},
    \begin{align*}
        \sum_{k=1}^{K} V_{1}^{\pi^{k}}(s_1^k)-V_{1}^{k}(s_1^k)
        & =
        \sum_{k=1}^{K} \sum_{h=1}^H \E^{\pi^{k}}\big[c_{h}^{k}(s^k_{h},a^k_{h}) - \hat{c}_{h}^{k}(s^k_{h},a^k_{h})\big] 
        \\
        & \qquad + 
        \E^{\pi^{k}}\left[\langle p_{h}(\cdot\mid s^k_{h},a^k_{h}) - \hat{p}_{h}^{k}(\cdot\mid s^k_{h},a^k_{h}) , V_{h + 1}^{k} \rangle \right]
        \\
        & \leq \underbrace{
        \sum_{k=1}^{K} \sum_{h=1}^H \E^{\pi^{k}}\big[c_{h}^{k}(s^k_{h},a^k_{h}) - \hat{c}_{h}^{k}(s^k_{h},a^k_{h})\big]}
        _{(A.1)}
        \\
        & \qquad + 
        \underbrace{
        \sum_{k=1}^{K} \sum_{h=1}^H \sum_{s'\in \Scal} \E^{\pi^{k}}\big[| p_{h}(s' \mid s^k_{h},a^k_{h}) - \hat{p}_{h}^{k}(s' \mid s^k_{h},a^k_{h})| V_{h + 1}^{\pi^{k}}(s') \big]}
        _{(A.2)}
        \\
        & \qquad + 
        \underbrace{
        \sum_{k=1}^{K}  \sum_{h=1}^H \E^{\pi^{k}}\left[\langle \epsilontilde_{h}^{k}(\cdot\mid s_{h},a_{h}) , V_{h + 1}^{k} - V_{h + 1}^{\pi^{k}} \rangle \right]}
        _{(A.3)},
    \end{align*}
    where we have used \Cref{lemma:consequences-good} for the inequality.
    For any $k,h,s$ and $a$. we have
    \begin{align*}
        c_{h}^{k}(s,a) - \hat{c}_{h}^{k}(s,a)
        & =
        c_{h}^{k}(s,a) - \E\left[ \hat{c}_{h}^{k}(s,a)\mid\mathcal{H}^{k - 1} \right] 
        + \E\left[ \hat{c}_{h}^{k}(s,a)\mid\mathcal{H}^{k - 1} \right] - \hat{c}_{h}^{k}(s,a)
        \\
        & =
        c_{h}^{k}(s,a)\left( 1 - \frac{q_{h}^{k}(s)\pi_{h}^{k}(a\mid s)}{u_{h}^{k}(s)\pi_{h}^{k}(a\mid s) + \gamma} \right) + \E\left[ \hat{c}_{h}^{k}(s,a)\mid\mathcal{H}^{k - 1} \right] - \hat{c}_{h}^{k}(s,a)
        \\
        & =
        c_{h}^{k}(s,a)\left( \frac{(u_{h}^{k}(s) - q_{h}^{k}(s))\pi_{h}^{k}(a\mid s) + \gamma}{u_{h}^{k}(s)\pi_{h}^{k}(a\mid s) + \gamma} \right) 
        \\
        & \qquad + 
        \E\left[\hat{c}_{h}^{k}(s,a)\mid\mathcal{H}^{k - 1}\right] - \hat{c}_{h}^{k}(s,a).
    \end{align*}
    Therefore (denoting $q^k_h(s) = q^{p,\pi^k}_h(s)$),
    \begin{align*}
        (A.1) 
        & =
        \sum_{k=1}^{K}  \sum_{h=1}^H \E^{\pi^k}\left[c_{h}^{k}(s_{h},a_{h}) \left(\frac{(u_{h}^{k}(s_{h})-q_{h}^{k}(s_{h}))\pi_{h}^{k}(a_{h}\mid s_{h})+\gamma}{u_{h}^{k}(s_{h})\pi_{h}^{k}(a_{h}\mid s_{h}) + \gamma} \right) \right]
        \\
        & \qquad +
        \sum_{k=1}^{K}  \sum_{h=1}^H \E^{\pi^k} \left[\E\left[\hat{c}_{h}^{k}(s_{h},a_{h}) \mid \mathcal{H}^{k-1}\right] - \hat{c}_{h}^{k}(s_{h},a_{h}) \right]
        \\
        & = 
        \underbrace{
        \sum_{k=1}^{K}  \sum_{h=1}^H \sum_{s,a} q_{h}^{k}(s)\pi_{h}^{k}(s \mid a)c_{h}^{k}(s,a) \left(\frac{(u_{h}^{k}(s)-q_{h}^{k}(s))\pi_{h}^{k}(a\mid s) + \gamma}{u_{h}^{k}(s)\pi_{h}^{k}(a\mid s) + \gamma} \right)}
        _{(A.1.1)}
        \\
        & \qquad + 
        \underbrace{
        \sum_{k=1}^{K}  \sum_{h=1}^H \sum_{s,a} q_{h}^{k}(s)\pi_{h}^{k}(s\mid a) \left( \E[\hat{c}_{h}^{k}(s,a)\mid\mathcal{H}^{k-1}]-\hat{c}_{h}^{k}(s,a)\right)}
        _{(A.1.2)}.
    \end{align*}
    Under the good event $p\in\mathcal{P}^{k-1}$. Hence by definition $u_{h}^{k}(s)\geq q_{h}^{k}(s)$.
    Therefore,
    \begin{align*}
        (A.1.1) 
        & \leq
        \sum_{k=1}^{K}  \sum_{h=1}^H \sum_{s,a} q_{h}^{k}(s)\pi^{k}(s\mid a)c_{h}^{k}(s,a) \left( \frac{(u_{h}^{k}(s)-q_{h}^{k}(s))\pi_{h}^{k}(a\mid s)+\gamma}{q_{h}^{k}(s)\pi_{h}^{k}(a\mid s)} \right)
        \\
        & =
        \sum_{k=1}^{K}  \sum_{h=1}^H \sum_{s,a} c_{h}^{k}(s,a)\left( (u_{h}^{k}(s)-q_{h}^{k}(s))\pi_{h}^{k}(a\mid s)+\gamma \right)
        \\
        & \leq
        \sum_{k=1}^{K}  \sum_{h=1}^H \sum_{s} \left( u_{h}^{k}(s)-q_{h}^{k}(s) \right) + \gamma K H S A
        \\
        & \le
        H S \sqrt{A K}
        + H S A d_{max} 
        + \gamma K H S A,
    \end{align*}
    where the last inequality follows similarly to the bound of Term (A.2) when combined with Lemma 20 of \citet{efroni2020optimistic}.
    
    Under the good event $G$ (in particular, $\lnot F_{1}^{cond}$),
    \begin{align*}
    (A.1.2) \leq H\sqrt{K\frac{\ln H}{2\delta'}}.
    \end{align*}
    By \Cref{lemma:l1norms-bound} and the fact that $V_{h+1}^{\pi^k} \leq H$,
    \begin{align*}
        (A.2) 
        & \leq 
        H^2 S \sqrt{A K} 
        + H^2 S A d_{max} + H^2 S^2 A.
    \end{align*}
    Finally, conditioned on the good event ($\lnot F_{3}^{cond}$),
    \begin{align*}
        (A.3) 
        & \leq
        \frac{H^{2}S}{2\gamma} \ln\frac{H^{2}SK}{\delta'}.
    \end{align*}
    We get that term $(A)$ can be bounded by,
    \begin{align*}
        \sum_{k = 1}^{K} V_{1}^{\pi^{k}}(s_1^k)-V_{1}^{k}(s_1^k) 
        & \leq
        (A.1.1)+(A.1.2)+(A.2)+(A.3)
        \\
        & \lesssim 
        H^2 S \sqrt{A K} 
        + H^2 S A d_{max}
        + \gamma K H S A
        + \frac{H^2 S}{\gamma},
    \end{align*}
    where $\lesssim$ ignores poly-logarithmic factors.
\end{proof}

\begin{lemma}   
    \label{lemma:l1norms-bound}
    Under the good event,
    \begin{align*}
        \sum_{k = 1}^{K} \sum_{h=1}^H \sum_{s' \in \mathcal{S}} \E^{\pi^{k}} \Bigl[ \bigl|p_{h}(s' \mid s^k_{h},a^k_{h}) - \hat{p}_{h}^{k}(s' \mid s^k_{h},a^k_{h}) \bigr| \Bigr]
        & \lesssim 
        H S \sqrt{A K} 
        +H S A d_{max} + H S^2 A.
    \end{align*}
\end{lemma}

\begin{proof}

    Define,
    \[
        \Ksa(s,a,h)=\left\{ k \in [K]:s_{h}^{k}=s,a_{h}^{k}=a,
        n_h^k(s,a)
        \leq d_{max}\right\},
    \]
    which are the episodes in which we visited $(s,a,h)$ but haven't observe haven't observed more than $d_{max}$ samples. For episodes in $\Ksa(s,a,h)$, we bound the estimation error of $p$ by a constant. For the rest of the episodes we utilize the fact that the amount unobserved feedback is smaller than the observed feedback (see \eqref{eq:observed-geq-unobserved} below).
    
    Note that $|\Ksa(s,a,h)|\leq2d_{max}$. 
    Thus, 
    \begin{align}
        \nonumber
        \sum_{k=1}^{K} & \sum_{h=1}^{H}\sum_{s'\in\mathcal{S}}  \E^{\pi^{k}}\Bigl[|p_{h}(s'\mid s_{h}^{k},a_{h}^{k})-\hat{p}_{h}^{k}(s'\mid s_{h}^{k},a_{h}^{k})|\Bigr]
        =
        \\
        \nonumber
        & =
        \sum_{k=1}^{K}\sum_{s,a,h} q_{h}^{k}(s,a)\sum_{s'\in\mathcal{S}}|p_{h}(s'\mid s,a)-\hat{p}_{h}^{k}(s'\mid s,a)|\\
        & \leq    
        \nonumber
        \sum_{k=1}^{K}\sum_{s,a,h} q_{h}^{k}(s,a)\min \{ 2,r_{h}^{k}(s,a) \}\\
        & \le
        \nonumber
        \sum_{k=1}^{K}\sum_{s,a,h} \bigl(q_{h}^{k}(s,a)-\indevent{s_{h}^{k}=s,a_{h}^{k}=a}\bigr)\min  \{ 2,r_{h}^{k}(s,a) \} \\
        & \qquad +
        \nonumber
        2\sum_{s,a,h} \sum_{k\in \Ksa(s,a,h)} \indevent{s_{h}^{k}=s,a_{h}^{k}=a}\\
        & \qquad +
        \nonumber
        \sum_{s,a,h} \sum_{k \notin \Ksa(s,a,h)} \indevent{s_{h}^{k}=s,a_{h}^{k}=a}r_{h}^{k}(s,a)\\
        & \lesssim    
        \nonumber
        \sqrt{K} + H S A d_{max} + \sum_{s,a,h} \sum_{k \notin \Ksa(s,a,h)}  \indevent{s_{h}^{k}=s,a_{h}^{k}=a}r_{h}^{k}(s,a)\\
        & \lesssim
        \sqrt{K} + H S A d_{max} + \sqrt{S} \sum_{s,a,h} \sum_{k \notin \Ksa(s,a,h)}  \frac{\indevent{s_{h}^{k}=s,a_{h}^{k}=a}}{\sqrt{n_{h}^{k}(s,a)\vee1}}
        \nonumber
        \\
        & \qquad + 
        S\sum_{s,a,h}  \sum_{k \notin \Ksa(s,a,h)}  \frac{\indevent{s_{h}^{k}=s,a_{h}^{k}=a}}{n_{h}^{k}(s,a)\vee1}.
        \label{eq:shrink-lemma-eq1}
    \end{align}
    The first inequality  
    follows the fact that $\Vert p_{h}(\cdot\mid s,a)-\hat{p}_{h}^{k}(\cdot\mid s,a)\Vert _{1} \leq 2$ and \Cref{lemma:consequences-good}. And the third inequality is by $\lnot F_{3}^{basic}$, and $|\Ksa(s,a,h)|\leq d_{max}$.
    Now,
    \begin{align*}
        & \sum_{k\notin\Ksa(s,a,h)} \frac{\indevent{s_{h}^{k}=s,a_{h}^{k}=a}}{\sqrt{n_{h}^{k}(s,a)\vee1}} 
        \le
        \\
        & \le
        \sum_{k\notin\Ksa(s,a,h)}\frac{\indevent{s_{h}^{k}=s,a_{h}^{k}=a}}{\sqrt{1\vee\sum_{j=1}^{k-1}\indevent{s_{h}^{j}=s,a_{h}^{j}=a}}}\cdot\sqrt{\frac{1\vee\sum_{j=1}^{k-1}\indevent{s_{h}^{j}=s,a_{h}^{j}=a}}{1\vee\sum_{j:j+d^{j}\le k-1}\indevent{s_{h}^{j}=s,a_{h}^{j}=a}}}\\
        & \le
        \sum_{k\notin\Ksa(s,a,h)}\frac{\indevent{s_{h}^{k}=s,a_{h}^{k}=a}}{\sqrt{1\vee\sum_{j=1}^{k-1}\indevent{s_{h}^{j}=s,a_{h}^{j}=a}}}\cdot\sqrt{1+\frac{1\vee\sum_{j:j<k,j+d^{j}\ge k}\indevent{s_{h}^{j}=s,a_{h}^{j}=a}}{1\vee\sum_{j:j+d^{j}\le k-1}\indevent{s_{h}^{j}=s,a_{h}^{j}=a}}}.
    \end{align*}
    Since for any $k\notin\Ksa(s,a,h)$,
    \begin{align}
    \sum_{j:j<k,j+d^{j}\ge k}\indevent{s_{h}^{j}=s,a_{h}^{j}=a}\leq d_{max}\leq\sum_{j:j+d^{j}\le k-1}\indevent{s_{h}^{j}=s,a_{h}^{j}=a}
    \label{eq:observed-geq-unobserved}
    \end{align}
    we have
    \begin{align}
        \sum_{k\notin\Ksa(s,a,h)}\frac{\indevent{s_{h}^{k}=s,a_{h}^{k}=a}}{\sqrt{n_{h}^{k}(s,a)\vee1}} 
        & \lesssim
        \nonumber
        \sum_{k\notin\Ksa(s,a,h)}\frac{\indevent{s_{h}^{k}=s,a_{h}^{k}=a}}{\sqrt{1\vee\sum_{j=1}^{k-1}\indevent{s_{h}^{j}=s,a_{h}^{j}=a}}}\\
        & \leq
        \sum_{k=1}^{K}\frac{\indevent{s_{h}^{k}=s,a_{h}^{k}=a}}{\sqrt{1\vee\sum_{j=1}^{k-1}\indevent{s_{h}^{j}=s,a_{h}^{j}=a}}}.
        \label{eq:shrink-lemma-bound-by-no-delay}
    \end{align}
    In the same way,
    \begin{align}
        \sum_{k\notin\Ksa(s,a,h)}\frac{\indevent{s_{h}^{k}=s,a_{h}^{k}=a}}{n_{h}^{k}(s,a)\vee1}
        & \lesssim
        \sum_{k=1}^{K}\frac{\indevent{s_{h}^{k}=s,a_{h}^{k}=a}}{1\vee\sum_{j=1}^{k-1}\indevent{s_{h}^{j}=s,a_{h}^{j}=a}}.
        \label{eq:shrink-lemma-bound-by-no-delay-no-sqrt}
    \end{align}

    Finally, \eqref{eq:shrink-lemma-bound-by-no-delay} and \eqref{eq:shrink-lemma-bound-by-no-delay-no-sqrt} appear in the non-delayed setting and can be bounded when summing over $(s,a,h)$ by $\tilde{O}(H\sqrt{SAK})$ and $\tilde{O}(H S A)$, respectively (see for example ,\citep[Lemma 4]{jin2019learning}). Plugging back in \eqref{eq:shrink-lemma-eq1} completes the proof.
\end{proof}

\subsubsection{Bounding Term (B)}
\label{sec:bound-term-b}

\begin{lemma}   
    \label{lemma:OMD}
    Let $s \in S$.
    Conditioned on the good event $G$,
    \begin{align*}
        \sum_{k = 1}^{K} \sum_{h=1}^H \left\langle Q_{h}^{k}(s,\cdot), \pi_{h}^{k}(\cdot\mid s) - \pi_{h}(\cdot\mid s) \right\rangle  
        \leq
        \frac{H\log(A)}{\eta} + \eta\frac{H^{3}}{\gamma}(K+D) + \frac{\eta} {\gamma^{2}} d_{max}H^{3}\ln\frac{H}{\delta}.
    \end{align*} 

\end{lemma}
\begin{proof}
We adopt the technique presented in \cite{gyorgy2020adapting} for MAB, in order to bound the term $\sum_{k} \langle Q_{h}^{k}(s,\cdot),\pi_{h}^{k}(\cdot\mid s) - \pi_{h}(\cdot\mid s)\rangle$, for each $s$ and $h$ separately. 

The proof uses a comparison to a ``cheating'' algorithm that does not experience delay and sees onoe step into the future.
Define
\begin{align*}
    \picheat_{h}^{k}(a\mid s) 
    & =
    \frac{\exp\left( -\eta\sum_{j:j\leq k-1}Q_{h}^{j}(s,a) \right)}{\sum_{a' \in \mathcal{A}} \exp\left( -\eta\sum_{j:j\leq k-1} Q_{h}^{j}(s,a') \right)}.
\end{align*}

We break the sum term in the following way:
\begin{align*}
    \sum_{k} \left\langle Q_{h}^{k}(s,\cdot),\pi_{h}^{k}(\cdot\mid s) - \pi_{h}(\cdot\mid s)\right\rangle
    & =
    \underbrace{\sum_{k} \left\langle Q_{h}^{k}(s,\cdot),\picheat_{h}^{k+1}(\cdot\mid s) - \pi_{h}(\cdot\mid s)\right\rangle }
    _{(B.1)}
    \\
    & \qquad +
    \underbrace{\sum_{k} \left\langle Q_{h}^{k}(s,\cdot),\pi_{h}^{k}(\cdot\mid s) - \picheat_{h}^{k+1}(\cdot\mid s)\right\rangle }
    _{(B.2)}.
\end{align*}
 
Term (B.1) is the regret of the ``cheating'' algorithm.
Using \citep[Theorem 3]{joulani2020modular},\footnote{We choose the regularizers in \citep{joulani2020modular} to be $q_0(x) = r_1(x) = \frac{1}{\eta}\sum_{i}x_{i}\log x_{i}$ and the rest are zero, which makes the update of their ADA-MD algorithm as in our policy improvement step. The statement now follows from  \citep[Theorem 3]{joulani2020modular}, the fact that the Bregman divergence is positive and that entropy is bounded by $\log A$.}
\begin{align}
    (B.1) & \leq \frac{\ln(A)}{\eta}.     \label{eq:OMD-cheat}
\end{align}

Now, using the definition of $\picheat_{h}^{k}$, 
\begin{align*}
    \frac{\picheat_{h}^{k+1}(a\mid s)}{\pi_{h}^{k}(a\mid s)} 
    & =
    \frac{\exp\left(-\eta\sum_{j: j\leq k} Q_{h}^{j}(s,a)\right)}
    {\sum_{a'}\exp\left(-\eta\sum_{j: j\leq k}  Q_{h}^{j}(s,a')\right)} 
    \cdot 
    \frac{\sum_{a'} \exp\left(-\eta\sum_{j:j+d^{j} \leq k-1}Q_{h}^{j}(s,a')\right)}
    {\exp\left(-\eta\sum_{j:j+d^{j}\leq k-1} Q_{h}^{j}(s,a)\right)}\\
    & =
    \exp\left(-\eta\sum_{j:j\leq k,j+d^{j}\geq k}Q_{h}^{j}(s,a)\right) \cdot \frac{\sum_{a'} \exp\left(-\eta\sum_{j:j+d^{j} \leq k-1} Q_{h}^{j}(s,a')\right)}{\sum_{a'} \exp\left(-\eta\sum_{j: j\leq k} Q_{h}^{j}(s,a')\right)}\\
    & \geq
    \exp\left(-\eta\sum_{j:j\leq k,j+d^{j}\geq k} Q_{h}^{j}(s,a)\right)\\
    & \geq
    1-\eta\sum_{j:j\leq k,j+d^{j}\geq k} Q_{h}^{j}(s,a)
\end{align*}
where in the first inequality we have used $\sum_{j:j+d^{j}\leq k-1}Q_{h}^{j}(s,a)\leq \sum_{j:j \le k} Q_{h}^{j}(s,a')$,
and for the second inequality we have used the fact that $e^{x}\geq1+x$
for any $x$. Using the above, 
\begin{align*}
    (B.2) 
    & =
    \sum_{k} \left\langle Q_{h}^{k}(s,\cdot),\pi_{h}^{k}(\cdot\mid s) - \picheat_{h}^{k+1}(\cdot\mid s)\right\rangle 
    \\
    & =
    \sum_{k} \sum_{a \in \mathcal{A}} Q_{h}^{k}(s,a)\left(\pi_{h}^{k}(a\mid s) - \picheat_{h}^{k+1}(a\mid s)\right)
    \\
    & =
    \sum_{k}\sum_{a \in \mathcal{A}} Q_{h}^{k}(s,a)\pi_{h}^{k}(a\mid s)\left(1-\frac{\picheat_{h}^{k+1}(a\mid s)}{\pi_{h}^{k}(a\mid s)}\right)
    \\
    & \leq
    \eta\sum_{k}\sum_{a \in \mathcal{A}} \pi_{h}^{k}(a\mid s)Q_{h}^{k}(s,a)\sum_{j:j\leq k,j+d^{j}\geq k}Q_{h}^{j}(s,a)
    \\
    & \leq
    \eta\sum_{k}\sum_{a \in \mathcal{A}} \pi_{h}^{k}(a\mid s)Q_{h}^{k}(s,a)\sum_{j:j\leq k,j+d^{j}\geq k} \left(\hat{c}_{h}^{j}(s,a)+\langle p(\cdot\mid s,a),V_{h+1}^{j}\rangle\right)
    \\
    \tag{\ensuremath{\hat{c}_{h}^{j}(s,a),V_{h+1}^{j}(s)\leq\frac{H}{\gamma}}}
    &\leq
    \eta\sum_{k}\sum_{a \in \mathcal{A}} \pi_{h}^{k}(a\mid s)Q_{h}^{k}(s,a)\sum_{j:j\leq k,j+d^{j}\geq k}\frac{H}{\gamma}
    \\
    & \leq
    \eta\frac{H}{\gamma}\sum_{k} |\{j:j\leq k,j+d^{j}\geq k\}| \sum_{a \in \mathcal{A}} \pi_{h}^{k}(a\mid s)Q_{h}^{k}(s,a)
    \\
    & =
    \eta\frac{H}{\gamma}\sum_{k} |\{j:j\leq k,j+d^{j}\geq k\}| V_{h}^{k}(s).
\end{align*}
Under the good event (in particular, $\lnot F_{4}^{cond}$),
\begin{align*}
    \sum_{k} |\{j:j\leq k,j+d^{j} \geq k\}| \left( V_{h}^{k}(s)-V_{h}^{\pi^{k}}(s) \right) 
    & \leq
    d_{max}\frac{H}{\gamma}\ln\frac{H}{\delta}.
\end{align*}
Therefore, 
\begin{align}
    (B.2)
    & \leq   \nonumber
    \eta\frac{H}{\gamma}\sum_{k} |\{j:j\leq k,j+d^{j}\geq k\}|V_{h}^{\pi^{k}}(s) + \frac{\eta}{\gamma^{2}}d_{max}H^2 \ln\frac{H}{\delta}\\
    & \leq   \nonumber
    \eta\frac{H^{2}}{\gamma} \sum_{k} |\{j:j\leq k,j+d^{j}\geq k\}| + \frac{\eta}{\gamma^{2}}d_{max}H^2 \ln\frac{H}{\delta}\\
    & \leq 
    \eta\frac{H^{2}}{\gamma}(K+D) + \frac{\eta}{\gamma^{2}}d_{max}H^2 \ln\frac{H}{\delta}.  \label{eq:OMD-drift}
\end{align}
where the last inequality holds since,
\begin{align*}
    \sum_{k} |\{j & :j\leq k,j+d^{j}\geq k\}| 
    = 
    \sum_{j}\sum_{k} \ind\{j\leq k\leq j+d^{j}\}
    \leq
    \sum_{j} (1 + d^{j})
    \leq 
    K+D.
\end{align*}
Combining \eqref{eq:OMD-cheat} and \eqref{eq:OMD-drift} and summing over $h$, completes the proof.
\end{proof}

\subsubsection{Bounding Term (C)}
\label{sec:bound-term-c}

\begin{lemma}  
    \label{lemma:optimism}
    Conditioned on the good event $G$,
    \begin{align*}
        \sum_{k = 1}^{K} \sum_{h=1}^H \E^{\pi}\left[ Q_{h}^{k}(s^k_{h},a^k_{h}) - c_{h}^{k}(s^k_{h},a^k_{h}) - \left\langle p_{h}(\cdot\mid s^k_{h},a^k_{h}) , V_{h + 1}^{k}\right\rangle \right] 
        \leq
        \frac{H}{2\gamma}\log \frac{SAHK}{\delta'}.
    \end{align*}
\end{lemma}

\begin{proof}
    Using Bellman equations,
    \begin{align*}
        \sum_{k  = 1}^{K} \sum_{h=1}^H \E^{\pi}\bigl[ Q_{h}^{k}(s^k_{h},a^k_{h}) & - c_{h}^{k}(s^k_{h},a^k_{h}) - \left\langle p_{h}(\cdot\mid s^k_{h},a^k_{h}) , V_{h + 1}^{k}\right\rangle \bigr]
        =
        \\
        & =
        \underbrace{\sum_{k  = 1}^{K}\sum_{h=1}^H \E^{\pi}\left[\hat{c}_{h}^{k}(s^k_{h},a^k_{h}) - c_{h}^{k}(s^k_{h},a^k_{h})\right]}_{(C.1)}
        \\
        & \qquad +
        \underbrace{\sum_{k  = 1}^{K}\sum_{h=1}^H \E^{\pi}\left[\left\langle \hat{p}_{h}^{k}(\cdot\mid s^k_{h},a^k_{h}),V_{h + 1}^{k}\right\rangle  - \left\langle p_{h}(\cdot\mid s^k_{h},a^k_{h}),V_{h + 1}^{k}\right\rangle \right]}_{(C.2)}.
    \end{align*}
    For any $h,s$ and $a$, under the good event,
    \begin{align*}
        \sum_{k  = 1}^{K} \hat{c}_{h}^{k}(s,a) - c_{h}^{k}(s,a)
        &\leq
        \sum_{k  = 1}^{K} \hat{c}_{h}^{k}(s,a) - \frac{q_{h}^{k}(s)}{u_{h}^{k}(s)}c_{h}^{k}(s,a)
        \leq
        \frac{\log(\frac{SAHK}{\delta'})}{2\gamma},
    \end{align*}
    
    where the first inequality is due to the fact that under the good
    event $p\in\mathcal{P}^{k - 1}$ and so $u_{h}^{k}(s)=\max_{\hat{p}\in\mathcal{P}^{k - 1}}q_{h}^{\hat p, \pi^k}(s) \geq q_{h}^{p, \pi^k}(s) = q_{h}^{k}(s)$.
    The second inequality follows directly from $\lnot F_{4}^{basic}$.
    Therefore,
    \begin{align*}
        (C.1) 
        & \leq
        \frac{H}{2\gamma}\log \frac{SAHK}{\delta'}.
    \end{align*}
    Once again, since under the good event $p\in\mathcal{P}^{k}$, then
    for all $h,s$ and $a$,
    \begin{align*}
        \left\langle \hat{p}_{h}^{k}(\cdot\mid s,a),V_{h + 1}^{k}\right\rangle  & =
        \min_{\hat{p}_{h}(\cdot\mid s,a)\in\mathcal{P}^{k}} \left\langle \hat{p}_{h}^{k}(\cdot\mid s,a) , V_{h + 1}^{k} \right\rangle 
        \leq
        \left\langle p_{h}(\cdot\mid s,a) , V_{h + 1}^{k}\right\rangle.
    \end{align*}
    Therefore $(C.2) \leq 0$,
    which completes the proof of the lemma.
\end{proof}

\subsection{Proof of \cref{thm:regret-bound-unknown-p-delayed-traj} under full-information feedback}     \label{appendix:basic-full-proof}

The proof follows almost immediately from the proof in \cref{appendix:basic-bandit-proof}, by noting that some of the terms become zero since we use the actual cost function and not an estimated one. In addition, in this setting we use the explicit exploration which yield a better bound on term $(A)$. 

Let $\Kexp(s,h)$ be the episodes in which we used the uniform policy in state $s$ at time $h$, because we did not receive enough feedback from that state. That is,
\[
  \Kexp(s,h) =  
  \left\{ k \in [K]:
  s_{h}^{k}=s,n_{h}^{k}(s)\leq d_{max}\log\frac{HSA}{\delta}\right\}.
\]
Also, define $\Kexp = \bigcup_{s,h} \Kexp(s,h)$. Recall that the algorithm keeps track of $\Kexp$, and preforms the policy improvement step only with respect to rounds that are not in $\Kexp$.
For any $(k,s,h)$ we have that $m_{h}^{k}\left(s\right) - n_{h}^{k}\left(s\right) \leq d_{max}$ and thus,    
\begin{align*}
    |\Kexp(s,h)|
    & =
    \left|\left\{ 
    k:s_{h}^{k} = s, n_{h}^{k}(s)\leq d_{max}\log\frac{HSA}{\delta}
    \right\}\right|  
    \\
    & \leq
    \left|\left\{ 
    k:s_{h}^{k} = s, m_{h}^{k}(s)\leq 2d_{max}\log\frac{HSA}{\delta}
    \right\}\right|
    \lesssim
    d_{max},
\end{align*}
By taking the union over $s$ and $h$ we have, $|\Kexp| \lesssim H S d_{max}$.

Similarly to the proof in \Cref{appendix:basic-bandit-proof}, we use the value difference lemma of \citet{efroni2020optimistic}, on episodes that are not in $\Kexp$,
\begin{align}
    \nonumber
    \regret
    & \lesssim
    H^2 S d_{max}
    + \sum_{k \notin \Kexp}V_{1}^{\pi^{k}}(s_1^k)-V_{1}^{\pi}(s_1^k)
    \\
    \nonumber
    & =
    H^2 S d_{max}
    + \underbrace{
    \sum_{k \notin \Kexp} V_{1}^{\pi^{k}}(s_1^k)-V_{1}^{k}(s_1^k)}
    _{(A)}
    \\
    \nonumber
    & \qquad +
    \underbrace{
    \sum_{k \notin \Kexp} \sum_{h=1}^H 
    \E^{\pi}\big[ \langle Q_{h}^{k}(s_h^k,\cdot),\pi_{h}^{k}(\cdot\mid s_h^k)-\pi^k_{h}(\cdot\mid s_h^k)\rangle ] }
    _{(B)}
    \\
    \label{eq:regret-decompose-full}
    & \qquad + 
    \underbrace{
    \sum_{k \notin \Kexp} \sum_{h=1}^H \E^{\pi} \big[ 
    \langle \hat{p}^{k}_{h}(\cdot\mid s_h^k,a_h^k) -  p_{h}(\cdot\mid s_h^k,a_h^k)
    ,V_{h+1}^{k}\rangle
     \big] }
    _{(C)}.
\end{align}

We continue bounding each of these terms separately.
Term (A) is bounded in \cref{sec:bound-term-a-full} by $\wt O \bigl(H^2 S \sqrt{A K} + H^2 S^{3/2} A^{3/2} \sqrt{d_{max}} + H^2 S^2 A^2 + H^2 S d_{max} \bigr)$.
Terms (B) and (C) are bounded in \cref{sec:bound-term-b-c-full} by $\wt O \bigl(  \frac{H}{\eta}+\eta H^{3}(D+K) \bigr)$.
This gives a total regret bound of
\begin{align*}
    \regret
    & = 
    \wt O \Bigl( H^2 S \sqrt{AK}
    +\frac{H}{\eta}
    +\eta H^{3}(D+K)
    + H^{2} S^{3/2} A^{3/2} \sqrt{d_{max}} 
    + H^2 S d_{max} + H^2 S^2 A^2 \Bigr)
    \\
    & \leq 
    \wt O \Bigl( H^2 S \sqrt{AK} 
    + H^2 S d_{max} + H^{2} S^2 A^3
    +\frac{H}{\eta}
    +\eta H^{3}(D+K)  \Bigr),
\end{align*}
where the last is because $H^{2} S^{3/2} A^{3/2} \sqrt{d_{max}} \leq O( H^2 S d_{max} + H^{2} S^2 A^3)$. Choosing $\eta = \frac{1}{H\sqrt{K + D}}$ gives the theorem's statement.

\begin{remark}
    Recall that we compute $\hat{p}^k$ at the end of episode $k+d^k$. However, in our analysis we utilize only feedbacks that returned before round $k$ (formally the shift is done in \Cref{lemma:consequences-good}). This was done to ensure the validity of our concentration bounds. For example, the concentration in $F_{3}^{basic}$ (see \cref{appendix:fail-events}) requires $r_{h}^{k}$ to depend only on the history up to round $k$. But, by round $k + d^k$ we might observe feedbacks from episodes that occur after episode $k$.
    
    With that being said, if $k + d^k$ is strictly monotone in $k$ (e.g., under fixed delay), then all feedback of episodes $j < k$ are available at time $k + d^k$. In that case we can essentially achieve the bounds of $\Cref{thm:regret-bound-unknown-p-non-delayed-traj}$, even when trajectory feedback is delayed, and even without explicit exploration.
\end{remark}

\subsubsection{Bounding Term (A) under full-information feedback}
\label{sec:bound-term-a-full}

Term $(A)$ under full-information can be written as,
\begin{align*}
    (A) 
    & = 
    \sum_{k \not\in \Kexp} V_{1}^{\pi^{k}}(s_1^k)-V_{1}^{k}(s_1^k)\\
    & =
    \sum_{k \not\in \Kexp} \sum_{h=1}^H
    \E^{\pi^{k}}\left[\langle p_{h}(\cdot\mid s^k_{h},a^k_{h}) - \hat{p}_{h}^{k}(\cdot\mid s^k_{h},a^k_{h}) , V_{h + 1}^{k} \rangle \right]\\
    & \leq
    H \sum_{k = 1}^{K} \sum_{h=1}^H \sum_{s'}
    \E^{\pi^{k}}\left[| p_{h}(s' \mid s^k_{h},a^k_{h}) - \hat{p}_{h}^{k}(s' \mid s^k_{h},a^k_{h}) | \right].
\end{align*}
The last is bounded by \Cref{lemma:l1norms-bound-exp}, which is analogues to \Cref{lemma:l1norms-bound}. This gives us the following bound on term (A):
\begin{align}
    (A)
    \lesssim
    H^2 S \sqrt{A K} + H^2 S^{3/2} A^{3/2} \sqrt{d_{max}} + H^2 S^2 A^2 + H^2 S d_{max}.
    \label{eq:estimation-err-full}
\end{align}
\begin{lemma}       \label{lemma:l1norms-bound-exp}
    Under the good event, with explicit exploration ($UseExplicitExploration = \texttt{true}$), 
    \begin{align*}
        \sum_{k = 1}^{K} \sum_{h=1}^H \sum_{s' \in \mathcal{S}} \E^{\pi^{k}} \Bigl[ \bigl|p_{h}(s' \mid s^k_{h},a^k_{h}) - \hat{p}_{h}^{k}(s' \mid s^k_{h},a^k_{h}) \bigr| \Bigr]
        & \lesssim 
        H S \sqrt{A K} 
        + H S d_{max}
        \\
        & \qquad
        + H S^2 A^2
        + H S^{3/2} A^{3/2} \sqrt{d_{max}}.
    \end{align*}
\end{lemma}

\begin{proof}
    Similarly to the proof of \Cref{lemma:l1norms-bound},
    \begin{align*}
        \sum_{k=1}^{K}\sum_{h=1}^{H} & \sum_{s'\in\mathcal{S}} \E^{\pi^{k}}\Bigl[|p_{h}(s'\mid s_{h}^{k},a_{h}^{k})-\hat{p}_{h}^{k}(s'\mid s_{h}^{k},a_{h}^{k})|\Bigr]
        =
        \\
        & =
        \sum_{k=1}^{K}\sum_{s,a,h} q_{h}^{k}(s,a)\sum_{s'\in\mathcal{S}}|p_{h}(s'\mid s,a)-\hat{p}_{h}^{k}(s'\mid s,a)|\\
        & \leq
        \sum_{k=1}^{K}\sum_{s,a,h} q_{h}^{k}(s,a)\min \{ 2,r_{h}^{k}(s,k) \} \\
        & =
        \sum_{k=1}^{K}\sum_{s,a,h} \bigl(q_{h}^{k}(s,a)-\indevent{s_{h}^{k}=s,a_{h}^{k}=a}\bigr)\min \{ 2,r_{h}^{k}(s,k) \} \\
        & \qquad +
        \sum_{k=1}^{K}\sum_{s,a,h} \indevent{s_{h}^{k}=s,a_{h}^{k}=a}\min \{ 2,r_{h}^{k}(s,k) \} \\
        & \lesssim      \tag{by $\lnot F_{3}^{basic}$}
        \sqrt{K} + \sum_{k=1}^{K}\sum_{s,a,h} \indevent{s_{h}^{k}=s,a_{h}^{k}=a}\min \{ 2,r_{h}^{k}(s,k) \} \\
        & \leq
        \sqrt{K}+2\sum_{s,h} \sum_{k\in\Kexp(s,h)}
        \underbrace{
        \sum_{a}\indevent{s_{h}^{k}=s,a_{h}^{k}=a}}
        _{ \leq 1}\\
        & \qquad +
        \sum_{k\notin\Kexp}\sum_{s,a,h} \indevent{s_{h}^{k}=s,a_{h}^{k}=a}r_{h}^{k}(s,k)\\
        & \lesssim      
        \tag{$|\Kexp(s,h)| \lesssim d_{max}$}
        \sqrt{K} + H S d_{max}\\
        & \qquad +
        \sqrt{S}\sum_{s,a,h}\sum_{k\notin\Kexp} \frac{\indevent{s_{h}^{k}=s,a_{h}^{k}=a}}{\sqrt{n_{h}^{k}(s,a)\vee1}}
        + S\sum_{s,a,h}\sum_{k\notin\Kexp}\frac{\indevent{s_{h}^{k}=s,a_{h}^{k}=a}}{n_{h}^{k}(s,a)\vee1}.
    \end{align*}
    The last two terms are bounded using \Cref{lemma:indicator-shrink-exp,lemma:indicator-shrink-not-sqrt-exp}, which completes the proof.
\end{proof}

\begin{lemma}
    \label{lemma:indicator-shrink-exp}
    It holds that
    \begin{align*}
        \sum_{s \in \mathcal{S}} \sum_{a \in \mathcal{A}} \sum_{h=1}^H \sum_{k\notin\Kexp} \frac{ \indevent{s_{h}^{k}=s,a_{h}^{k}=a}}{\sqrt{n_{h}^{k}(s,a)\vee1}} 
        & \lesssim
        H \sqrt{SAK} + HS A^{3/2} \sqrt{d_{max}}.
    \end{align*}
\end{lemma}

\begin{proof}
    For any $s,a$ and $h$,
    \begin{align*}
        & \sum_{k\notin\Kexp} \frac{ \indevent{s_{h}^{k}=s,a_{h}^{k}=a}}{\sqrt{n_{h}^{k}(s,a)\vee1}}
        \le
        \\
        & \le
        \sum_{k\notin\Kexp(s,h)} \frac{ \indevent{s_{h}^{k}=s,a_{h}^{k}=a}}{\sqrt{1\vee\sum_{j=1}^{k-1} \indevent{s_{h}^{j}=s,a_{h}^{j}=a}}} \cdot \sqrt{ \frac{1\vee\sum_{j=1}^{k-1} \indevent{s_{h}^{j}=s,a_{h}^{j}=a}}{1\vee\sum_{j:j + d^{j}\le k-1} \indevent{s_{h}^{j}=s,a_{h}^{j}=a}}}
        \\
        & \le
        \sum_{k\notin\Kexp(s,h)} \frac{ \indevent{s_{h}^{k}=s,a_{h}^{k}=a}}{\sqrt{1\vee\sum_{j=1}^{k-1} \indevent{s_{h}^{j}=s,a_{h}^{j}=a}}} \cdot \sqrt{1 +  \frac{1\vee\sum_{j:j<k,j + d^{j}\ge k} \indevent{s_{h}^{j}=s,a_{h}^{j}=a}}{1\vee\sum_{j:j + d^{j}\le k-1} \indevent{s_{h}^{j}=s,a_{h}^{j}=a}}}
        \\
        & \leq
        \sum_{k\notin\Kexp(s,h)} \frac{ \indevent{s_{h}^{k}=s,a_{h}^{k}=a}}{\sqrt{1\vee\sum_{j=1}^{k-1} \indevent{s_{h}^{j}=s,a_{h}^{j}=a}}} \cdot \sqrt{1 +  \frac{1 + \sum_{j:j<k,j + d^{j}\ge k} \indevent{s_{h}^{j}=s,a_{h}^{j}=a}}{1\vee\sum_{j:j + d^{j}\le k-1} \indevent{s_{h}^{j}=s,a_{h}^{j}=a}}}
        \\
        & \leq
        \sum_{k\notin\Kexp(s,h)} \frac{ \indevent{s_{h}^{k}=s,a_{h}^{k}=a}}{\sqrt{1\vee\sum_{j=1}^{k-1} \indevent{s_{h}^{j}=s,a_{h}^{j}=a}}} \cdot \sqrt{2 +  \frac{\sum_{j:j<k,j + d^{j}\ge k} \indevent{s_{h}^{j}=s,a_{h}^{j}=a}}{1\vee\sum_{j:j + d^{j}\le k-1} \indevent{s_{h}^{j}=s,a_{h}^{j}=a}}}
        \\
        & \lesssim
        \underbrace{
        \sum_{k=1}^{K} \frac{ \indevent{s_{h}^{k}=s,a_{h}^{k}=a}}{\sqrt{1\vee\sum_{j=1}^{k-1} \indevent{s_{h}^{j}=s,a_{h}^{j}=a}}}}
        _{(D.1)}
        \\
        & \qquad + 
        \underbrace{
        \sum_{k\notin\Kexp(s,h)} \frac{ \indevent{s_{h}^{k}=s,a_{h}^{k}=a}}{\sqrt{1\vee\sum_{j=1}^{k-1} \indevent{s_{h}^{j}=s,a_{h}^{j}=a}}} \cdot \sqrt{ \frac{\sum_{j:j<k,j + d^{j}\ge k} \indevent{s_{h}^{j}=s,a_{h}^{j}=a}}{1\vee\sum_{j:j + d^{j}\le k-1} \indevent{s_{h}^{j}=s,a_{h}^{j}=a}}}}
        _{(D.2)}.
    \end{align*}
    As mentioned before, $(D.1)$ appears in the non-delayed setting and can be bounded when summing over $(s,a,h)$ by $\tilde{O}(H\sqrt{SAK})$.
    For bounding $(D.2)$ we'll use the notion of $\Kcal(s,a,h)$ defined in the proof of \cref{lemma:l1norms-bound} and the fact that outside $\Kcal(s,a,h)$ the inequality in  \eqref{eq:observed-geq-unobserved} holds. Recall
    that $|\Kcal(s,a,h)| \le 2 d_{max}$ since every visit is observable after $d_{max}$ episodes.
    
    Now summing $(D.2)$ over $s$ and $a$,
    \begin{align*}
    & \sum_{s,a} \sum_{k\notin\Kexp(s,h)} \frac{ \indevent{s_{h}^{k}=s,a_{h}^{k}=a}}{\sqrt{1\vee\sum_{j=1}^{k-1} \indevent{s_{h}^{j}=s,a_{h}^{j}=a}}} \cdot \sqrt{ \frac{\sum_{j:j<k,j + d^{j}\ge k} \indevent{s_{h}^{j}=s,a_{h}^{j}=a}}{1\vee\sum_{j:j + d^{j}\le k-1} \indevent{s_{h}^{j}=s,a_{h}^{j}=a}}}
    \\
    & =
    \underbrace{
    \sum_{s,a}\sum_{\substack{k\notin\Kexp(s,h) \\ k\in\Kcal(s,a,h)}} \frac{ \indevent{s_{h}^{k}=s,a_{h}^{k}=a}}{\sqrt{1\vee\sum_{j=1}^{k-1} \indevent{s_{h}^{j}=s,a_{h}^{j}=a}}} \cdot \sqrt{ \frac{\sum_{j:j<k,j + d^{j}\ge k} \indevent{s_{h}^{j}=s,a_{h}^{j}=a}}{1\vee\sum_{j:j + d^{j}\le k-1} \indevent{s_{h}^{j}=s,a_{h}^{j}=a}}}}
    _{(D.2.1)}
    \\
    & + 
    \underbrace{
    \sum_{s,a}\sum_{\substack{k\notin\Kexp(s,h) \\ k\notin\Kcal(s,a,h)}} \frac{ \indevent{s_{h}^{k}=s,a_{h}^{k}=a}}{\sqrt{1\vee\sum_{j=1}^{k-1} \indevent{s_{h}^{j}=s,a_{h}^{j}=a}}} \cdot \sqrt{ \frac{\sum_{j:j<k,j + d^{j}\ge k} \indevent{s_{h}^{j}=s,a_{h}^{j}=a}}{1\vee\sum_{j:j + d^{j}\le k-1} \indevent{s_{h}^{j}=s,a_{h}^{j}=a}}}}
    _{(D.2.2)}.
    \end{align*}
    Now, under the good event ($\lnot F_{5}^{basic}$) we have that for any $k\notin\Kexp(s,h)$,
    \begin{align*}
        \sum_{j:j + d^{j}\le k-1} \indevent{s_{h}^{j}=s,a_{h}^{j}=a}
        = n_{h}^{k}(s,a)
        \geq \Omega\left( \frac{d_{max}}{A} \right).
    \end{align*}
    Also, deterministically we have $\sum_{j:j<k,j + d^{j}\ge k} \indevent{s_{h}^{j}=s,a_{h}^{j}=a}\leq d_{max}$.
    Hence,
    \begin{align*}
        (D.2.1) 
        \leq
        \sqrt{A}\sum_{s,a}\sum_{k\notin\Kexp(s,h),k\in\Kcal(s,a,h)} \frac{\indevent{s_{h}^{k}=s,a_{h}^{k}=a}} {\sqrt{1\vee\sum_{j=1}^{k-1}\indevent{s_{h}^{j}=s,a_{h}^{j}=a}}}
        \lesssim
        S A^{3/2} \sqrt{d_{max}},
    \end{align*}
    where the last inequality follows from the fact that for any time the nominator is $1$, the sum in the denominator has increased by 1 as well. Hence it is bounded by the sum $\sum_{i=1}^{2d_{max}}\frac{1}{\sqrt{i}}\leq O(\sqrt{d_{max}})$.
    For last, recall that \eqref{eq:observed-geq-unobserved} holds for all $k\notin \Kcal(s,a,h)$.
    Hence,
    \begin{align*}
    (D.2.2) & \leq
    \sum_{s,a}\sum_{k\notin\Kexp(s,h),k\notin\Kcal(s,a,h)} \frac{ \indevent{s_{h}^{k}=s,a_{h}^{k}=a}}{\sqrt{1\vee\sum_{j=1}^{k-1} \indevent{s_{h}^{j}=s,a_{h}^{j}=a}}}\\
    & \leq
    \sum_{s,a}
    \underbrace{
    \sum_{k=1}^K \frac{ \indevent{s_{h}^{k}=s,a_{h}^{k}=a}}{\sqrt{1\vee\sum_{j=1}^{k-1} \indevent{s_{h}^{j}=s,a_{h}^{j}=a}}}}
    _{\text{Term $(D.1)$ which we bounded before}}
    \lesssim
    \sqrt{SAK}.
    \end{align*}
\end{proof}

\begin{lemma}
    \label{lemma:indicator-shrink-not-sqrt-exp}
    It holds that
    \begin{align*}
        \sum_{s \in \mathcal{S}} \sum_{a \in \mathcal{A}} \sum_{h=1}^H \sum_{k\notin\Kexp} \frac{ \indevent{s_{h}^{k}=s,a_{h}^{k}=a}}{n_{h}^{k}(s,a)\vee1} 
        \lesssim
        H S A^2.
    \end{align*}
\end{lemma}

\begin{proof}
    For any $s,a$ and $h$,
    \begin{align*}
        \sum_{k\notin\Kexp} & \frac{ \indevent{s_{h}^{k}=s,a_{h}^{k}=a}}{n_{h}^{k}(s,a)\vee1}
        \le
        \\
        & \le
        \sum_{k\notin\Kexp(s,h)} \frac{ \indevent{s_{h}^{k}=s,a_{h}^{k}=a}}{1\vee\sum_{j=1}^{k-1} \indevent{s_{h}^{j}=s,a_{h}^{j}=a}} \cdot \frac{1\vee\sum_{j=1}^{k-1} \indevent{s_{h}^{j}=s,a_{h}^{j}=a}}{1\vee\sum_{j:j + d^{j}\le k-1} \indevent{s_{h}^{j}=s,a_{h}^{j}=a}}
        \\
        & \leq
        \sum_{k\notin\Kexp(s,h)} \frac{ \indevent{s_{h}^{k}=s,a_{h}^{k}=a}}{1\vee\sum_{j=1}^{k-1} \indevent{s_{h}^{j}=s,a_{h}^{j}=a}} \cdot \Bigl( 2 +  \frac{\sum_{j:j<k,j + d^{j}\ge k} \indevent{s_{h}^{j}=s,a_{h}^{j}=a}}{1\vee\sum_{j:j + d^{j}\le k-1} \indevent{s_{h}^{j}=s,a_{h}^{j}=a}} \Bigr)
        \\
        & \le
        \sum_{k\notin\Kexp(s,h)} \frac{ \indevent{s_{h}^{k}=s,a_{h}^{k}=a}}{1\vee\sum_{j=1}^{k-1} \indevent{s_{h}^{j}=s,a_{h}^{j}=a}} \cdot \Bigl( 2 +  \frac{d_{max}}{d_{max} / A} \Bigr)
        \\
        & \le
        3 A \sum_{k\notin\Kexp(s,h)} \frac{ \indevent{s_{h}^{k}=s,a_{h}^{k}=a}}{1\vee\sum_{j=1}^{k-1} \indevent{s_{h}^{j}=s,a_{h}^{j}=a}}
        \lesssim
        A \log (KH),
    \end{align*}
    where we used the fact that $\sum_{j:j<k,j + d^{j}\ge k} \indevent{s_{h}^{j}=s,a_{h}^{j}=a}$ is always bounded by $d_{max}$, and that $\sum_{j:j + d^{j}\le k-1} \indevent{s_{h}^{j}=s,a_{h}^{j}=a}$ is at least $\Omega(\nicefrac{d_{max}}{A})$ for episodes that are not in $\Kexp$ under the good event.
    The last inequality follows from standard arguments \citep[Lemma B.18]{cohen2020ssp}, since there is no delay involved in this sum.
\end{proof}

\subsubsection{Bounding Terms (B) and (C) under full-information feedback}
\label{sec:bound-term-b-c-full}

Following the analysis of term (B.2) in the proof of \Cref{lemma:OMD}, since we run the policy improvement step only over $[K] \backslash \Kexp$, and since under full-information $V_{h}^{k} \leq H$ and the costs are bounded by $1$,
\begin{align*}
    (B.2)
    & \leq
    \eta H^2(K+D).
\end{align*}
Hence, term $(B)$ can be bounded by,
\begin{align}
    (B) 
    & \leq
    \frac{H\log A}{\eta}+\eta H^{3}(D+K).
    \label{eq:OMD-full}
\end{align}
For last, term $(C.1)$ is now zero, and so
\begin{align}
    (C) & \leq 0.
    \label{eq:optimisem-full}
\end{align}

\subsection{Proof of \cref{thm:regret-bound-unknown-p-non-delayed-traj}}   \label{appendix:no-trajectory-delay}

When the trajectories are observed without delay, Term (A) is bounded similarly to \citet{efroni2020optimistic} since it is no longer affected by the delay.
Moreover, since there is no explicit exploration we also do not have the extra $H^2 S d_{max}$ factor.
Terms (B) and (C) remain unchanged.

Thus, with bandit feedback, we obtain the regret bound
\begin{align*}
    \regret
    =
    \wt O \Bigl( H^2 S \sqrt{A K} + H^2 S^2 A + \gamma K H S A + \frac{H}{\eta} + \frac{\eta}{\gamma} H^3 (K + D) + \frac{H}{\gamma} + \frac{\eta}{\gamma^2} H^3 d_{max} \Bigr),
\end{align*}
and choosing $\eta = \frac{1}{H(A^{3/2}K + D)^{2/3}}$ and $\gamma = \frac{1}{(A^{3/2}K + D)^{1/3}}$ gives the theorem's statement.

Similarly, with full-information feedback, we obtain the regret bound
\[
    \regret
    =
    \wt O \Bigl( H^{3/2} S \sqrt{A K} + H^2 S^2 A + \frac{H}{\eta} + \eta H^3 (K + D) \Bigr)
\]
and choosing $\eta = \frac{1}{H\sqrt{K + D}}$ gives the theorem's statement.

We note that \citet{efroni2020optimistic} bound Term (A) by $\wt O ( H^2 S \sqrt{A K} )$, but with full-information feedback it can actually be bounded by $\wt O ( H^{3/2} S \sqrt{A K} )$.
This is obtained by known Bernstein-based confidence bounds analysis \citep{azar2017minimax,zanette2019tighter}.
For example, one can follow Lemmas 4.6,4.7,4.8 of \citet{cohen2020ssp} which are more general.

The reason that this bound (that improves by a factor of $\sqrt{H}$) does not hold with delayed trajectory feedback is \cref{lemma:indicator-shrink-not-sqrt-exp} where we bound $\sum_{s \in \mathcal{S}} \sum_{a \in \mathcal{A}} \sum_{h=1}^H \sum_{k\notin\Kexp} \frac{ \indevent{s_{h}^{k}=s,a_{h}^{k}=a}}{n_{h}^{k}(s,a)\vee1}$ by $\wt O (H S A^2)$ instead of $\wt O (H S A)$ when the trajectory feedback is not delayed.
Thus, the analysis of \citet{cohen2020ssp} gets a bound of $\wt O (H^{3/2} S A \sqrt{K})$ instead of $\wt O (H^{3/2} S \sqrt{A K})$, since it bounds Term (A) roughly by
\[
    H \sqrt{S K} \sqrt{\sum_{s \in \mathcal{S}} \sum_{a \in \mathcal{A}} \sum_{h=1}^H \sum_{k\notin\Kexp} \frac{ \indevent{s_{h}^{k}=s,a_{h}^{k}=a}}{n_{h}^{k}(s,a)\vee1}}.
\]

\subsection{Proof of \cref{thm:regret-bound-known-p}}        
\label{appendix:known-dynamics}

When dynamics are known, we use the actual transition function instead of the estimated one. 
Under the bandit-feedback, the terms $(A.2)$, $(A.3)$ in \cref{appendix:basic-bandit-proof} become zero. 
Since we use the actual occupancy measure of the policy (and do not compute it using some transition function from the confidence set), Term $(A.1.1)$ is now bounded by  $\gamma KHSA$. 
Term $(A.2.1)$ remains unchanged. 
Therefore,
\begin{align*}
    (A) 
    \lesssim
    \gamma KHSA + H\sqrt{K}.
\end{align*}     
Term $(B)$ remains unchanged and Term $(C)$ is now bounded by $\tilde{O}(H / \gamma)$, as (C.2) zeroes.

Thus, with known transition function and bandit feedback, we obtain the regret bound
\begin{align*}
    \regret
    =
    \wt O \Bigl( H \sqrt{K} + \gamma K H S A + \frac{H}{\eta} + \frac{\eta}{\gamma} H^3 (K + D) + \frac{H}{\gamma} + \frac{\eta}{\gamma^2} H^3 d_{max} \Bigr),
\end{align*}
and choosing $\eta = \frac{1}{H(A^{3/2}K + D)^{2/3}}$ and $\gamma = \frac{1}{(A^{3/2}K + D)^{1/3}}$ gives the theorem's statement.

Similarly, with known transition function and full-information feedback (the only non-zero term now is Term (B)), we obtain the regret bound
\[
    \regret
    =
    \wt O \Bigl( \frac{H}{\eta} + \eta H^3 (K + D) \Bigr)
\]
and choosing $\eta = \frac{1}{H\sqrt{K + D}}$ gives the theorem's statement.

\newpage

\section{Skipping scheme for handling large delays}
\label{appendix:skipping}

In this section we show that by skipping episodes with large delays, we can substitute the $d_{max}$ term in \cref{thm:regret-bound-unknown-p-non-delayed-traj,thm:regret-bound-unknown-p-delayed-traj,thm:regret-bound-known-p} by $\sqrt{D}$. 
This was presented by \citet{thune2019nonstochastic} for MAB with delays and can be easily applied to our setting as well.
The idea is to skip episodes with delay larger than $\sqrt{D}$ and bound the regret on skipped episodes trivially by $H$. 
That way, effectively the maximum delay is $\sqrt{D}$ and the number of skipped episodes is at most $\sqrt{D}$ as well. 
The skipping scheme can be generalized for arbitrary threshold as presented in \Cref{alg:skipper}.

\begin{algorithm}
    \caption{Skipping Wrapper}      \label{alg:skipper}
    \begin{algorithmic}
        \STATE \textbf{Input:} Algorithm $ALG$,  Skipping threshold $\beta > 0$.
        \FOR{$k = 1, 2, 3,...$}
            \STATE Get policy $\pi^k$ from $ALG$ and play the $k$-th episode with $\pi^k$.
            \STATE Observe feedback from all episodes in $\mathcal{F}^k = \{j:j+d^j = k\}$. 
            \STATE Feed $ALG$ all episodes $j\in \mathcal{F}^k$ such that $d^j \leq \beta$.
        \ENDFOR
    \end{algorithmic}
\end{algorithm}

\begin{lemma}       \label{lemma:skipper-bound}
    Assume that we have a regret bound for $ALG$ that depends on the number of episodes, the sum of delays and the maximum delay: $R^{ALG}(K,D,d_{max})$. Assume also that the $ALG$ choices depend only on the feedback.
    Then the regret of \Cref{alg:skipper} when simulating $ALG$ with a threshold $\beta > 0$ is at most,
    \begin{align*}
        R^{ALG}(|\mathcal{K}_{\beta}|, D_{\beta},\beta) + H(K - |\mathcal{K}_{\beta}|)
    \end{align*}
    where $\mathcal{K}_{\beta} = \{k:d^k \leq \beta\}$ and $D_{\beta} = \sum_{k\in \mathcal{K}_{\beta}} d^k$.
\end{lemma}

\begin{proof}
    Fix some threshold $\beta$ and a policy $\pi$. 
    \begin{align*}
        \sum_{k=1}^{K}V_{1}^{\pi^{k}}(s_1^k)-V_{1}^{\pi}(s_1^k) 
        & =
        \sum_{k\in\mathcal{K}_{\beta}} V_{1}^{\pi^{k}}(s_1^k)-V_{1}^{\pi}(s_1^k)
        +\sum_{k\notin\mathcal{K}_{\beta}} V_{1}^{\pi^{k}}(s_1^k)-V_{1}^{\pi}(s_1^k).
    \end{align*}
    Since the the algorithm policies $\pi^k$ are affected only by feedback from $\mathcal{K}_{\beta}$,
    and the total delay on those rounds is $D_{\beta}$, the first sum
    is bounded by $R^{ALG}(|\mathcal{K}_{\beta}|,D_{\beta},\beta)$.
    
    Since the value function is bounded by $H$, the second sum is bounded
    by $H(K - |\mathcal{K}_{\beta}|)$.
\end{proof}

\begin{remark}
    The proof of \Cref{lemma:skipper-bound} relies on the fact that the algorithm does not observe feedback outside of $\mathcal{K}_{\beta}$. However, if the trajectory feedback is available immediately at the end of the episode, we can also feed the algorithm with trajectory feedback outside of $\mathcal{K}_{\beta}$. This can only shrink the confidence intervals and reduce the regret.
\end{remark}

\begin{lemma}       \label{lemma:num-skipped-bound}
    For any threshold $\beta > 0$, the number of skipped rounds under \Cref{alg:skipper} is bounded by $K - \mathcal{K}_{\beta} < \frac{D}{\beta}$.
\end{lemma}

\begin{proof}
    First note that,
    \begin{align*}
        K-\mathcal{K}_{\beta} 
        & =
        \sum_{k=1}^{K}\left( 1-\ind\{d^{k}\leq\beta\} \right)
        =
        \sum_{k=1}^{K}\ind\{d^{k}>\beta\}.
    \end{align*}
    Now, we can bound the sum of delays by,
    \begin{align*}
        D 
        & \geq
        \sum_{k=1}^{K}d^{k} \ind\{d^{k}>\beta\}
        >
        \sum_{k=1}^{K}\beta\ind\{d^{k}>\beta\}
        =
        \beta\left( K-\mathcal{K}_{\beta} \right).
    \end{align*}
    Dividing both sides by $\beta$ completes the proof.
\end{proof}

Using the last two lemmas and the regret guarantees that we show in previous sections, we can now deduce regret bounds for delayed OPPO, when simulated by \Cref{alg:skipper}.
In some settings there was no dependence on $d_{max}$, and thus no skipping is needed.

\begin{corollary}   \label{cor:skiping-regret}
    Running Delayed OPPO, results in the following regret bounds (with probability at least $1 - \delta$):
    \begin{itemize}
        \item Under bandit feedback, known dynamics, and with threshold $\beta = \sqrt{\nicefrac{D}{HS}}$,
        \begin{align*}
            \regret
            =
            \wt{O}\left( 
            HS\sqrt{A}K^{2/3} 
            + H^2 D^{2/3}
            \right).
        \end{align*}
        \item Under bandit feedback, unknown dynamics, non-delayed trajectory feedback, and with threshold $\beta = \sqrt{\nicefrac{D}{HS}}$,
        \begin{align*}
            \regret
            =
            \wt{O}\left( 
            HS\sqrt{A}K^{2/3} 
            + H^2 D^{2/3}
            + H^2 S^2 A
            \right).
        \end{align*}
        \item Under full-information feedback, unknown dynamics delayed trajectory feedback, and with threshold $\beta = \sqrt{\nicefrac{D}{HS}}$,
        \begin{align*}
            \regret
            =
            \wt{O}\left( 
            H^{2}S\sqrt{AK}
            + H^{3/2}\sqrt{SD}
            +H^{2}S^{2}A^{3}
            \right).
        \end{align*}
        \item Under bandit feedback, unknown dynamics, delayed trajectory feedback, and with threshold $\beta = \sqrt{\nicefrac{D}{HSA}}$,
        \begin{align*}
            \regret
            =
            \wt{O}\left( 
            HS\sqrt{A}K^{2/3}
            + H^2 D^{2/3}
            + H^2 S^2 A^3
            + S^3 A^3
            \right).
        \end{align*}
    \end{itemize}
\end{corollary}
\begin{proof}
    The first three regret bounds follow immediately from the regret bounds we show for Delayed OPPO, \cref{lemma:skipper-bound} and \cref{lemma:num-skipped-bound}. For the last bound, we directly get a bound of,
    \begin{align*}
        \tilde{O}\left( 
        HS\sqrt{A}K^{2/3}
        + H^2 D^{2/3}
        + H^{3/2} \sqrt{S A D}
        + H^2 S^2 A^3
        + S^3 A^3
        \right).
    \end{align*}
    Note that if the third term dominates over the second, then $D \leq (\nicefrac{SA}{H})^3$, which implies that 
    \begin{align*}
        H^{2/3} \sqrt{S D} 
        \leq
        S^3 A^3.
    \end{align*}
\end{proof}

\newpage

\section{Doubling trick for handling unknown number of episodes and total delay}
\label{appendix:doubling}





\Cref{alg:OPPO-doubling} for unknown $D$ and $K$, uses the well-known doubling trick. 
Unlike \cite{bistritz2019online} that aim to estimate $K+D$, we aim to estimate the value $(A^{3/2}K + D)$ under bandit feedback which is the core difference that allows us to obtain the exact same bounds as the case where $K$ and $D$ are known. In fact, using the proper tuning and our analysis, one can achieve optimal regret in Delayed MAB using a doubling trick (see \cref{remark:doubling-improve}). The second difference is that we incorporate the skipping scheme and tune the threshold using our estimate which allows us to preserve the merits of the skipping procedure. 

On the technical side, the proof of \cref{thm:doubling-bandit-knwon-dynamics} adopts some ideas from \cite{bistritz2019online}, but requires more delicate analysis, especially when handling term $(i)$ in the proof. 

Denote by $M^k$ the number of missing samples at the end of episode $k$. That is, $M^k = k - \sum_{j=1}^k |\mathcal{F}^j|$.At time $k$ we estimate the value of $(A^{3/2}K + D)$ by $A^{3/2}k+\sum_{j=1}^{k}M^j$, and initializes a new phase whenever the estimation doubles itself.
Note that this is an optimistic estimation of $(A^{3/2}K + D)$.
If the feedback from episode $j$ arrived, then we estimate its delay exactly by $d^j$.
However, if the feedback did not arrive, we estimate it as if the feedback will arrive in the next episode.

\begin{algorithm}
    \caption{Delayed OPPO with known transition function, bandit feedback and unknown $D$ and $K$}  
    \label{alg:OPPO-doubling}
    \begin{algorithmic}
        \STATE \textbf{Input:} State space $\mathcal{S}$, Action space $\mathcal{A}$, Horizon $H$, Transition function $p$.
        
        \STATE \textbf{Initialization:} 
        Set $\pi_{h}^{1}(a \mid s) = \nicefrac{1}{A}$ for every $(s,a,h) \in \mathcal{S} \times \mathcal{A} \times [H]$, $e = 1$, $\eta_{e} = H^{-1} 2^{- 2e / 3}$, $\gamma_{e} = 2^{- e / 3 }$, $\beta_{e} = 2^{e / 2 }$.
        
        \FOR{$k=1,2,\dots,K$}
            
            \STATE Play episode $k$ with policy $\pi^k$.
            
            \STATE Observe feedback from all episodes $j \in \mathcal{F}^k$.
            
            \STATE {\color{gray} \# Policy Evaluation}
            
            \FOR{$j\in \mathcal{F}^k$ such that $d^j \le \beta_e$}
            
                \STATE $\forall s \in \mathcal{S} : V_{H+1}^j(s)=0$.
                
                \FOR{$h = H,\dots,1$ and $(s,a) \in \mathcal{S} \times \mathcal{A}$}
                    
                    \STATE $\hat c^j_h(s,a) = \frac{c_h^j(s,a) \cdot \mathbb{I} \{ s_h^j=s,a_h^j=a \}}{q^{p,\pi^j}_h(s) \pi^j_h(a \mid s) + \gamma_e}$.
                    
                    \STATE $Q_{h}^{j}(s,a) = \hat{c}_{h}^{j}(s,a)+\langle p_{h}(\cdot\mid s,a),V_{h+1}^{j} \rangle$.
                    
                    \STATE $V_{h}^{j}(s) = \langle Q_{h}^{j}(s,\cdot),\pi_{h}^{j}(\cdot \mid s)\rangle$.
                
                \ENDFOR
            
            \ENDFOR
            
            \STATE {\color{gray} \# Policy Improvement}
            
            \FOR{$(s,a,h) \in \mathcal{S} \times \mathcal{A} \times [H]$}
            
                \STATE $\pi^{k+1}_h(a | s) = \frac{\pi^k_h(a \mid s) \exp (-\eta_e \sum_{j\in\mathcal{F}^{k} : d^j \le \beta_e} Q_{h}^{j}(s,a))}{\sum_{a' \in \mathcal{A}} \pi^k_h(a' \mid s) \exp (-\eta_e \sum_{j\in\mathcal{F}^{k} : d^j \le \beta_e} Q_{h}^{j}(s,a'))}$.
                
            \ENDFOR
            
            \STATE {\color{gray} \# Doubling}
            
            \IF{$A^{3/2}k + \sum_{j=1}^k M^j = A^{3/2}k + \sum_{j=1}^k (j - \sum_{i=1}^j |\mathcal{F}^i|) > 2^e$}
            
                \STATE Set $\pi_{h}^{1}(a \mid s) = \nicefrac{1}{A}$ for every $(s,a,h) \in \mathcal{S} \times \mathcal{A} \times [H]$, $e = e + 1$, $\eta_{e} = H^{-1} 2^{- 2e / 3}$, $\gamma_{e} = 2^{- e / 3 }$, $\beta_{e} = 2^{e / 2 }$.
            
            \ENDIF
            
        \ENDFOR
        
    \end{algorithmic}
\end{algorithm}


\begin{theorem}     \label{thm:doubling-bandit-knwon-dynamics}
    Under bandit feedback, if the transition function is known, then (with probability at least $1 - \delta$), the regret of \cref{alg:OPPO-doubling} is bounded by,
    \begin{align*}
         \regret
         & \leq
         \tilde{O}\Big( 
         HS\sqrt{A}K^{2/3} 
         + H^{2}D^{2/3}
         \Big).
    \end{align*}
\end{theorem}

\begin{proof}
    Let $\mathcal{K}_{e}$ be the episodes in phase $e$, $\Mcal'_{e} = \{k\in \mathcal{K}_{e}: k+d^k\notin \mathcal{K}_{e} \}$, $\Mcal_{\beta_{e}} = \{k\in \mathcal{K}_{e}: d^k \geq \beta_{e} \}$, and $\Mcal_{e} = \Mcal'_{e} \cup \Mcal_{\beta_{e}}$ which is the set of episodes with missing feedback in phase $e$.
    Also, denote the last episode of phase $e$ by $K_e$. 
    Using \cref{thm:regret-bound-known-p} and \Cref{lemma:skipper-bound}.
    \begin{align}
        \nonumber
         \sum_{k \in \Kcal_{e}}V_{1}^{\pi^{k}}(s_1^k)-V_{1}^{\pi}(s_1^k)
         & =
         \tilde{O}\Big( 
         \gamma_e |\Kcal_{e}|HSA
         + \frac{H}{\eta_e} 
         + \frac{\eta_e}{\gamma_e}H^{3}\sum_{k\in \Kcal_e \backslash \Mcal_e }(1+d^k) 
         \\
         & \qquad \quad
         + \frac{H}{\gamma_e}
         + H^3 \frac{\eta_e}{\gamma_{e}^{2}} \beta_e
         + H \sqrt{|\Kcal_e|}
         + H|\Mcal_e|
         \Big).
         \label{eq:doubling-phase-regret}
    \end{align}
    By definition of our doubling scheme,
    \begin{align}
        \nonumber
        2^{e} & \geq A^{3/2} K_{e} + \sum_{k=1}^{K_{e}} M^{k}\\
        \nonumber
        & =
        A^{3/2} K_{e} + \sum_{k=1}^{K_{e}} \sum_{j=1}^{K} \ind\{j\leq k,j+d^{j}>k\}\\
        \nonumber
        & =
        A^{3/2} K_{e} + \sum_{j=1}^{K} \sum_{k=1}^{K_{e}} \ind\{j\leq k,j+d^{j}>k\}\\
        \nonumber
        & \geq 
        K_{e} + \sum_{j\in\Kcal_{e} \backslash \Mcal_{e}} \sum_{k=1}^{K_{e}} \ind\{j\leq k,j+d^{j}>k\}\\
        \nonumber
        & \underset{(*)}{\geq}
        |\Kcal_{e}| + \sum_{j\in\Kcal_{e} \backslash \Mcal_{e}} d^{j}\\
        & \geq
        \sum_{j\in\Kcal_{e} \backslash \Mcal_{e}} (d^{j}+1),
        \label{eq:doubling-D-phase-bound}
    \end{align}
    where $(*)$ holds since $j+d^{j}\leq K_{e}$ for any $j\in\Kcal_{e} \backslash \Mcal_{e}$.
    We now bound $|\Mcal'_{e}|$. Similarly to the above,
    \begin{align*}
        2^{e} & \geq\sum_{j=1}^{K}\sum_{k=1}^{K_{e}}\ind\{j\leq k,j+d^{j}>k\}\\
        & \geq
        \sum_{j\in\Mcal'_{e}} \sum_{k=K_{e-1}+1}^{K_{e}} \ind\{j\leq k,j+d^{j}>k\}\\
        & \underset{(*)}{=}
        \sum_{j\in\Mcal'_{e}}K_{e}-j+1\\
        & \underset{(**)}{\geq}
        \sum_{j=K_{e}-|\Mcal'_{e}|}^{K_{e}}K_{e}-j+1\\
        & \geq
        \sum_{j=1}^{|\Mcal'_{e}|+1}j\\
        & =
        \frac{1}{2}(|\Mcal'_{e}|+1)(|\Mcal'_{e}|+2)\\
        & \geq
        \frac{1}{2}|\Mcal'_{e}|^{2},
    \end{align*}
    where $(*)$ follows from the fact that for any $k\in \Kcal_{e}$ we have that $j+d^j > k$, and $(**)$ follows by choosing the 
    largest possible $|\Mcal'_{e}|$ indices. 
    The above implies that,
    \begin{align}
        |\Mcal'_{e}|
        \leq 
        2^{\frac{e+1}{2}}.       
        \label{eq:doubling-proof-missing-bound}
    \end{align}
    For last, we bound $|\Mcal_{\beta_{e}} \setminus \Mcal'_{e}|$,
    \begin{align*}
        2^{e} 
        & \geq
        \sum_{j=1}^{K}\sum_{k=1}^{K_{e}}\ind\{j\leq k,j+d^{j}>k\}\\
        & \geq
        \sum_{j\in\Mcal_{\beta_{e}} \setminus \Mcal'_{e}} \sum_{k=1}^{K_{e}}\ind\{j\leq k,j+d^{j}>k\}\\
        & \underset{(*)}{=}
        \sum_{j\in\Mcal_{\beta_{e}} \setminus \Mcal'_{e}} d^j\\
        & \underset{(**)}{\ge}
        \sum_{j\in\Mcal_{\beta_{e}} \setminus \Mcal'_{e}}\beta_{e}\\
        & =
        |\Mcal_{\beta_{e}} \setminus \Mcal'_{e}|\beta_{e},
    \end{align*}
    where $(*)$ follows because $j\notin \Mcal_{e}'$ and so $j+d^j \leq K_e$. And $(**)$ follows the definition of $\Mcal_{\beta_e}$
    This implies that,
    \begin{align*}
        |\Mcal_{\beta_{e}} \setminus \Mcal'_{e}| 
        \leq
        2^{\frac{e}{2}}.
    \end{align*}
    Combining with \eqref{eq:doubling-proof-missing-bound} gives us,
    \begin{align}
        |\Mcal_e|
        =
        |\Mcal'_{e} \cup \Mcal_{\beta_{e}}|
        \leq
        2^{\frac{e+3}{2}}.
        \label{eq:doublind-missing-bound2}
    \end{align}
    Plugging the bounds of \eqref{eq:doubling-D-phase-bound} and \eqref{eq:doublind-missing-bound2} into \eqref{eq:doubling-phase-regret}, noting that $|\Kcal_e| \leq 2^e$, plugging the values of $\eta_e$, $\gamma_e$ and $\beta_e$, and summing over all phases gives us,
    \begin{align*}
        \regret 
        \leq 
        \tilde{O} \Biggl(
        HSA \underbrace{\sum_{e=1}^{E}|\Kcal_e|2^{-\frac{e}{3}}}_{(i)}
        + H^2 \underbrace{\sum_{e=1}^{E} 2^\frac{2e}{3}}_{(ii)}
        \Biggr),
    \end{align*}
    where $E \leq \log(A^{3/2} K + D)$ is the number of phases. Term $(ii)$ above is a sum of geometric series and can be bounded by $O\bigl( (D + A^{3/2}K)^{2/3} \bigr) \leq O\bigl( D^{2/3} + S K^{2/3} \bigr)$ since $A \leq S$.
    Term $(i)$ is bounded by the value of the following maximization problem,
    \begin{align*}
        & \max\sum_{e=1}^{E}|\Kcal_{e}|2^{-\frac{e}{3}}\\
        \text{subject to } 
        & A^{3/2}|\Kcal_{e}|\leq2^{e}, \sum_{e=1}^{E}|\Kcal_{e}|=K,
    \end{align*}
    which is necessarily bounded by
    \begin{align*}
        & \max \sum_{e=1}^{\infty}x_{e}2^{-\frac{e}{3}}\\
        \text{subject to } 
        & 0 \leq x_{e}\leq \frac{2^{e}}{A^{3/2}}, \sum_{e=1}^{\infty}x_{e}=K.
    \end{align*}
    Since $2^{-\frac{e}{3}}$ is decreasing, this is maximized whenever
    the first $x_{e}$s are at maximum value. There are at most $\lceil \log (KA^{3/2})\rceil $
    non-zero $x_{e}$s and so,
    \begin{align*}
        \sum_{e=1}^{E} |\Kcal_{e}|2^{-\frac{e}{3}} 
        & \leq
        \frac{1}{A^{3/2}}\sum_{e=1}^{\lceil \log (KA^{3/2})\rceil }2^{\frac{2e}{3}}
        \leq
        O \left( \frac{1}{A^{3/2}} (A^{3/2}K)^\frac{2}{3} \right)
        \leq O \left( \frac{1}{\sqrt{A}} K^\frac{2}{3} \right),
    \end{align*}
    which gives us the desired bound.
\end{proof}

\begin{remark}
    The exact same proof holds for the rest of settings, in which we get the exact same regret as in \Cref{cor:skiping-regret},  \Cref{thm:regret-bound-unknown-p-non-delayed-traj} and \Cref{thm:regret-bound-known-p}. For full information, one should set the doubling condition to be $k+\sum_{j=1}^k M^j > 2^e$, and $\eta_e = H^{-1} 2^{-e/2}$.
    
\end{remark}

\begin{remark}  \label{remark:doubling-improve}
    \citet{bistritz2019online} set, for delayed MAB, the doubling condition to be $k+\sum_{j=1}^k M^j > 2^e$ which results in sub-optimal $\tilde O\bigl( A\sqrt{K} + \sqrt{D}\bigr)$ regret, instead of $\tilde O\bigl( \sqrt{AK} + \sqrt{D}\bigr)$. Changing the condition to be $Ak + \sum_{j=1}^k M^j > 2^e$ and following the new ideas in our analysis leads to the optimal $\tilde O\bigl( \sqrt{AK} + \sqrt{D}\bigr)$ regret.
    Interestingly, our analysis shows that the doubling condition of \citet{bistritz2019online} is appropriate for full-information feedback but not to bandit feedback, where our new condition is crucial in order to get optimal regret.
\end{remark}

\newpage

\section{Delayed O-REPS}
\label{appendix:delayed-o-reps}

Given an MDP $\mathcal{M}$, a policy $\pi$ induces an occupancy measure $q = q^\pi$, which satisfies the following:

\begin{align}
    \sum_{a,s'} q_{1}(s_\text{init},a,s') 
    & = 1
    & 
    \label{eq:o-measure-starts-s0}
    \\
    \sum_{s,a,s'} q_{h}(s,a,s')
    & = 1
    & \forall h=1,\dots,H
    \label{eq:o-meausre-sum1}
    \\
    \sum_{s',a}q_{h}(s,a,s')
    & = \sum_{s',a} q_{h-1}(s',a,s)
    & \forall s \in \mathcal{S} \text{ and } h=2,\dots,H
    \label{eq:o-meausre-in-eq-out}
\end{align}
If $q$ satisfies \eqref{eq:o-measure-starts-s0}, \eqref{eq:o-meausre-sum1} and \eqref{eq:o-meausre-in-eq-out}, then it induces a policy and a transition function in the following way:
\begin{align*}
    \pi_{h}^{q}(a \mid s)
    & =
    \frac{\sum_{s'}q_{h}(s,a,s')}{\sum_{a'}\sum_{s'}q_{h}(s,a',s')}\\
    p^{q}_{h}(s'\mid s,a) 
    & =
    \frac{q(s',a,s)}{\sum_{s,a} q(s',a,s)}
\end{align*}
The next lemma characterize the occupancy measures induced by some policy $\pi$.

\begin{lemma}[Lemma 3.1, \citet{rosenberg2019full}]
    An occupancy measure $q$ that satisfies \eqref{eq:o-measure-starts-s0}, \eqref{eq:o-meausre-sum1} and \eqref{eq:o-meausre-in-eq-out} is induced by some policy $\pi$ \textbf{if and only if} $p^{q} = p$.
\end{lemma}
\begin{definition}
    Given an MDP $\mathcal{M}$, we define $\Delta(\mathcal{M})$  to be the set of all $q\in [0,1]^{H \times S \times A \times S}$ that satisfies \eqref{eq:o-measure-starts-s0}, \eqref{eq:o-meausre-sum1} and \eqref{eq:o-meausre-in-eq-out} such that $p^q = p$ (where $p$ is the transition function of $\mathcal{M}$).
\end{definition}

For convenience, in this section we let the cost functions to be a function of the current state, the action taken and the next state: $c_{h}(s,a,s')$. So the value of a policy, $\pi$ is given by 
\begin{align*}
    V^\pi(s_\text{init}) 
    = \langle q^{\pi},c\rangle 
    = \sum_{h} \langle q_{h}^{\pi},c_{h}\rangle 
\end{align*}
and the regret with respect to a policy $\pi$ is given by
\begin{align*}
    \regret = \sum_{k=1}^{K} \langle q^{\pi^{k}}-q^{\pi} , c^{k} \rangle.
\end{align*}
The above can be treated as an online linear optimization problem. Indeed, O-REPS \cite{zimin2013online} treats it as such by running Online Mirror Decent (OMD) on the set of occupancy measures. That is, at each episode the algorithm plays the policy induced by the occupancy measure $q^k$ and updates the occupancy measure for the next episode by,
\begin{align*}
    q^{k+1}
    =
    \arg\min_{q\in\Delta(\mathcal{M})}\{ \eta\langle q,c^{k}\rangle +D_{R}(q\Vert q^{k})\},
\end{align*}
where is $R$ the unnormalized negative entropy. That is,
\[
    R(q)
    =
    \sum_{h}\sum_{s,a,s'} q_{h}(s,a,s')\log q_{h}(s,a,s') - q_h(s,a,s')
\]
and $D_R$ is the Bregman divergence associated with $R$ ($D_R$ is also known as the Kullback-Leibler divergence). If the feedback is delayed we would update the occupancy measure using all the feedback that arrives at the end of the current episode, i.e., 
\begin{align*}
    q^{k+1}
    =
    \arg\min_{q\in\Delta(\mathcal{M})}\Big\{ \eta \Big\langle q, \sum_{j\in\mathcal{F}^{k}}c^{j} \Big\rangle +D_{R}(q\Vert q^{k}) \Big\}.
\end{align*}
Whenever the transition function is unknown, $\Delta(\mathcal{M})$ can not be computed. In this case we adopt the method of \citet{rosenberg2019full} and extend $\Delta(\mathcal{M})$ by the next definition.
\begin{definition}
    For any $k \in [K]$, we define $\Delta(\mathcal{M},k)$ to be
    the set of all $q\in[0,1]^{H\times S\times A\times S}$
    that satisfies  \eqref{eq:o-measure-starts-s0}, \eqref{eq:o-meausre-sum1}, \eqref{eq:o-meausre-in-eq-out}  and
    \begin{align*}
    \forall h,s',a,s:
    | p_{h}^{q}(s' \mid s,a)-\bar{p}_{h}^{k}(s' \mid s,a) | \leq\epsilon_{h}^{k}(s' \mid s,a).
    \end{align*}
\end{definition}

The update step will now be with respect to $\Delta(\mathcal{M},k)$. We have that with high probability $\Delta(\mathcal{M},k)$ contain $\Delta(\mathcal{M})$ for all $k$, and so the estimation of the value function is again optimistic.
Delayed O-REPS for unknown dynamics is presented in \Cref{alg:o-reps}.
\begin{algorithm}
    \caption{Delayed O-REPS} \label{alg:o-reps}
    \begin{algorithmic}
        \STATE \textbf{Input:} State space $\mathcal{S}$, Action space $\mathcal{A}$, Learning rate $\eta > 0$, Confidence parameter $\delta > 0$.
        \STATE \textbf{Initialization:} Set $\pi^{1}_{h}(a\mid s)=\nicefrac{1}{A}$, $q_{h}^{1}(s,a,s')=\nicefrac{1}{S^{2}A}$ for every $(s,a,s',h) \in \mathcal{S} \times \mathcal{A} \times \mathcal{S} \times [H]$.
        
        \FOR{$k=1,2,...,K$}
        
            \STATE Play episode $k$ with policy $\pi^k$.
            
            \STATE Observe feedback from all episodes $j\in \mathcal{F}^k$, and last trajectory $U^{k}=\{(s_{h}^{k},a_{h}^{k})\}_{h=1}^{H}$.
            
            \STATE Update transition function estimation.
            
            \STATE {\color{gray} \# Update Occupancy Measure}
            \IF{transition function is known}
            
                \STATE $q^{k+1}=\arg\min_{q\in\Delta(\mathcal{M})}\{ \eta\langle q,\sum_{j\in\mathcal{F}^{k}}c^{j}\rangle +D_{R}(q\Vert q^{k})\} $
            
            \ELSE
            
                \STATE $q^{k+1}=\arg\min_{q\in\Delta(\mathcal{M},k)}\{ \eta\langle q,\sum_{j\in\mathcal{F}^{k}}c^{j}\rangle +D_{R}(q\Vert q^{k})\} $

            \ENDIF
            \STATE {\color{gray} \# Update Policy}
            \STATE Set $\pi_{h}^{k+1}(a\mid s)
            =\frac{\sum_{s'}q_{h}^{k+1}(s,a,s')}{\sum_{a'}\sum_{s'}q_{h}^{k+1}(s,a',s')}$ for every $(s,a,h) \in \mathcal{S} \times \mathcal{A} \times [H]$.
        \ENDFOR
    \end{algorithmic}
    
\end{algorithm}

The update step can be implemented by first solving the unconstrained convex optimization problem,
\begin{align}    \label{eq:O-REPS-unconstrained-update}
    \tilde{q}^{k+1}
    =
    \arg\min
    \left\{ 
    \eta\left\langle q,\sum_{j\in\mathcal{F}^{k}}c^{j} \right\rangle + D_{R}(q\Vert q^{k})
    \right\},
\end{align}

and then projecting onto the set $\Delta(\mathcal{M},k)$ with respect to $D_{R}(\cdot\Vert \tilde{q}^{k+1})$. That is,
\[
    q^{k+1}
    =
    \arg\min_{q\in\Delta(\mathcal{M},k)}D_{R}(q\Vert\tilde{q}^{k+1}).
\]

The solution for \eqref{eq:O-REPS-unconstrained-update} is simply given by,
\[
    \tilde{q}_{h}^{k+1}(s,a,s')
    =
    q_{h}^{k}(s,a,s')e^{-\eta\sum_{j\in\mathcal{F}^{k}}c^{j}(s,a,s')}.
\]

\begin{theorem}
    \label{thm:regret-bound-o-reps}
    Running Delayed O-REPS under full-information feedback and delayed cost feedback guarantees the following regret, with probability at least $1 - \delta$,
    \[
        \regret
        =
        \wt O (H^{3/2} S \sqrt{A K} + H \sqrt{D} + H^2 S^2 A).
    \]
    Moreover, if the transition function is known, we obtain regret of
    \[
        \regret
        =
        \wt O ( H \sqrt{K+D}).
    \]
\end{theorem}

\subsection{Proof of \cref{thm:regret-bound-o-reps}}
Given a policy $\pi$ and a transition $p$, we denote the occupancy measure of $\pi$ with respect to $p$, by $q^{p,\pi}$. That is, $q^{p,\pi}_h(s,a,s') = \Pr[s_{h}^{k}=s, a_{h}^{k}=a, s_{h+1}^{k}=s'\mid s_1^k=s_\text{init},\pi,p]$. Also, denote $p^k = p^{q^k}$, and note that by definition $q^{p^k, \pi^k} = q^k$. We define the following good event $G = \lnot F_{1}^{basic}$ where $F_{1}^{basic}$ defined in \Cref{appendix:fail-events}. As shown in \Cref{lemma:basic-good-event}, $G$ occurs with probability of at least $1-\delta$. As consequence we have that for all episodes, $\Delta(\mathcal{M}) \subseteq \Delta(\mathcal{M},k)$. From this point we analyse the regret given that $G$ occurred.

We break the regret in the following way:

\begin{align}
    \sum_{k=1}^{K}\langle q^{p,\pi^{k}}-q^{p,\pi},c^{k}\rangle 
    & =
    \sum_{k=1}^{K}\langle q^{p,\pi^{k}}-q^{p^{k},\pi^{k}},c^{k}\rangle
    \nonumber
    +\sum_{k=1}^{K}\langle q^{p^{k},\pi^{k}}-q^{p,\pi},c^{k}\rangle \\
    & =
    \sum_{k=1}^{K}\langle q^{p,\pi^{k}}-q^{p^{k},\pi^{k}},c^{k}\rangle
    +\sum_{k=1}^{K}\langle q^k - q^{\pi},c^{k}\rangle.
    \label{eq:O-REPS-regret-decomposition}
\end{align}

The first term, under the good event, is bounded similarly as in the proof of \Cref{thm:regret-bound-unknown-p-non-delayed-traj} by,
\begin{align}
    \sum_{k=1}^{K}\langle q^{p,\pi^{k}}-q^{p^{k},\pi^{k}},c^{k}\rangle
    & \lesssim
    H^{3/2} S \sqrt{A K} + H^2 S^2 A.
    \label{eq:O-REPS-error-term}
\end{align}

For the second term, we adopt the approach of \citep{thune2019nonstochastic,bistritz2019online}, and modify the standard analysis of OMD. We start with \cref{lemma:standart-OMD-O-REP} which bounds the regret of playing $\pi^{k+d^k}$ at episode $k$.

\begin{lemma}   \label{lemma:standart-OMD-O-REP}
    If $\Delta(\mathcal{M})\subseteq\Delta(\mathcal{M},k)$ for all $k$, then for any $q \in \Delta(\mathcal{M})$, delayed O-REPS satisfies
    \begin{align*}
        \sum_{k=1}^{K}\langle c^{k},q^{k+d^{k}}-q\rangle  
        & \leq
        \frac{2H\log(HSA)}{\eta}+\eta HK.
    \end{align*}
\end{lemma}

\begin{proof}
    Note that $\tilde{q}_{h}^{k+1}(s,a,s')
    =q_{h}^{k}(s,a,s')e^{-\eta\sum_{j \in \Fcal^{k}}c_{h}^{j}(s,a,s')}$.
    Taking the log, 
    \[
        \eta\sum_{j \in \Fcal^{k}}c_{h}^{j}(s,a,s')
        = \log q_{h}^{k}(s,a,s')-\log\tilde{q}_{h}^{k+1}(s,a,s').
    \]
    Hence for any $q$ 
    \begin{align*}
        \eta\left\langle \sum_{j \in \Fcal^{k}}c_{h}^{j},q^{k}-q \right\rangle  
        & =
        \left\langle \log q^{k} - \log\tilde{q}^{k+1},q^{k}-q \right\rangle \\
        & =
        D_{R}(q||q^{k}) - D_{R}(q||\tilde{q}^{k+1}) + D_{R}(q^{k}||\tilde{q}^{k+1})\\
        & \leq 
        D_{R}(q||q^{k}) - D_{R}(q||q^{k+1}) - D_{R}(q^{k+1}||\tilde{q}^{k+1})+D_{R}(q^{k}||\tilde{q}^{k+1})\\
        & \leq 
        D_{R}(q||q^{k}) - D_{R}(q||q^{k+1}) + D_{R}(q^{k}||\tilde{q}^{k+1}),
    \end{align*}
    where the first equality follows directly the definition of Bregman divergence. The first inequality is by \citep[Lemma 1.2]{ziminonline} and
    the assumption that $\Delta(\mathcal{M})\subseteq\Delta(\mathcal{M},k)$.
    The second inequality is since Bregman divergence is non-negative.
    Now, the last term is bounded by, 
    \begin{align*}
        D_{R}(q^{k}||\tilde{q}^{k+1}) 
        & \leq 
        D_{R}(q^{k}||\tilde{q}^{k+1}) + D_{R}(\tilde{q}^{k+1}||q^{k})\\
        & =
        \sum_{h}\sum_{s,a,s'} \tilde{q}_{h}^{k+1}(s,a,s') \log\frac{\tilde{q}_{h}^{k+1}(s,a,s')}{q_{h}^{k}(s,a,s')} 
        \\
        & \qquad 
        + \sum_{h}\sum_{s,a,s'} q_{h}^{k}(s,a,s') \log\frac{q_{h}^{k}(s,a,s')}{\tilde{q}_{h}^{k+1}(s,a,s')}\\
        & =
        \langle q^{k}-\tilde{q}^{k+1},\log q^{k}-\log\tilde{q}^{k+1}\rangle \\
        & =
        \eta\biggl\langle q^{k}-\tilde{q}^{k+1},\sum_{j \in \Fcal^{k}}c^{j} 
        \biggr\rangle.
    \end{align*}
    We get that 
    \[
        \eta\left\langle \sum_{j \in \Fcal^{k}}c^{j},q^{k}-q \right\rangle 
        \leq 
        D_{R}(q||q^{k})-D_{R}(q||q^{k+1}) 
        + \eta\left\langle q^{k}-\tilde{q}^{k+1},\sum_{j \in \Fcal^{k}}c^{j}\right\rangle.
    \]
    Summing over $k$ and dividing by $\eta$, we get 
    \begin{align*}
        \underbrace{
        \sum_{k=1}^{K}\sum_{j \in \Fcal^{k}}\left\langle  c^{j},q^{k}-q \right\rangle}
        _{(*)}
        & \leq
        \frac{D_{R}(q||q^{1})-D_{R}(q||q^{K+1})}{\eta}
        +\sum_{k=1}^{K}\left\langle  q^{k}-\tilde{q}^{k+1},\sum_{j \in \Fcal^{k}}c^{j} \right\rangle \\
        & \leq
        \frac{D_{R}(q||q^{1})}{\eta}
        +\sum_{k=1}^{K}\left\langle  q^{k}-\tilde{q}^{k+1},\sum_{j \in \Fcal^{k}}c^{j} \right\rangle \\
        & \leq
        \frac{2H\log(SA)}{\eta}
        + \underbrace{
        \sum_{k=1}^{K}\left\langle  q^{k}-\tilde{q}^{k+1},\sum_{j \in \Fcal^{k}}c^{j} \right\rangle}
        _{(**)},
    \end{align*}
    where the last inequality is a standard argument (see \citep{ziminonline,hazan2019introduction}).
    We now
    rearrange $(*)$ and $(**)$: 
    \begin{align*}
        (*)
        & =
        \sum_{k=1}^{K}\sum_{j=1}^{K}\ind\{ j+d^{j}=k\} \langle c^{j},q^{k}-q\rangle \\
        & =
        \sum_{j=1}^{K}\sum_{k=1}^{K}\ind\{ j+d^{j}=k\} \langle c^{j},q^{k}-q\rangle \\
        & =
        \sum_{j=1}^{K}\langle c^{j},q^{j+d^{j}}-q\rangle \\
        & =
        \sum_{k=1}^{K}\langle c^{k},q^{k+d^{k}}-q\rangle.
    \end{align*}
    In a similar way, 
    \begin{align*}
        (**)  
        & =
        \sum_{k=1}^{K}\sum_{j \in \Fcal^{k}}\langle q^{k}-\tilde{q}^{k+1},c^{j}\rangle \\
        & =
        \sum_{k=1}^{K}\sum_{j=1}^{K}\ind\{ j \in \Fcal^{k}\} \langle q^{k}-\tilde{q}^{k+1},c^{j}\rangle \\
        & =
        \sum_{j=1}^{K}\sum_{k=1}^{K}\ind\{ j \in \Fcal^{k}\} \langle q^{k}-\tilde{q}^{k+1},c^{j}\rangle \\
        & =
        \sum_{k=1}^{K}\langle q^{k+d^{k}}-\tilde{q}^{k+d^{k}+1},c^{k}\rangle.
    \end{align*}
    This gives us, 
    \[
    \sum_{k=1}^{K}\langle c^{k},q^{k+d^{k}}-q\rangle \leq\frac{2H\log(S A)}{\eta}+\sum_{k=1}^{K}\langle q^{k+d^{k}}-\tilde{q}^{k+d^{k}+1},c^{k}\rangle.
    \]
    It remains to bound the second term on the right hand side:
    
    \begin{align*}
        \sum_{k}\langle & q^{k+d^{k}}-\tilde{q}^{k+d^{k}+1},c^{k}\rangle =
        \sum_{k}\sum_{h}\sum_{s,a,s'} c_{h}^{k}(s,a,s')(q_{h}^{k+d^{k}}(s,a,s')-\tilde{q}_{h}^{k+d^{k}+1}(s,a,s'))\\
        & =
        \sum_{k}\sum_{h}\sum_{s,a,s'} c_{h}^{k}(s,a,s') \left(
        q_{h}^{k+d^{k}}(s,a,s')-q_{h}^{k+d^{k}}(s,a,s')e^{-\eta\sum_{j \in \Fcal^{k}+d^{k}+1}c_{h}^{j}(s,a,s')} 
        \right)\\
        & \leq
        \sum_{k}\sum_{h}\sum_{s,a,s'}q_{h}^{k+d^{k}}(s,a,s') \left(
        1-e^{-\eta\sum_{j \in \Fcal^{k}+d^{k}+1}c_{h}^{j}(s,a,s')}
        \right)\\
        \tag{\ensuremath{1-e^{-x}\leq x}} 
        & \leq
        \eta\sum_{k}\sum_{h}\sum_{s,a,s'}q_{h}^{k+d^{k}}(s,a,s') \left(
        \sum_{j \in \Fcal^{k}+d^{k}+1}c_{h}^{j}(s,a,s')
        \right)\\
        \tag{\ensuremath{c_{h}^{k}(s,a,s') \leq 1}} 
        & \leq
        \eta\sum_{k}\sum_{h}\sum_{s,a,s'}q_{h}^{k+d^{k}}(s,a,s')\big| \mathcal{F}^{k+d^{k}+1} \big|\\
        & =
        \eta H\sum_{k} \big| \mathcal{F}^{k+d^{k}+1} \big|\\
        & \leq
        \eta HK.
    \end{align*}
    This completes the proof of the lemma.
\end{proof}
Using \Cref{lemma:standart-OMD-O-REP}, we can bounds the regret as,

\begin{align}
    \sum_{k=1}^{K}\langle c^{k},q^{k}-q\rangle  
    & \leq
    \frac{2H\log (HSA)}{\eta}
    +\eta HK
    +\sum_{k=1}^{K} \langle c^{k},q^{k}-q^{k+d^{k}}\rangle.
    \label{eq:O-REPS-regret-with-drift}
\end{align}
The next lemma is a generalization of Lemma 1 in \citep{zimin2013online}, which allows us to bound the distance between two consecutive occupancy measures.
\begin{lemma}   \label{lemma:adjacent-o-mesure-KL-bound}
For $q_{h}^{k}$ that are generated by delayed O-REPS, we have that,

    \begin{align*}
        \sum_{h}D_{R}(q_{h}^{k}\Vert q_{h}^{k+1}) 
        & \leq
        \sum_{h}\sum_{s,a,s'} q_{h}^{k}(s,a,s')\frac{(\eta\sum_{j\in\mathcal{F}^{k}}c_{h}^{j}(s,a,s'))^{2}}{2}.
    \end{align*}
\end{lemma}

\begin{proof}
    
First we present some notations.
Given $v_h,e_h:\Scal\times\Acal\times\Scal \to \mathbb{R},$ define 
\[
    B_{h}^{k}(s,a,s'\mid v,e)
    =
    e_h(s,a,s')
    + v_h(s,a,s')
    - \eta\sum_{j\in\mathcal{F}^{k}} c_{h}^{j}(s,a,s')
    - \sum_{s''} \bar p_{h}^{k}(s''\mid s,a)v_{h+1}(s,a,s'').
\]
Given $\beta_{h}:\Scal \to \mathbb{R}$ and $\mu_{h}^{+},\mu_{h}^{-}:\Scal\times\Acal\times\Scal \to \mathbb{R}_{\geq0}$,
define 
\begin{align*}
    v^{\mu_{h}}(s,a,s')
    & =
    \mu_{h}^{-}(s,a,s')-\mu_{h}^{+}(s,a,s')\\
    e^{\mu_{h},\beta_{h}}_h(s,a,s')
    & =
    ( \mu_{h}^{-}(s,a,s') + \mu_{h}^{+}(s,a,s') ) \epsilon_{h}^{k}(s' \mid s,a) + \beta_{h}(s)-\beta_{h+1}(s'),
\end{align*}
where we always set $\beta_{1}=\beta_{H}=0$. For last, define 
\[
Z_{h}^{k}(v,e)=\sum_{s,a,s'}q_{h}^{k}(s,a,s')e^{B_{h}^{k}(s,a,s'\mid v,e)}.
\]
By \citep[Theorem 4.2]{rosenberg2019full}, we have that
\[
    q_{h}^{k+1}(s,a,s')
    =
    \frac{q_{h}^{k}(s,a,s') 
    e^{B_{h}^{k}(s,a,s' \mid  v^{\mu_{h}^{k}}, e^{\mu_{h}^{k}, \beta_{h}^{k}})}} {Z_{h}^{k}(v^{\mu_{h}^{k}},e^{\mu_{h}^{k},\beta_{h}^{k}})},
\]
where
\[
    \mu^{k},\beta^{k}
    =
    \arg\min_{\beta,\mu\geq0} \sum_{h=1}^{H} \log Z_{h}^{k}(v^{\mu_{h}},e^{\mu_{h},\beta_{h}}).
\]
Now, we have that
\begin{align*}
    \sum_{h}D_{R}(q_{h}^{k}\Vert q_{h}^{k+1}) 
    & =
    \sum_{h}\sum_{s,a,s'}q_{h}^{k}(s,a,s') \log\frac{q_{h}^{k}(s,a,s')}{q_{h}^{k+1}(s,a,s')}\\
    & =
    \sum_{h}\sum_{s,a,s'} q_{h}^{k}(s,a,s')\log\frac{Z_{h}^{k}(v^{\mu_{h}^{k}},e^{\mu_{h}^{k},\beta_{h}^{k}})} {e^{B_{h}^{k}(s,a,s' \mid v^{\mu_{h}^{k}},e^{\mu_{h}^{k},\beta_{h}^{k}})}}\\
    & =
    \underbrace{
    \sum_{h} \log Z_{h}^{k}(v^{\mu_{h}^{k}},e^{\mu_{h}^{k},\beta_{h}^{k}})}
    _{(A)} 
    -
    \underbrace{
    \sum_{h}\sum_{s,a,s'} q_{h}^{k}(s,a,s')B_{h}^{k}(s,a,s' \mid v^{\mu_{h}^{k}},e^{\mu_{h}^{k},\beta_{h}^{k}})}
    _{(B)}.
\end{align*}
By definition of $\mu_{h}^{k},\beta_{h}^{k}$, term $(A)$
can be bounded by
\begin{align*}
    (A) 
    & \leq
    \sum_{h}\log Z_{h}^{k}(0,0)\\
    & =
    \sum_{h}\log(\sum_{s,a,s'}q_{h}^{k}(s,a,s')e^{B_{h}^{k}(s,a,s'\mid0,0)})\\
    & =
    \sum_{h}\log(\sum_{s,a,s'}q_{h}^{k}(s,a,s')e^{-\eta\sum_{j\in\mathcal{F}^{k}}c_{h}^{j}(s,a,s')})\\
    \tag{\ensuremath{\forall x\geq0:e^{-x}\leq1-x+\frac{x^{2}}{2}}} 
    & \leq
    \sum_{h} \log\left(\sum_{s,a,s'}q_{h}^{k}(s,a,s')\left(1-\eta\sum_{j\in\mathcal{F}^{k}}c_{h}^{j}(s,a,s')+\frac{(\eta\sum_{j\in\mathcal{F}^{k}}c_{h}^{j}(s,a,s'))^{2}}{2}\right)\right)\\
    & =
    \sum_{h}\log\left(1-\eta\sum_{s,a,s'}\sum_{j\in\mathcal{F}^{k}}q_{h}^{k}(s,a,s')c_{h}^{j}(s,a,s')+\sum_{s,a,s'}q_{h}^{k}(s,a,s')\frac{(\eta\sum_{j\in\mathcal{F}^{k}}c_{h}^{j}(s,a,s'))^{2}}{2}\right)
    \\
    \tag{\ensuremath{\log(1+x)\leq x}} 
    & \leq
    - \eta\sum_{h}\sum_{s,a,s'}\sum_{j\in\mathcal{F}^{k}} q_{h}^{k}(s,a,s')c_{h}^{j}(s,a,s')+\sum_{h}\sum_{s,a,s'}q_{h}^{k}(s,a,s')\frac{(\eta\sum_{j\in\mathcal{F}^{k}}c_{h}^{j}(s,a,s'))^{2}}{2}.
\end{align*}
Term $(B)$ can be rewritten as
\begin{align*}
    (B)
    & =
    \sum_{h}\sum_{s,a,s'}q_{h}^{k}(s,a,s')(e^{\mu_{h}^{k},\beta_{h}^{k}}(s,a,s')
    +v^{\mu_{h}^{k}}(s,a,s')
    \\
    & \qquad-\eta\sum_{j\in\mathcal{F}^{k}}c_{h}^{j}(s,a,s')
    -\sum_{s''}p^{k}(s''\mid s,a) v^{\mu_{h}^{k}}(s,a,s''))\\
    & =
    \sum_{h}\sum_{s,a,s'} q_{h}^{k}(s,a,s') e^{\mu_{h}^{k},\beta_{h}^{k}}(s,a,s')
    +\sum_{h}\sum_{s,a,s'}q_{h}^{k}(s,a,s')v^{\mu_{h}^{k}}(s,a,s')\\
    & \qquad -
    \eta\sum_{h}\sum_{s,a,s'}\sum_{j\in\mathcal{F}^{k}} q_{h}^{k}(s,a,s')c_{h}^{j}(s,a,s')
    \\
    & \qquad -\sum_{h}\sum_{s,a,s'}\sum_{s''}q_{h}^{k}(s,a,s')p^{k}_{h}(s''\mid s,a)v^{\mu_{h}^{k}}(s,a,s'')\\
    & =
    \sum_{h}\sum_{s,a,s'} q_{h}^{k}(s,a,s')e^{\mu_{h}^{k},\beta_{h}^{k}}(s,a,s')+\sum_{h}\sum_{s,a,s'}q_{h}^{k}(s,a,s')v^{\mu_{h}^{k}}(s,a,s')\\
    & \qquad -
    \eta\sum_{h}\sum_{s,a,s'}\sum_{j\in\mathcal{F}^{k}}q_{h}^{k}(s,a,s')c_{h}^{j}(s,a,s')
    \\
    & \qquad -\sum_{h}\sum_{s,a}\sum_{s''}q_{h}^{k}(s,a)p^{k}(s''\mid s,a)v^{\mu_{h}^{k}}(s,a,s'')\\
    & =
    \sum_{h}\sum_{s,a,s'}q_{h}^{k}(s,a,s')e^{\mu_{h}^{k},\beta_{h}^{k}}(s,a,s')+\sum_{h}\sum_{s,a,s'} q_{h}^{k}(s,a,s')v^{\mu_{h}^{k}}(s,a,s')\\
    & \qquad -
    \eta\sum_{h}\sum_{s,a,s'}\sum_{j\in\mathcal{F}^{k}}q_{h}^{k}(s,a,s')c_{h}^{j}(s,a,s')-\sum_{h}\sum_{s,a,s''} q_{h}^{k}(s,a,s'')v^{\mu_{h}^{k}}(s,a,s'')\\
    & =
    \sum_{h}\sum_{s,a,s'}q_{h}^{k}(s,a,s')e^{\mu_{h}^{k},\beta_{h}^{k}}(s,a,s')
    -\eta\sum_{h}\sum_{s,a,s'}\sum_{j\in\mathcal{F}^{k}} q_{h}^{k}(s,a,s')c_{h}^{j}(s,a,s').
\end{align*}
Overall we get
\begin{align*}
    \sum_{h}D_{R}(q_{h}^{k}\Vert q_{h}^{k+1})
    & \leq
    \sum_{h}\sum_{s,a,s'} q_{h}^{k}(s,a,s')\frac{(\eta\sum_{j\in\mathcal{F}^{k}}c_{h}^{j}(s,a,s'))^{2}}{2}
    \\
    & \qquad -\sum_{h}\sum_{s,a,s'} q_{h}^{k}(s,a,s')e^{\mu_{h}^{k},\beta_{h}^{k}}(s,a,s')\\
    \tag{\ensuremath{\mu_{h}^{k}\geq0}} 
    & \leq
    \sum_{h}\sum_{s,a,s'} q_{h}^{k}(s,a,s')\frac{(\eta\sum_{j\in\mathcal{F}^{k}}c_{h}^{j}(s,a,s'))^{2}}{2}
    \\
    & \qquad - \sum_{h}\sum_{s,a,s'} q_{h}^{k}(s,a,s')(\beta_{h}^{k}(s)-\beta_{h+1}^{k}(s')).
\end{align*}
For last, we show that the second term is $0$, which completes
the proof
\begin{align*}
    \sum_{h}\sum_{s,a,s'} & q_{h}^{k}(s,a,s')(\beta_{h}^{k}(s) - \beta_{h+1}^{k}(s'))
    =
    \\
    & =
    \sum_{h}\Biggl[ \sum_{s,a,s'}q_{h}^{k}(s,a,s')\beta_{h}^{k}(s)-\sum_{s,a,s'}q_{h}^{k}(s,a,s')\beta_{h+1}^{k}(s') \Biggr]\\
    & =
    \sum_{h}\Biggl[ \sum_{s}q_{h}^{k}(s)\beta_{h}^{k}(s)-\sum_{s'}q_{h+1}^{k}(s')\beta_{h+1}^{k}(s') \Biggr]\\
    & =
    \sum_{s}\sum_{h}\bigl[ q_{h}^{k}(s)\beta_{h}^{k}(s)-q_{h+1}^{k}(s)\beta_{h+1}^{k}(s) \bigr]\\
    & =
    \sum_{s}\bigl[ q_{1}^{k}(s)\beta_{1}^{k}(s)-q_{H}^{k}(s)\beta_{H}^{k}(s) \bigr]
    = 0,
\end{align*}
where the last equality follows since $\beta_{1}^{k} = \beta_{H}^{k} = 0$.
\end{proof}

We now use the lemma above to bound the last term in \eqref{eq:O-REPS-regret-with-drift}:
\begin{align*}
    \sum_{k=1}^{K}\langle c^{k},q^{k}-q^{k+d^{k}}\rangle
    & \leq
    \sum_{k=1}^{K}\sum_{h}\sum_{s,a,s'} |q_{h}^{k}(s,a,s')-q_{h}^{k+d^{k}}(s,a,s')|\\
    & \leq
    \sum_{k=1}^{K}\sum_{j=k}^{k+d^{k}-1}\sum_{h}\sum_{s,a,s'} |q_{h}^{j}(s,a,s')-q_{h}^{j+1}(s,a,s')|\\
    \tag{by Pinsker's inequality} 
    & \leq
    2\sum_{k=1}^{K}\sum_{j=k}^{k+d^{k}-1}\sum_{h} \sqrt{2D_{R}(q_{h}^{j}\Vert q_{h}^{j+1})}\\
    \tag{by Jensen's inequality} 
    & \leq
    2\sum_{k=1}^{K}\sum_{j=k}^{k+d^{k}-1} \sqrt{2H\sum_{h}D_{R}(q_{h}^{j}\Vert q_{h}^{j+1})}\\
    \tag{by \cref{lemma:adjacent-o-mesure-KL-bound}} 
    & \lesssim
    \sum_{k=1}^{K}\sum_{j=k}^{k+d^{k}-1}\sqrt{H\sum_{h}\sum_{s,a,s'}q_{h}^{j}(s,a,s')\Bigl(\eta\sum_{i\in\mathcal{F}^{j}}c_{h}^{i}(s,a,s')\Bigr)^{2}}\\
    & \leq
    \sum_{k=1}^{K}\sum_{j=k}^{k+d^{k}-1}\sqrt{H\sum_{h}\sum_{s,a,s'}q_{h}^{j}(s,a,s')(\eta|\mathcal{F}^{j}|)^{2}}\\
    & =
    \eta H\sum_{k=1}^{K}\sum_{j=k}^{k+d^{k}-1}|\mathcal{F}^{j}|
    \leq
    \eta HD,
\end{align*}
where the last inequality is shown in the proof of Theorem 1 in \citep{thune2019nonstochastic}. Combining the above with \eqref{eq:O-REPS-regret-decomposition}, \eqref{eq:O-REPS-error-term} and \eqref{eq:O-REPS-regret-with-drift} completes the proof.

\newpage

\section{Full the details of the empirical evaluation and more experiments}
\label{appendix:experiments}

We conducted our experiments on a grid-world of size $10 \times 10$ i.e., $S = 100$ and horizon $H = 50$. There are four types of states: \textit{Initial state}, $s_\text{init}$, which is always the top-left corner, \textit{goal state}, $s_\text{goal}$, which is always the buttom-right corner, \textit{wall states} which are not reachable and \textit{regular states} which are the rest of the states on the grid. There are four actions $A = \{\texttt{up},\texttt{down},\texttt{right},\texttt{left}\}$. After taking an action, the agent moves with probability $0.9$ towards the adjacent state in the corresponding direction, provided that this is not a wall state or falls outside the grid. With probability $0.1$ the direction is perturbed uniformly. The cost function is defined as $c(s,a) = \indevent{s\ne s_\text{goal}}$.
\paragraph{Implementation of the algorithms.} As presented by \cite{efroni2020optimistic}, under stochastic MDP, the estimated transition $\hat{p}_h^j$ can be replaced with the observed empirical transitions, and instead reduce the cost by order of $1/\sqrt{n_h^j(s,a)}$ during the policy evaluation step. This forces the algorithm to explore states that were not visited enough and keeps the estimated $Q$-function optimistic as our algorithm does. For better computational efficiency we implemented the policy evaluation step in our algorithm with these kind of estimates. All algorithms where run under full-information cost feedback. All algorithms where tested with a fixed learning rate $\eta = 0.1$. Reduction maintains $d_{max} + 1$ copies of OPPO where $d_{max}$ is realized maximal delay. Effectively each copy suffers no delay so this is reduced to standard OPPO \citep{efroni2020optimistic}.  
Each of our experiment take 2-5 hours of computation time on a CPU.

\begin{figure}
    \centering 
    \includegraphics[width=0.48\textwidth]{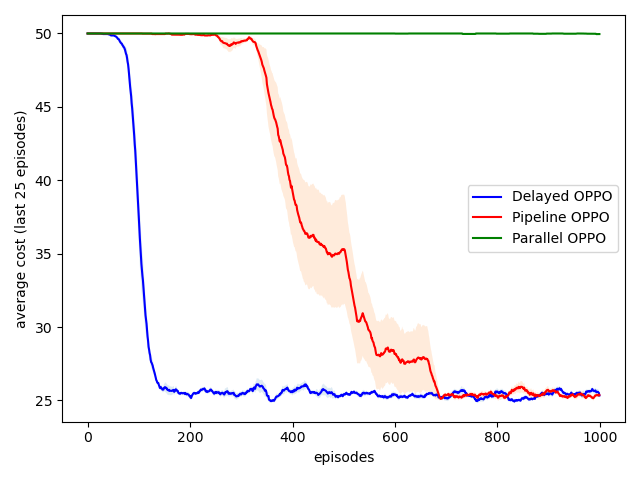}
    \includegraphics[width=0.5\textwidth]{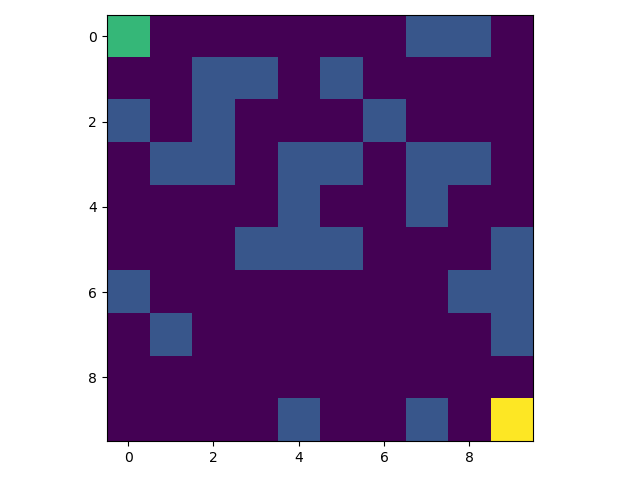}
  \caption{On the left: Average cost of delayed algorithms in grid world with walls with geometrically distributed delays with mean 10.
  On the right: the grid environment with the wall states, where green is $s_{\text{init}}$, yellow is $s_{\text{goal}}$, dark blue are regular states and blue are the wall states.}
  \label{fig:wall_exp}
\end{figure}

\paragraph{Adding wall states.} The results of \cref{fig:dmax_main} where tested on a simple grid without wall states. In \cref{fig:wall_exp} we added wall states so that a more complex dynamics needs to be learned.

Note that the convergence time of all algorithms increases compared to \cref{fig:dmax_main}, as a more complex policy need to be learned. However, Delayed OPPO still keeps its great advantage over the other two alternatives.

\begin{figure}
    \centering 
    \includegraphics[width=0.65\textwidth]{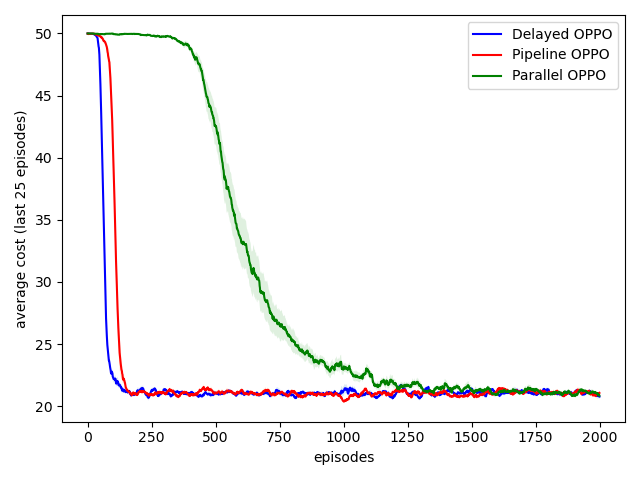}
    \caption{Average cost of delayed algorithms in grid world with walls with geometrically distributed delays with mean $2$.}
    \label{fig:geo2}
\end{figure}

\paragraph{Convergence time of Parallel-OPPO.} In all the experiments we presented so far, Parallel-OPPO have not shown an improvement over time. In order to show the difference on convergence time, we changed the delay distribution to be geometric with mean 2 and increased the number of episodes to $K = 2000$. The results are averaged over 10 runs and appear in \cref{fig:geo2}.

The maximal delay scale approximately as $2\log(K)\approx 15$. While each copy of OPPO that Parallel-OPPO maintains suffers effectively no delay, this is insignificant compared to the fact that Delayed OPPO observe approximately $15$ times more observation then each copy. Pipeline-OPPO performs quite well in this case, as the maximal delay is quite small and approximately after $15$ episodes it has a pipeline of observations. With that being said, note that even when the maximal delay quite small, Delayed OPPO definitely outperforms Pipline-OPPO. In addition, it is sufficient to have a single large delay in order to have a major reduction in the performance of Pipline-OPPO. In real-world application it is quite common to have few large delays. In fact, few delays might be infinite, for example due to packet loss over a network. This would make Pipline-OPPO and Parallel-OPPO completely degenerate, while the effect of few missing observations on Delayed-OPPO is only minor.
\end{document}